\documentclass[10pt,twocolumn,letterpaper]{article}

\usepackage{cvpr}              

\definecolor{cvprblue}{rgb}{0.21,0.49,0.74}
\usepackage[pagebackref,breaklinks,colorlinks,allcolors=cvprblue]{hyperref}

\usepackage{bm}  
\usepackage{soul}
\usepackage{comment}

\def\paperID{16} 
\def\confName{CVPR}
\def\confYear{2025}

\title{MaskAdapt: Unsupervised Geometry-Aware Domain Adaptation Using Multimodal Contextual Learning and RGB-Depth Masking}

\author{
Numair Nadeem\textsuperscript{1} \quad
Muhammad Hamza Asad\textsuperscript{2} \quad
Saeed Anwar\textsuperscript{3} \quad
Abdul Bais\textsuperscript{1} \\
\textsuperscript{1}University of Regina \\
\textsuperscript{2}University Canada West \\
\textsuperscript{3} The University of Western Australia \\
{\tt\small nng794@uregina.ca, muhammadhamza.asad@ucanwest.ca} \\
{\tt\small saeed.anwar@uwa.edu.au, abdul.bais@uregina.ca}
}

\begin{document}
\maketitle
\begin{abstract}

Semantic segmentation of crops and weeds is crucial for site-specific farm management; however, most existing methods depend on labor intensive pixel-level annotations. A further challenge arises when models trained on one field (source domain) fail to generalize to new fields (target domain) due to domain shifts, such as variations in lighting, camera setups, soil composition, and crop growth stages. Unsupervised Domain Adaptation (UDA) addresses this by enabling adaptation without target-domain labels, but current UDA methods struggle with occlusions and visual blending between crops and weeds, leading to misclassifications in real-world conditions. To overcome these limitations, we introduce MaskAdapt, a novel approach that enhances segmentation accuracy through multimodal contextual learning by integrating RGB images with features derived from depth data. By computing depth gradients from depth maps, our method captures spatial transitions that help resolve texture ambiguities. These gradients, through a cross-attention mechanism, refines RGB feature representations, resulting in sharper boundary delineation. In addition, we propose a geometry-aware masking strategy that applies horizontal, vertical, and stochastic masks during training. This encourages the model to focus on the broader spatial context for robust visual recognition. Evaluations on real agricultural datasets demonstrate that MaskAdapt consistently outperforms existing State-of-the-Art (SOTA) UDA methods, achieving improved segmentation mean Intersection over Union (mIOU) across diverse field conditions.

\end{abstract}    
\section{Introduction}
\label{sec:intro}
\begin{figure}[t!]
  \centering
  \includegraphics[width=0.49\textwidth]{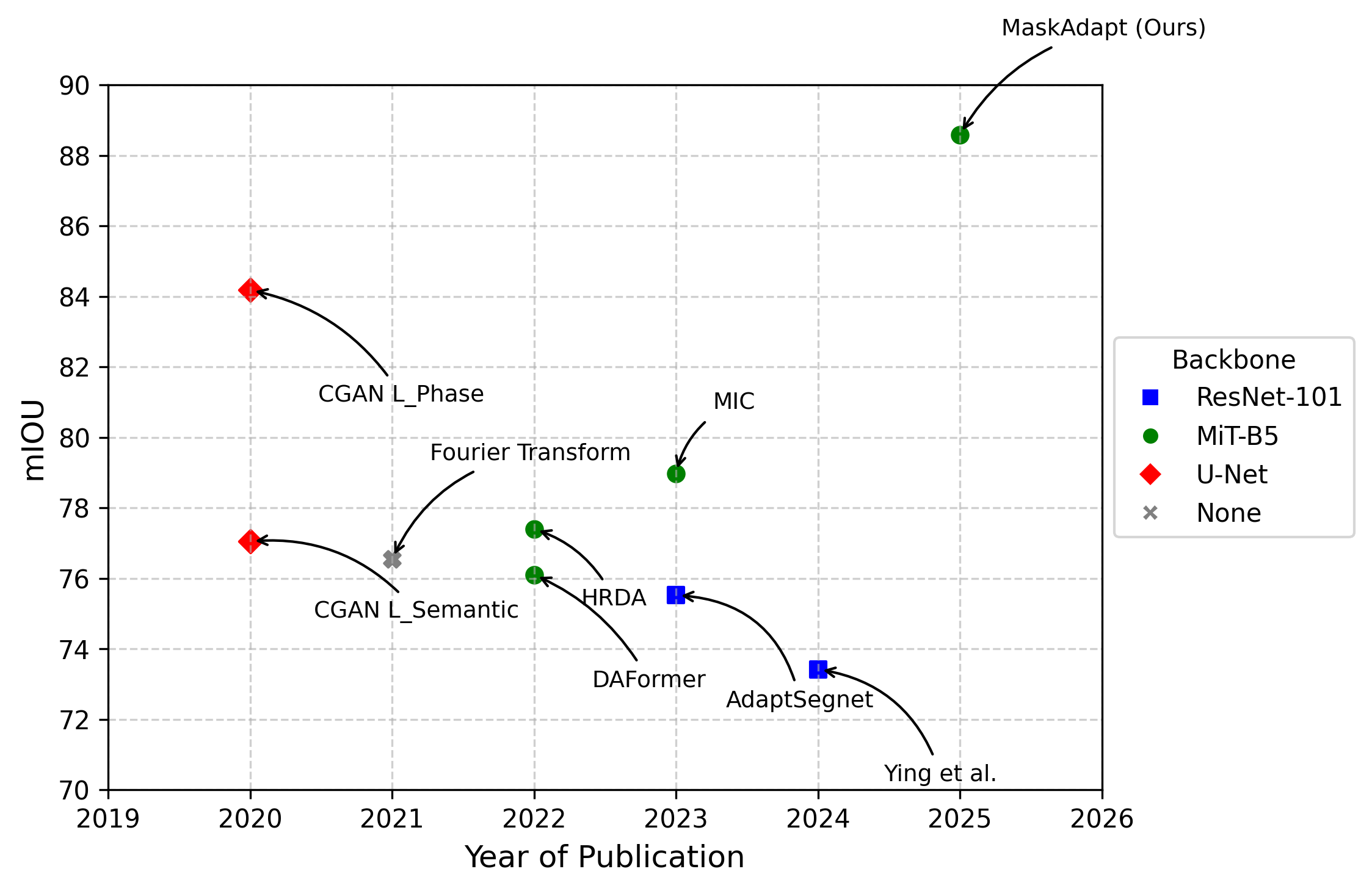}
\caption{Performance comparison of our proposed method with the SOTA UDA methods. Our method MaskAdapt achieves the highest mIOU of 88.5\%.}
\label{fig:placeholder}
\end{figure}
Site-Specific Farm Management (SSFM) is crucial to improve crop yield while optimizing input usage through the precise detection and quantification of crops, weeds, and soil characteristics~\cite{finch2016seed}. Semantic segmentation using RGB field imagery is widely employed to classify and quantify crop, weed, and soil pixels~\cite{asad2024improved, asad2020weed, asad2019weed}. However, the performance of segmentation models deteriorates when trained on one agricultural field (source domain) and tested on another (target domain). This decline is primarily due to domain shifts caused by variations in lighting, camera configurations \& perspectives, soil types, and plant growth stages, which present significant challenges~\cite{bertoglio2023comparative, gogoll2020unsupervised, vasconcelos2021low}. Addressing this issue by annotating diverse agricultural datasets is labor-intensive and time-consuming~\cite{goodfellow2014generative}. In such scenarios, Unsupervised Domain Adaptation (UDA) is essential to adapt models without requiring labeled data from the target domain~\cite{venkateswara2020introduction}.

Although UDA has made significant strides~\cite{hoyer2023domain, hoyer2022hrda}, challenges persist, particularly with RGB-based methods that struggle to distinguish similar textures\cite{10.1007/978-3-031-72933-1_19}, such as those between crops and weeds, leading to false segmentation and lost details~\cite{hoyer2023mic}. Incorporating a geometric representation such as depth offers a solution by providing spatial cues\cite{10.1007/978-3-031-72933-1_19}: uniform depth in crop rows and sharp edges at soil boundaries~\cite{ranftl2021vision}. Yet, this approach falters with classes having minimal height variation, like weeds, which often lack significant height differences from crops, blending into the depth profile and reducing separability. This prompts a pivotal research question: What new feature can we derive from depth, and how to integrate it with depth and RGB features to enhance segmentation accuracy in UDA for classes with minimal elevation variance?

Depth gradients derived from depth maps emerge as a promising solution, capturing spatial transitions, gradual within crop rows and abrupt at crop-weed and crop-soil interfaces from depth maps. Refinement of texture ambiguities and sharpening of boundaries boosts segmentation when depth alone is insufficient, as evidenced in Table \ref{tab:fusion_components}. Integrating these gradients with RGB data and depth maps strengthens multimodal coordination, suggesting that the new feature and its fusion strategy could unlock robust crop-weed-soil delineation in UDA, addressing the limitations of height-dependent depth reliance. 

To leverage the mentioned potential, we adopt a UDA framework for agricultural semantic segmentation, featuring a dual-encoder system: a pre-trained RGB encoder captures visual details, while a lightweight depth encoder processes geometric information. From depth maps, we derive a new feature, depth gradients, computed to highlight spatial transitions, providing a critical signal to distinguish ambiguous regions (such as crop-weed occlusion, overlap regions, or shadowed areas) where RGB alone fails. These features are fused across scales using a depth-gradient-guided cross-attention mechanism, where depth maps and depth gradients refine RGB features to sharpen boundaries and enhance texture clarity. This approach directly leverages the gradient's ability to emphasize structural edges, improving segmentation under domain shifts. We also introduce a geometry-aware multimodal masking strategy, utilizing horizontal patch masks for crop row features, vertical patch masks for crop interrow features, and stochastic masks to break local details for global pattern detection~\cite{hoyer2023mic}. This encourages the model to rely on broader context across modalities~\cite{hoyer2023mic}. Designed as a plug-in network, this solution integrates effortlessly with pre-trained models, enhancing adaptability without requiring extensive retraining.

Our framework enhances semantic segmentation for the agricultural industry by addressing domain shifts in UDA, which is validated using publicly available agricultural datasets. It introduces the following primary contributions:
\begin{itemize}
\item We enhance UDA with depth gradients from RGB-derived depth maps, computed via first-order differences, to guide precise multiscale feature fusion.
\item We propose an enhanced feature fusion module that incorporates a depth-gradient-guided cross-attention mechanism, utilizing depth and its gradient cues to refine RGB features, thereby enhancing fine-detail segmentation in agriculture.
\item We introduce a geometry-aware multimodal masking strategy that leverages horizontal, vertical, and stochastic patch masks to exploit geometric patterns in RGB and depth, complementarily applied to enhance cross-domain adaptation.
\end{itemize}

\section{Related Work}
\label{sec:lit}
\subsection{RGB-D Semantic Segmentation}
Integration of depth information has advanced semantic segmentation, enhancing RGB-based models with geometric cues to improve accuracy and robustness~\cite{zhang2019pattern, chen20203d}. In this context, depth is used in two primary ways. First, it refines RGB features through techniques such as averaging modality output scores to enhance inter-object discrimination, as shown in indoor settings~\cite{long2015fully}. Second, it improves via multilevel fusion across network stages~\cite{wang2016learning} and depth-aware convolutions or pooling~\cite{wang2018depth} that adapt 2D operations to depth-consistent regions. Gated fusion modules~\cite{cheng2017locality} further refine depth integration by dynamically weighting modality contributions. Alternatively, depth serves as a supervisory signal in multitask learning, regularizing RGB features bidirectionally~\cite{zhang2019pattern} or aiding UDA by narrowing domain gaps with geometric consistency~\cite{lin2017cascaded}. Recent methods favor depth-to-RGB refinement, employing attention mechanisms like softmax aggregation~\cite{wang2018understanding} for cross-modal integration or excitation blocks~\cite{hu2018squeeze} to emphasize relevant depth cues.

Modern depth estimation models have significantly improved the ability to derive depth from RGB images. MonoDepth2~\cite{godard2019digging} predicts depth from a single RGB image while UniMatch~\cite{xu2023unifying} utilizes stereo pairs to create precise depth maps. In addition, recently, vision transformers~\cite{ranftl2021vision} have replaced CNNs for depth prediction tasks. Drawing inspiration from these advances, we derive the depth maps using \cite{ranftl2021vision} approach. Our method leverages a depth gradient-based attention map, detailed in Section~\ref{sec:global}, which uses gradients to highlight geometric transitions, guiding RGB features as a supervisory signal. Unlike prior RGB-Depth (RGB-D) methods~\cite{liu2020learning, zhang2023cmx, chen2020bi}, we combine depth and its gradients to refine segmentation robustly across domains.

\subsection{Unsupervised Domain Adaptation}
UDA transfers knowledge from a labeled source domain to an unlabeled target domain, addressing domain shifts that hinder semantic segmentation performance. UDA methods fall into distribution alignment and self-supervised learning categories. Distribution alignment minimizes the feature gaps between source and target domains. By applying statistical moment matching, metrics such as maximum mean discrepancy are minimized to facilitate this alignment (e.g., \cite{long2015learning, sun2017correlation}). This approach draws on various statistical techniques (e.g., \cite{li2021semantic}). Optimal transport aligns domains by connecting source-target samples and minimizing pairwise distances~\cite{hsu2020progressive}, allowing for precise adaptation. Generative methods like CycleGAN~\cite{zhu2017unpaired} align domains at pixel and feature levels via cycle consistency, enhancing visual alignment~\cite{hoffman2018cycada}. Adversarial learning, which is popular in segmentation~\cite{ganin2016domain, schwonberg2023survey}, uses discriminators to enforce domain-invariant features; however, instability can occur~\cite{wang2019evolutionary, hoyer2022hrda}. Self-training, a stable alternative~\cite{wang2019evolutionary, hoyer2023mic}, employs pseudo-labels refined by confidence thresholds~\cite{mei2020instance} or consistency across augmentations~\cite{tranheden2021dacs}, often using teacher networks~\cite{tarvainen2017mean}. Recent efforts integrate multi-resolution feature maps~\cite{hoyer2022hrda} and depth-informed adaptation~\cite{xu2023unifying}, aligning with our RGB-depth focus, while hybrid approaches blending adversarial and self-training~\cite{wang2020classes} tackle instability. Challenges in pseudo-label reliability and target supervision persist, motivating our multi-scale, depth-enhanced UDA approach.

\begin{figure}[tbp]
\centering
 \includegraphics[width=0.65\textwidth]{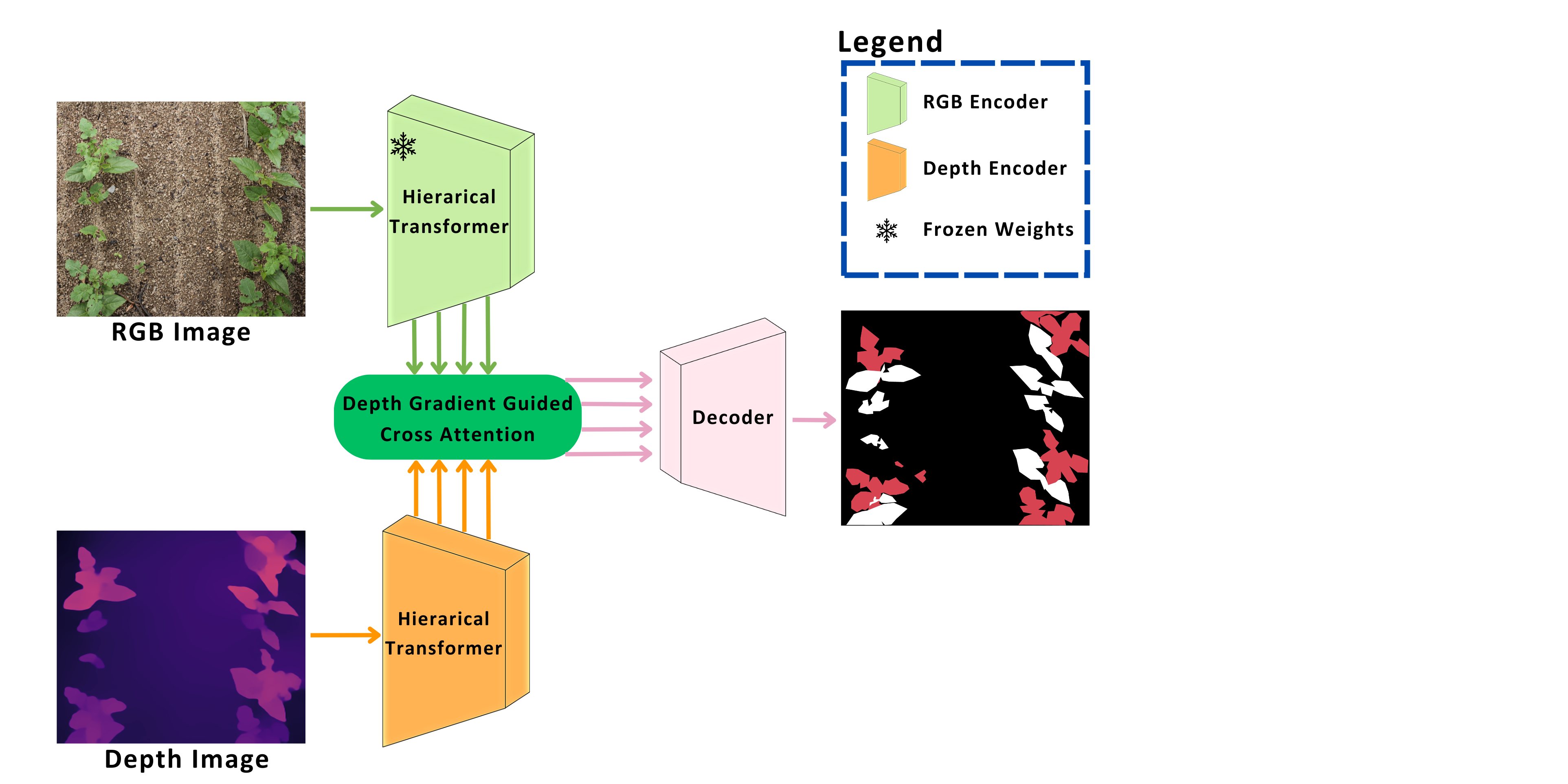}  
\caption{Enhanced Feature Fusion: A dual-encoder architecture combines a pre-trained RGB encoder for visual features with a custom lightweight depth encoder for depth cues, fused via a depth-gradient-guided cross-attention mechanism across multiple scales to improve boundary refinement and segmentation accuracy under domain shifts. } 
\label{fig:main_sas_curve} 
\end{figure}

\subsection{Masked Image Modeling}

Masked image modeling involves learning robust representations by reconstructing images intentionally corrupted through masking, first explored with denoising autoencoders~\cite{venkateswara2017deep} and context encoders~\cite{pan2019transferrable}, which laid the groundwork for self-supervised learning in vision.

The Vision Transformers (ViT)~\cite{dosovitskiy2021image} sparked a new wave of self-supervised methods for pre-training vision models using masked images. These techniques train models to reconstruct hidden parts of images, such as raw pixels~\cite{chen2019learning, dong2021peco, hoffman2018cycada, hoyer2019grid, wu2021one}, image patch tokens~\cite{zhang2019category, choi2019self}, or deeper feature layers~\cite{chen2021scale, vu2019dada}, allowing efficient learning from large image datasets for tasks like scene analysis~\cite{wei2022masked}. Building on ViT, the Masked Autoencoder (MAE)~\cite{hoyer2019grid} offers an efficient pre-training framework through partial image occlusion, masking random subsets of input patches and training a robust ViT-based encoder~\cite{dosovitskiy2021image} to process only visible patches, while a lightweight decoder reconstructs occluded regions. Similarly, Masked Image Consistency (MIC)~\cite{hoyer2023mic} module enforces prediction consistency between masked target images and pseudo-labels via an EMA teacher, while MICDrop~\cite{10.1007/978-3-031-72933-1_19} extends this by masking RGB and depth features in complementary fashion.

Building upon MIC's pretraining efficacy and MICDrop's cross-modality masking, our methodology leverages a pretrained RGB encoder and a trained depth encoder with a geometry-aware multimodal masking strategy, applying horizontal, vertical, and stochastic masks across RGB and depth modalities to exploit spatial structures, such as crop row alignments and crop-weed boundaries, for UDA.

\begin{figure}[!tbp]
         \centering
         \includegraphics[width=0.5\textwidth]{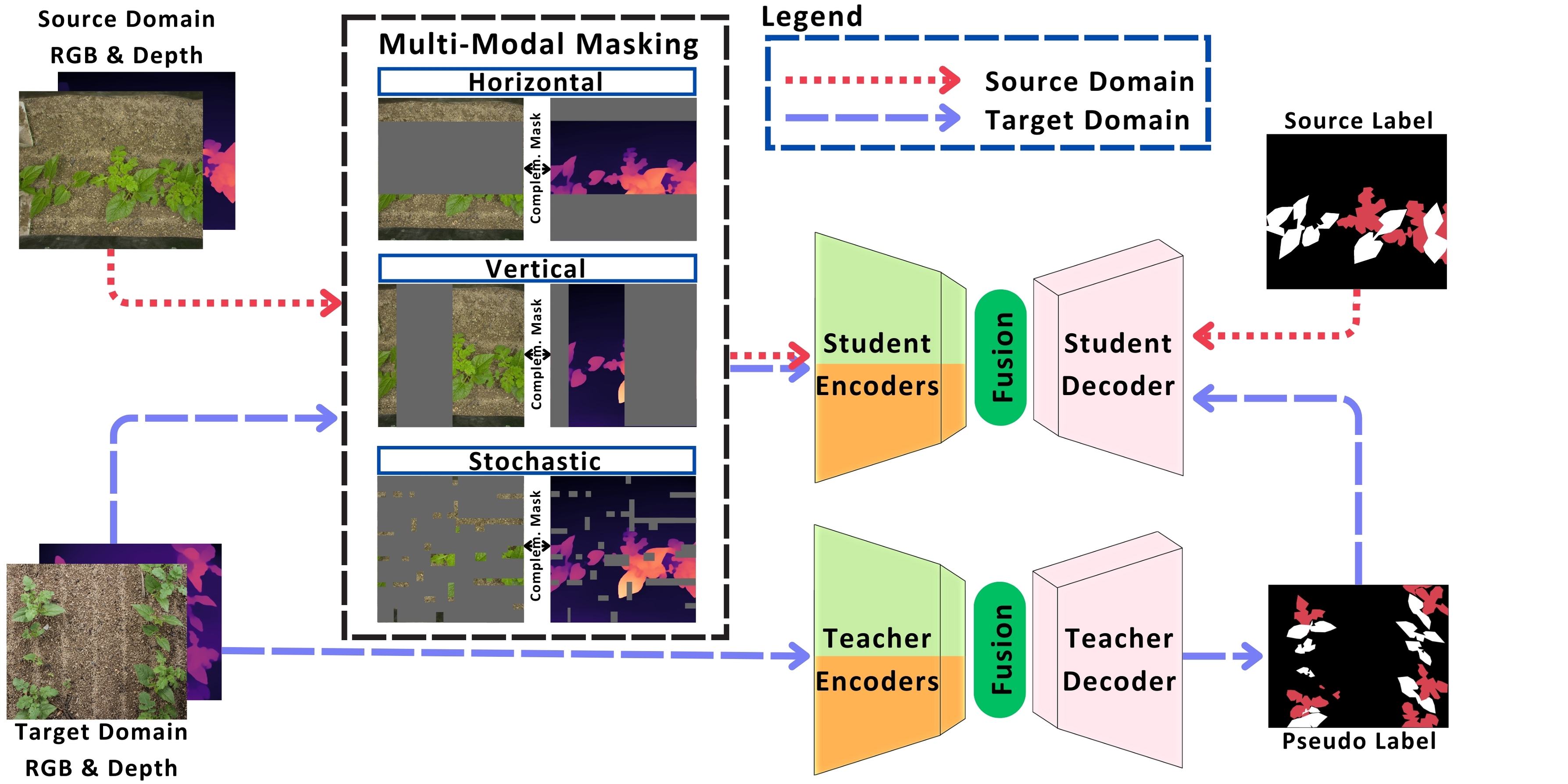}
         \caption{Illustration of our training pipeline featuring Multimodal Masking in Cross-Domain Learning: Source and target images are masked at the input level—initially masking source domain inputs, then shifting to target domain inputs as pseudo-label confidence increases—using three strategies (horizontal, vertical, stochastic) with a dynamic, time-dependent masking ratio, where RGB and depth inputs are masked complementarily. Student encoders process these masked inputs, integrated via Enhanced Feature Fusion block, and fed to the decoder for final segmentation prediction. }
         \label{multi_resolution_training}
\end{figure}

\section{Methodology}
\label{sec:formatting}
\vspace{1mm}\noindent\textbf{Problem Statement}. In the UDA setting, we address a scenario involving a labeled source domain \( D_s = \{(x_s^i, y_s^i)\}_{i=1}^{n_s} = (X_s, Y_s) \) with labeled samples \( n_s \), where \( X_s = \{x_s^i\} \) represents the source input data and \( Y_s = \{y_s^i\} \) denotes the corresponding labels, and a target domain \( D_t = \{(x_t^i)\}_{i=1}^{n_t} \ = (X_t) \) with labeled samples \( n_t \), where \( X_t = \{x_t^i\} \) represents the target input data. Both domains share the same label space \( \{1, 2, \dots, K\} \), where \( K \) is the number of semantic classes for segmentation, but have distinct distributions. Our objective is to bridge the domain gap between \( X_s \) and \( X_t \) using labeled source data \( (X_s, Y_s) \) and unlabeled target data \( X_t \), with performance evaluated on a set of labeled holdout validation of the target domain.

\vspace{1mm}\noindent\textbf{Preliminaries}.
In semantic segmentation, models are trained on a labeled source domain using supervised methods. However, adapting to an unlabeled target domain (\( X_t \)) requires bridging the domain gap, as source-only training falters due to distributional shifts. UDA balances supervised source loss with unsupervised target loss, shaped by adversarial training~\cite{chen2021scale,tsai2018learning} or self-training~\cite{hoyer2022daformer,tranheden2021dacs}. We follow recent methods~\cite{hoyer2022hrda,hoyer2023mic} using a student-teacher framework, where the teacher updates via EMA~\cite{tarvainen2017mean}, crafts pseudo-labels for target images to guide the student. Following~\cite{tranheden2021dacs}, the student sees heavily augmented images, and the teacher weakly augmented ones, ensuring consistency. Moreover, we utilize hierarchical encoders to generate multiscale feature maps, enabling precise segmentation with fine-grained detail. We assess our approach by extending pre-trained encoders~\cite{hoyer2022daformer,hoyer2022hrda} with depth from RGB via Vision Transformers~\cite{ranftl2021vision}, enhancing cross-domain accuracy.

\subsection{Enhanced Feature Fusion Module}
To address the challenges in agricultural semantic segmentation, our method employs a dual-encoder system: a pre-trained RGB encoder extracts visual features, while a lightweight depth encoder, initialized with pre-trained ImageNet\cite{geirhos2018imagenet} weights and then fine-tuned, processes depth information. We derive depth gradients from depth maps computed via first-order differences to enhance feature representation. These are fused with RGB features across multiple scales using a depth-gradient-guided cross-attention mechanism, as detailed in Section~\ref{sec:global}. This streamlined approach optimizes segmentation accuracy under domain shifts by refining boundaries and efficiently resolving ambiguities. 
\begin{figure}[tbp]
    \centering
        \includegraphics[width=0.3\textwidth]{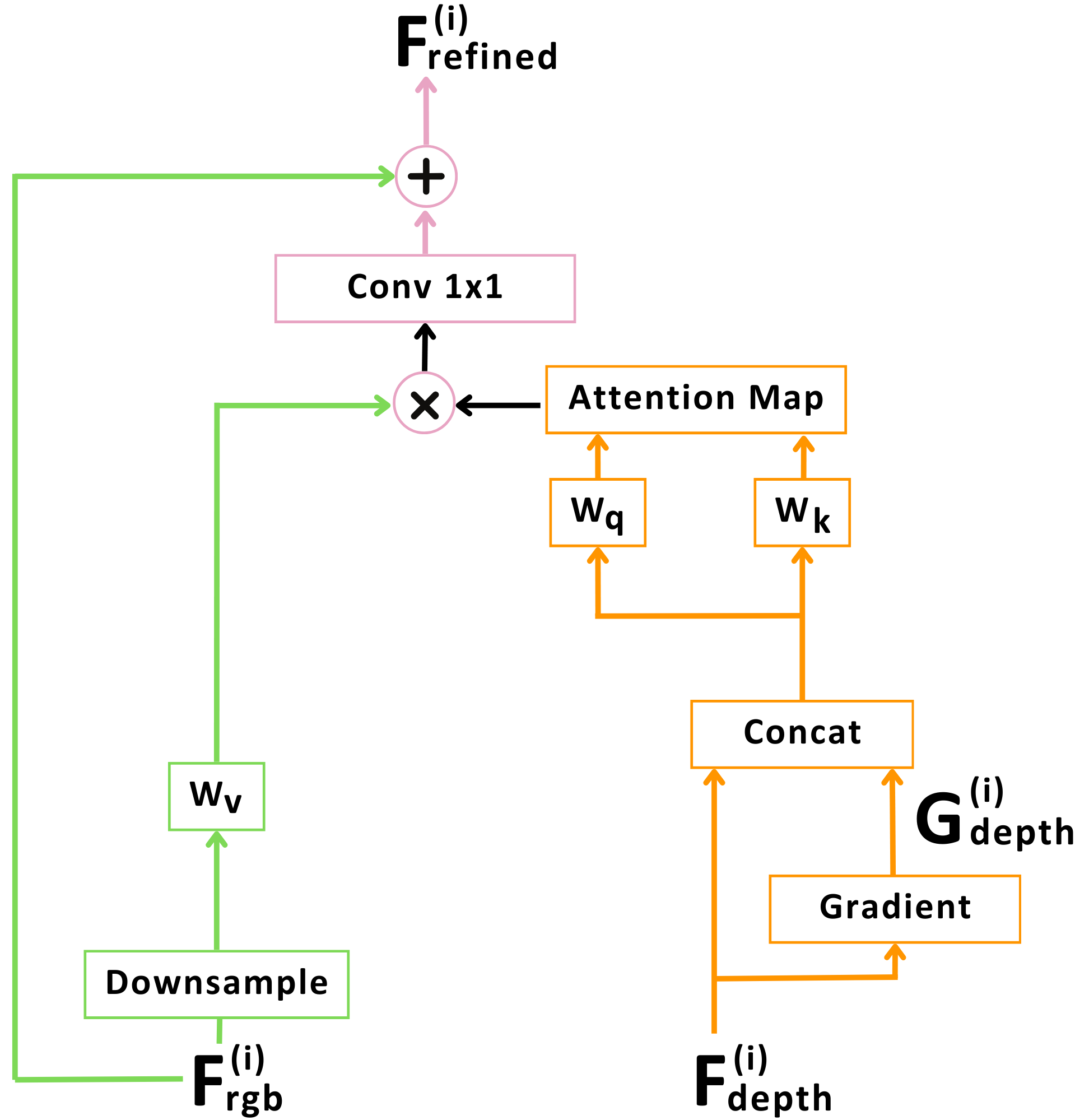}
    
    \caption{Our Depth Gradient-Guided Cross-Attention mechanism comprises depth features with gradients and RGB features. The depth stream computes spatial gradient magnitude to highlight geometric transitions (e.g., crop-weed boundaries), while the RGB stream uses cross-attention guided by these depth cues to aggregate contextual features. The unified representation combines precise boundary information from depth with robust appearance features from RGB for accurate segmentation.} 
    \label{fig:cross-attn} 
\end{figure}

\subsubsection{Depth Gradient-Guided Cross-Attention}
\label{sec:global}

 To utilize depth features as a guiding principle for RGB features, we propose a depth gradient-guided cross-attention mechanism, shown in Figure \ref{fig:cross-attn}, which leverages multiscale depth features and their spatial gradients to guide the aggregation of RGB features, allowing precise segmentation of large-scale agricultural structures. The approach computes spatial depth gradients \( G_{\text{depth}} \) from depth features \( F_{\text{depth}} \) using first-order differences, concatenates them channel-wise, and downsamples the fused depth gradient features to project into queries and keys, while RGB features act as values, allowing the cross-attention mechanism to correlate regions with similar depth profiles (e.g., crop rows) and sharp gradients (e.g., crop-weed-soil boundaries).

Let \( F_{\text{rgb}}^{(i)} \in \mathbb{R}^{H_i \times W_i \times C} \) denote the RGB and \( F_{\text{depth}}^{(i)} \in \mathbb{R}^{H_i \times W_i \times C} \) denote depth feature maps with \( i \) representing encoder level (\( i \in \{1, 2, 3, 4\} \)). Here \( H_i = \frac{H}{2^{i-1}} \), \( W_i = \frac{W}{2^{i-1}} \), and \( C \) is the channel dimension. To explicitly leverage geometric cues critical for agricultural scenes (e.g., boundaries between crop rows, weed and soil), we compute spatial depth gradient magnitude \( G_{\text{depth}}^{(i)} \in \mathbb{R}^{H_i \times W_i \times 1} \) using first-order differences. These gradients highlight regions of sharp depth transitions (e.g., crop-weed-soil boundaries) and gradual changes (e.g., within crop rows). The raw depth features and gradients are concatenated channel-wise:
\begin{equation}
F_{\text{depth+grad}}^{(i)} = \text{Concat}(F_{\text{depth}}^{(i)}, G_{\text{depth}}^{(i)}) \in \mathbb{R}^{H_i \times W_i \times (C+1)}.
\label{eq:concat_features}
\end{equation}

To address the quadratic complexity \( O((HW)^2 C + HW C^2) \) of standard global attention as suggested in \cite{10.1007/978-3-031-72933-1_19}, we bilinearly downsample \( F_{\text{depth+grad}}^{(i)} \) and \( F_{\text{rgb}}^{(i)} \) during training through various pooling factors, reducing spatial dimensions while preserving structural patterns. For inference, pooling factors are adjusted to retain finer details in higher-resolution levels. Let \( F_{\text{depth+grad}}^{(i), \downarrow} \in \mathbb{R}^{\frac{H_i}{p_i} \times \frac{W_i}{p_i} \times (C+1)} \) denote the downsampled features, where \( p_i \) is the pooling factor for level \( i \). These are linearly projected into queries \( Q^{(i)} \) and keys \( K^{(i)} \):
\begin{equation}
Q^{(i)} = F_{\text{depth+grad}}^{(i), \downarrow} W_q^{(i)} \in \mathbb{R}^{\frac{H_i}{p_i} \times \frac{W_i}{p_i} \times d_k},
\label{eq:query}
\end{equation}
\begin{equation}
K^{(i)} = F_{\text{depth+grad}}^{(i), \downarrow} W_k^{(i)} \in \mathbb{R}^{\frac{H_i}{p_i} \times \frac{W_i}{p_i} \times d_k},
\label{eq:key}
\end{equation}
where \( W_q^{(i)}, W_k^{(i)} \in \mathbb{R}^{(C+1) \times d_k} \) are learnable weights. The corresponding RGB features \( F_{\text{rgb}}^{(i), \downarrow} \in \mathbb{R}^{\frac{H_i}{p_i} \times \frac{W_i}{p_i} \times C} \) are projected to values \( V^{(i)} \):
\begin{equation}
V^{(i)} = F_{\text{rgb}}^{(i), \downarrow} W_v^{(i)} \in \mathbb{R}^{\frac{H_i}{p_i} \times \frac{W_i}{p_i} \times d_v},
\label{eq:value}
\end{equation}
with \( W_v^{(i)} \in \mathbb{R}^{C \times d_v} \). The cross-attention computes the aggregation of RGB features across all spatial positions as: 
\begin{equation}
F_{\text{global}}^{(i)} = \left( \text{softmax} \left( \frac{Q^{(i)} K^{(i)\top}}{\sqrt{d_k}} \right) \right) V^{(i)} \in \mathbb{R}^{\frac{H_i}{p_i} \times \frac{W_i}{p_i} \times d_v}
\label{eq:combined_attention_global}
\end{equation}
The aggregated features are upsampled to \( H_i \times W_i \) via bilinear interpolation and fused with the original RGB features through a residual connection:
\begin{equation}
F_{\text{refined}}^{(i)} = F_{\text{rgb}}^{(i)} + \text{Conv}_{1 \times 1} ( F_{\text{global}}^{(i), \uparrow} ) \in \mathbb{R}^{H_i \times W_i \times C}.
\label{eq:refined_features}
\end{equation}

\subsection{Geometry Aware Multimodal Masking in Cross-Domain Learning}
To effectively learn from multimodal RGB and depth data, models must integrate local details and contextual relationships. In supervised learning, labels guide this process, but the lack of target labels complicates it in UDA, especially to distinguish similar classes such as crop from weeds~\cite{asad2024improved}. Masking aids UDA by forcing the model to learn missing data from unmasked regions, learning domain-invariant contextual details that enhance generalization across domains~\cite{hoyer2023mic}. This occurs through occlusion, forcing reliance on spatial patterns and modality synergy rather than domain-specific features.

We propose \emph{geometry aware multimodal masking in cross-domain learning} to boost context learning between labeled source and unlabeled target domains. We employ a scheduling approach parameterized by iteration $t$, initially masking source domain data and later incorporating target domain data as prediction confidence exceeds a threshold, reducing pseudo-label noise. We use complementary masking which corrupts information across modalities in a complementary fashion; hence the model relies on features from both RGB and depth, enhancing cross-model inference. For agricultural images with crops planted in rows, we employ three geometry-aware masking strategies: horizontal, vertical, and stochastic. At each iteration $t$, we randomly select one of these mask types to apply. The masking ratio $m_t$ is time-dependent: during initial training, $m_t$ remains low (e.g., 10\%-20\%) to allow supervised learning on source data, then increases as training progresses, reaching high values like 80\% to elevate difficulty and enhance adaptation. The masking formulation is defined as:
\begin{itemize}
    \item $M_{\text{$\Theta$,rgb}}(a,b) = [\gamma > m_t], \quad \gamma \sim \text{Uniform}(0,1)$
    \item $M_{\text{$\Theta$,depth}}(a,v) = 1 - M_{\text{$\Theta$,rgb}}(a,b)$
\end{itemize}
where \(\text{$\Theta$}\in \{horizontal, vertical, stochastic\} \), $(a,b)$ is the block index, and $\gamma \sim \text{Uniform}(0,1)$ indicates $\gamma$ is a random variable drawn from a uniform distribution over the interval $[0, 1]$, determining whether a block is masked based on the threshold $m_t$.

\vspace{1mm}\noindent\textbf{Horizontal Masking}.
For horizontal crop row images, horizontal masking utilizes complete row patches to occlude crop rows, forcing the model to deduce row structure from vertical context or depth.

\vspace{1mm}\noindent\textbf{Vertical Masking}.
This strategy applies vertical column patches to target soil bands between rows for images with horizontal crop rows. It disrupts continuity across rows, compelling the model to infer inter-row properties using horizontal context or depth inputs.

\vspace{1mm}\noindent\textbf{Stochastic Masking}.
Stochastic masking employs random patches throughout the image, regardless of row orientation. It disrupts local details, encouraging global reasoning to identify broader patterns, aiding in detecting widespread field anomalies across modalities.

Designed as a plug-in network, this depth gradient-guided cross-attention and geometry aware multi-modal masking-based approach integrates with pre-trained models, leveraging these masking strategies to balance modality use and promote robust, spatially aware feature learning.

\begin{table*}[!tbp]
\caption{Comparison of our model's performance with SOTA methods on different source-target datasets using IOU-Background, IOU-Crop, IOU-Weed, and mIOU, averaged over 3 random data sampling seeds. The boldface indicates the best-performing model on the target domain. Our method outperforms all SOTA UDA methods.}
\label{SOTA_comp}
\centering
\resizebox{0.5\textwidth}{!}{
\begin{tabular}{c|c|l|ccclcclccc}
\hline
\textbf{Source} & \textbf{Target}  & \textbf{Metric} & \rotatebox{90}{Fourier Trans. \cite{vasconcelos2021low}} & \rotatebox{90}{CGAN L\_semantic \cite{gogoll2020unsupervised}} & \rotatebox{90}{CGAN L\_phase \cite{yang2020phase}} & \rotatebox{90}{AdaptSegNet \cite{ilyas2023overcoming}} & \rotatebox{90}{CBST \cite{zou2018unsupervised}} & \rotatebox{90}{\cite{huang2024unsupervised}} & \rotatebox{90}{DAFormer \cite{hoyer2022daformer}}& \rotatebox{90}{HRDA \cite{hoyer2022hrda}} & \rotatebox{90}{MIC \cite{hoyer2023mic}} & \rotatebox{90}{Ours} \\ 
\hline
Bean & Bean & IOU-B & 97.41 & 97.67 & 97.55 & 97.10 & 55.70 & \textbf{99.16} & 97.10 & 97.43 & 97.91 & 98.18\\
     BIPBIP & WeedElec & IOU-C & 79.24 & 82.62 & 81.12 & 78.38 & 25.76 & 83.49 & 78.41 & 81.78 & 82.36 & \textbf{85.16}\\
     & & IOU-W & 59.69 & 67.86 & 63.71 & 67.28 & 11.54 & 64.34 & 57.62 & 62.01 & 71.64 & \textbf{74.62} \\
     & & mIOU & 78.78 & 82.71 & 80.79 & 80.92 & 31.00 & 82.30 & 77.71 & 80.41 & 83.97 & \textbf{85.99} \\ \hline
Bean & Bean & IOU-B & 97.92 & 98.03 & 98.01 & 96.37 & 60.41 & 96.08 & \textbf{98.23} & 97.74 & 99.11 & 98.11 \\
     WeedElec & BIPBIP & IOU-C & 82.22 & 82.24 & 83.92 & 76.50 & 41.07 & 77.38 & 80.67 & 83.52 & 85.59 & \textbf{87.01} \\
     & & IOU-W & 70.43 & 67.91 & 71.38 & 66.71 & 05.03 & 75.06 & 65.61 & 68.37 & 69.12 & \textbf{76.35} \\
     & & mIOU & 83.52 & 82.73 & 84.43 & 79.86 & 35.50 & 82.84 & 81.50 & 83.21 & 84.60 & \textbf{87.15} \\ \hline
Maize & Maize & IOU-B & 98.61 & 98.14 & 94.67 & 98.10 & 74.33 & 97.04 & 95.32 & 98.41 & 99.02 & \textbf{99.73} \\
      BIPBIP & WeedElec & IOU-C & 83.89 & 52.93 & 48.71 & 83.58 & 36.87 & 83.86 & 85.94 & 87.32 & 88.12 & \textbf{89.66} \\
      & & IOU-W & 69.01 & 25.18 & 50.19 & 63.72 & 15.43 & \textbf{77.40} & 67.49 & 69.89 & 68.79 & 74.97 \\
      & & mIOU & 83.83 & 58.75 & 64.52 & 81.80 & 42.21 & 86.10 & 82.91 & 85.21 & 85.31 & \textbf{88.12} \\ \hline
Maize & Maize & IOU-B & 96.32 & 96.07 & 97.33 & 93.86 & 65.72 & 94.08 & 93.81 & 94.71 & 94.18 & \textbf{96.78} \\
      WeedElec & BIPBIP & IOU-C & 73.99 & 73.23 & 82.96 & 67.18 & 20.12 & 72.63 & 69.79& 80.10 & 66.48 & \textbf{88.64} \\
      & & IOU-W & 63.96& 61.85 & 72.27 & 65.52 & 20.64 & 53.55 & 64.70 & 63.39 & 76.28 & \textbf{80.32} \\
      & & mIOU & 76.54& 77.05 & 84.18 & 75.52 & 35.50 & 73.42 & 76.10& 77.40& 78.98 & \textbf{88.58} \\ \hline
Bean & Bean & IOU-B & 93.57 & 82.94 & 85.51 & 92.53 & 75.94 & 93.46 & \textbf{95.61} & 94.62 & 94.45 & 92.05 \\
      2019 & 2021 & IOU-C & 79.43 & 44.37 & 83.42 & 68.17 & 36.33 & 64.34 & 86.12 & 86.76 & 84.83 & \textbf{87.57} \\
      & & IOU-W & 48.41 & 34.61 & 27.71 & 41.17 & 05.24 & \textbf{66.81} & 56.77 & 54.47 & 62.36 & 66.32 \\
      & & mIOU & 73.83 & 53.97 & 65.54 & 67.20 & 39.17 & 74.87 & 79.49 & 78.62 & 80.55 & \textbf{81.98} \\ \hline
Bean & Bean & IOU-B & 97.79 & 87.72 & 92.38 & 97.36 & 86.15 & 97.23 & 95.85 & \textbf{97.93} & 95.34 & 93.81 \\
     2021 & 2019 & IOU-C & 77.24 & 35.35 & 54.81 & 76.50 & 44.22 & 82.58 & 75.11 & 81.59 & 83.73 & \textbf{85.39} \\
     & & IOU-W & 61.22 & 04.79 & 41.32 & 57.86 & 16.42 & 66.61 & 64.72 & 68.31 & 70.23 & \textbf{73.07} \\
     & & mIOU & 78.75 & 42.62 & 62.83 & 77.24 & 48.93 & 82.14 & 78.56 & 82.61 & 83.10 & \textbf{84.09} \\ \hline
Maize & Maize & IOU-B & 91.24 & 67.28 & 68.37 & 89.27 & 86.32 & 93.31 & 90.29 & 91.95 & 91.88 & \textbf{93.47} \\
      2019 & 2021 & IOU-C & 83.38 & 03.23 & 14.47 & 79.73 & 27.10 & 83.14 & 85.39 & 87.26 & 85.50 & \textbf{88.51} \\
      & & IOU-W & 47.69 & 01.21 & 00.12 & 44.24 & 06.76 & 57.34 & 54.21 & 55.82 & 62.15 & \textbf{66.71} \\
      & & mIOU & 74.10 & 23.90 & 27.65 & 71.08 & 40.06 & 77.93 & 76.63 & 78.34 & 79.84 & \textbf{82.89} \\ \hline
Maize & Maize & IOU-B & 97.10 & 96.64 & 82.73 & 96.97 & 87.17 & 97.14 & 94.31 & 93.40 & 94.96 & \textbf{97.98} \\
      2021 & 2019 & IOU-C & 80.05 & 66.53 & 34.07 & 81.56 & 69.22 & 83.68 & 66.87 & 68.13 & 69.54 & \textbf{89.26} \\
      & & IOU-W & 73.51 & 48.34 & 21.11 & 65.67 & 17.82 & 73.88 & 60.61 & 62.27 & 64.29 & \textbf{74.12} \\
      & & mIOU & 83.70 & 70.50 & 46.00 & 81.40 & 58.07 & 84.90 & 73.93 & 74.60 & 76.26 & \textbf{87.12} \\ \hline
\end{tabular}}
\end{table*}
\section{Experiments}
\subsection{Implementation Details}
\textbf{Datasets}. We use six maize and bean datasets from the ROSE Challenge~\cite{bertoglio2023comparative}, collected by two distinct robotic platforms (BIPBIP and WeedElec teams)  under varying camera and field conditions, introducing significant domain gaps. Datasets include RGB images and masks (four weed classes combined into one~\cite{bertoglio2023comparative, huang2024unsupervised}), with 1000 labeled images from 2019 (125 per crop per team) and BIPBIP's 250 images per crop from 2019 and 2021. We train on the source domain, split target domain data for training/validation, and apply augmentations suggested in \cite{tranheden2021dacs}. 
We perform two domain adaptations: images from different team's datasets within May 2019 (BIPBIP and WeedElec for maize and beans) and the same team dataset across different growth stages ( BIPBIP-May 2019 and September 2021). This assess our method's effectiveness in handling domain gaps.

\vspace{1mm}\noindent\textbf{Depth Images}. We obtain depth image estimations for all source and target images from Vision Transformer ~\cite{ranftl2021vision} via a pre-trained model trained on large-scale RGB data.

\vspace{1mm}\noindent\textbf{Network Architecture}. Our network is based on the \emph{MMSegmentation} \cite{mmseg2020} framework. \emph{MIC} \cite{hoyer2023mic} is the base architecture/baseline with \emph{MiT-B5} \cite{xie2021segformer} as the RGB-encoder block and \emph{Atrous Spatial Pyramid Pooling (ASPP)} decode head as context-aware feature fusion decoder. For depth feature extractor, we use the lightweight MiT-B3~\cite{mitb3}, processing these depth maps to extract features for our dual-encoder UDA system.

\begin{figure*}[!tbp]
\label{CCUDA_comp}
\begin{center}
\resizebox{\textwidth}{!}{
\begin{tabular}{c@{ }c@{ }c@{}c@{ }c@{ }c@{ }c}

\includegraphics[width=.15\textwidth]{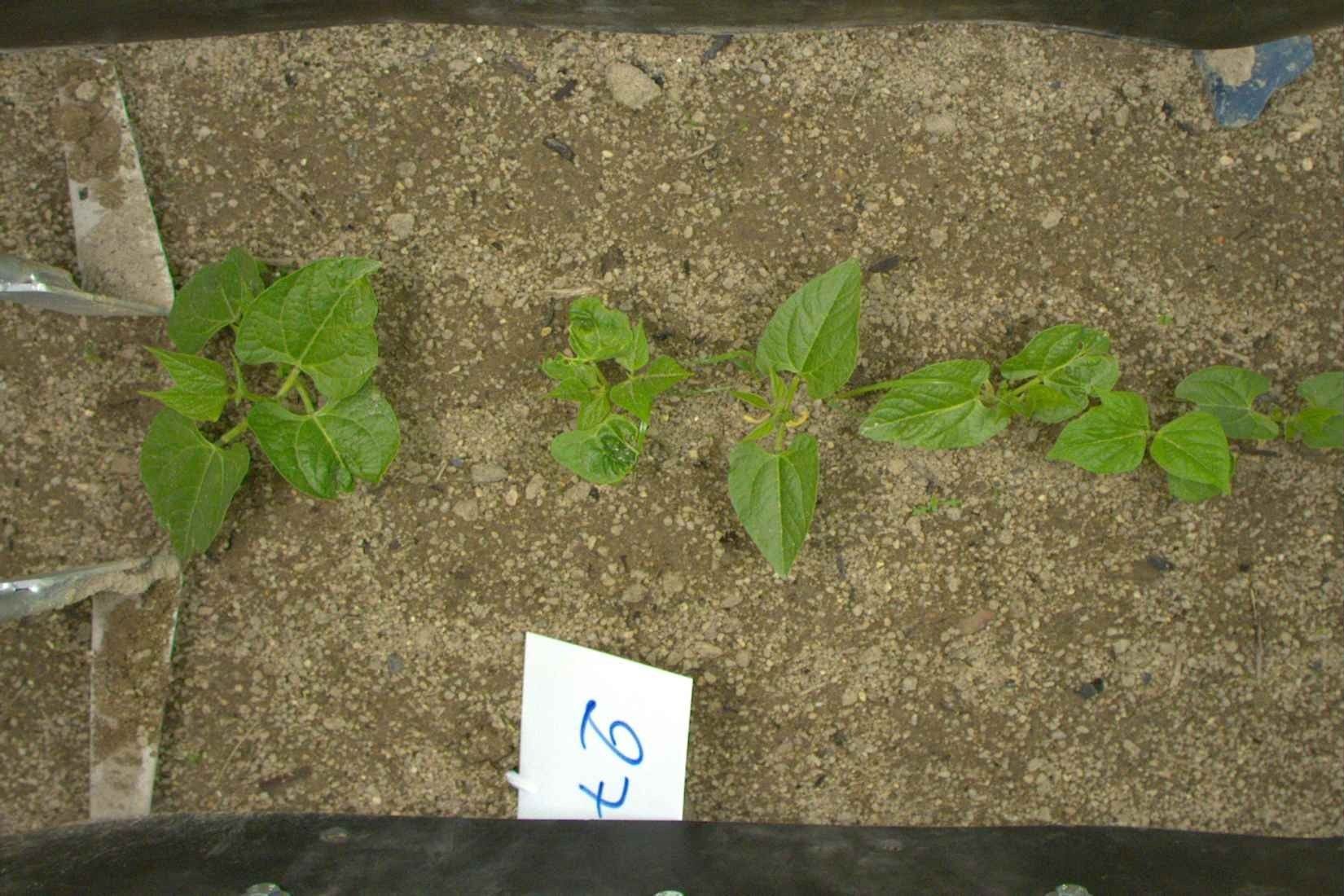}&\includegraphics[width=.15\textwidth]{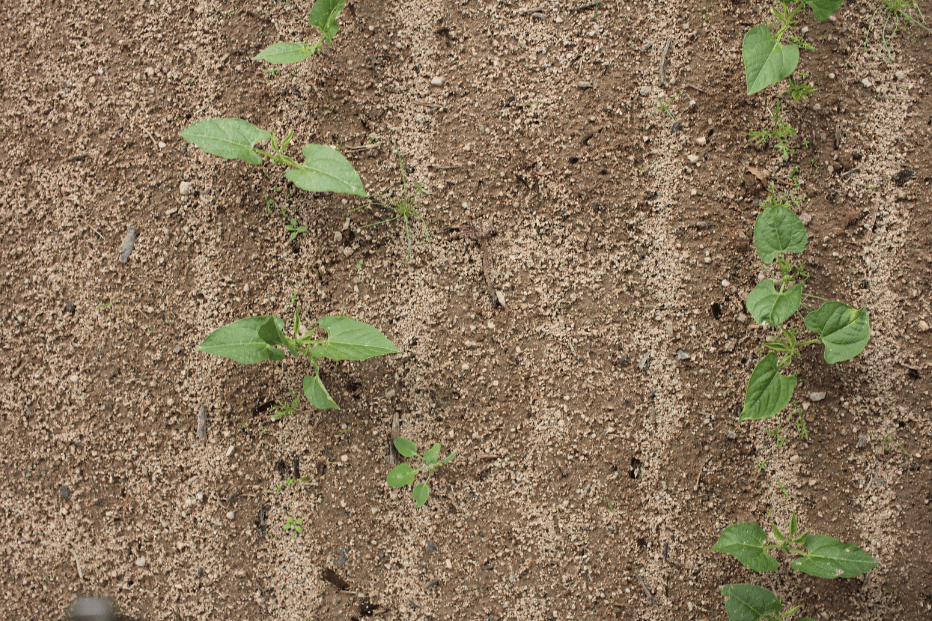}&\includegraphics[width=.15\textwidth]{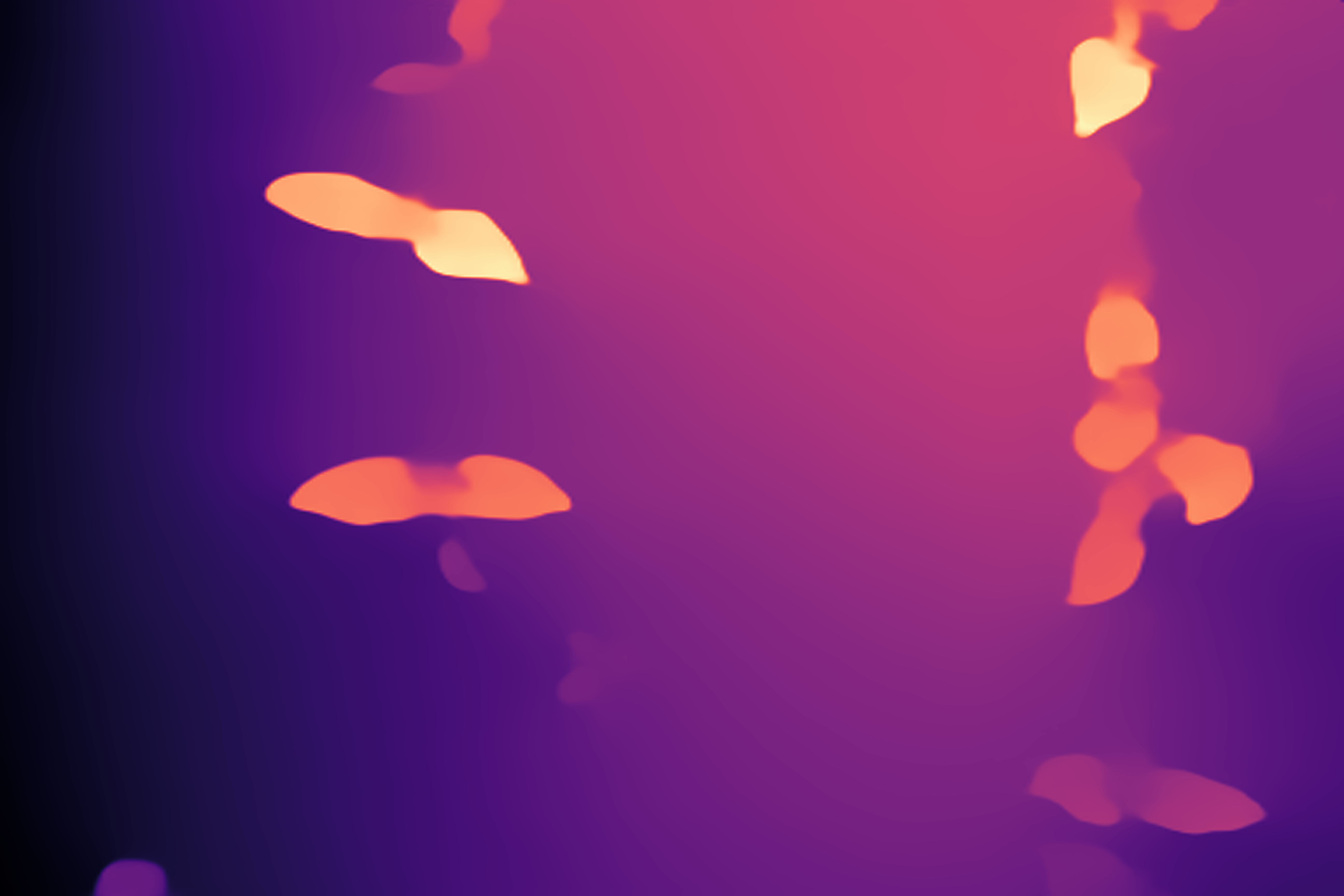}&
\includegraphics[width=.15\textwidth]{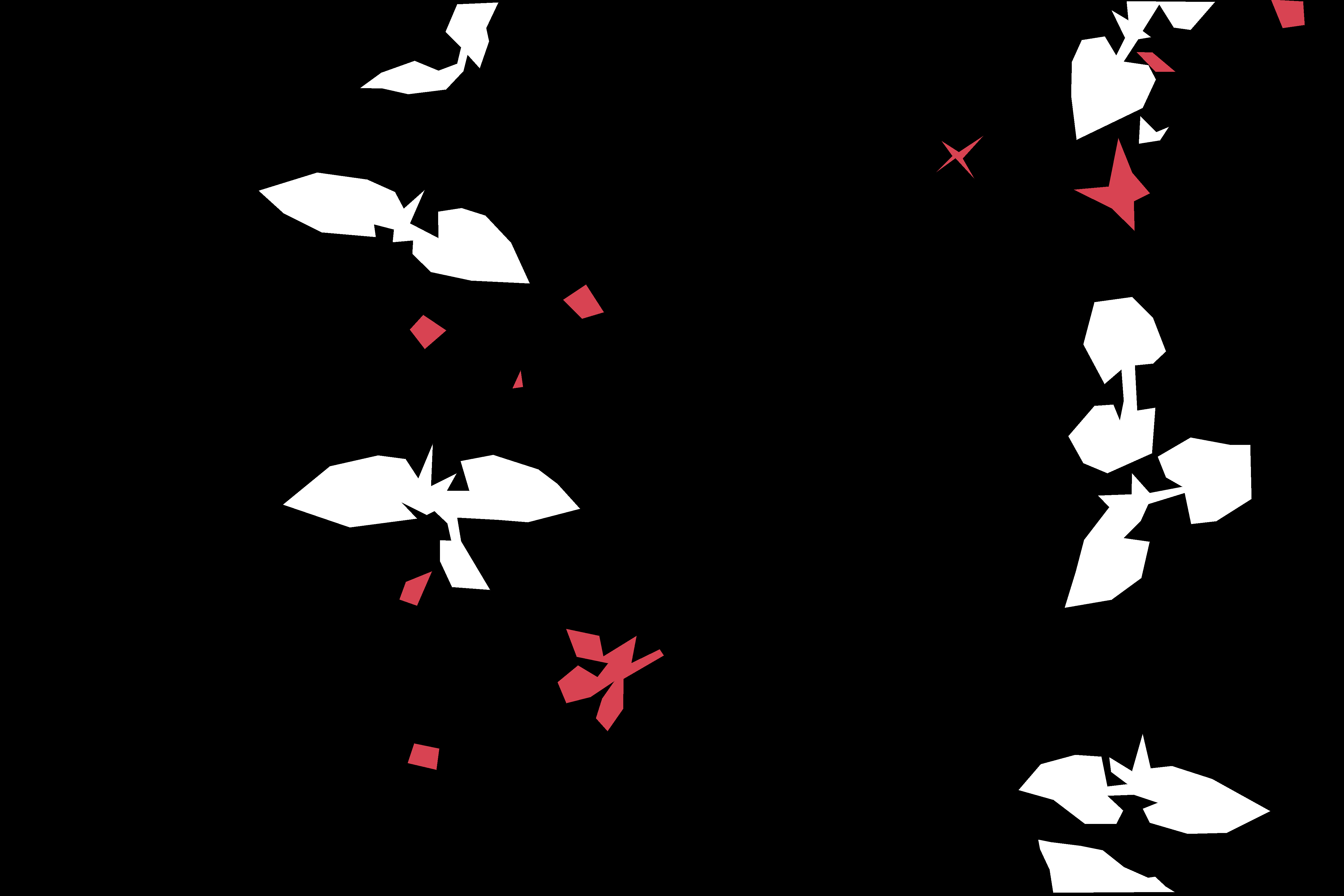}&
\includegraphics[width=.15\textwidth]{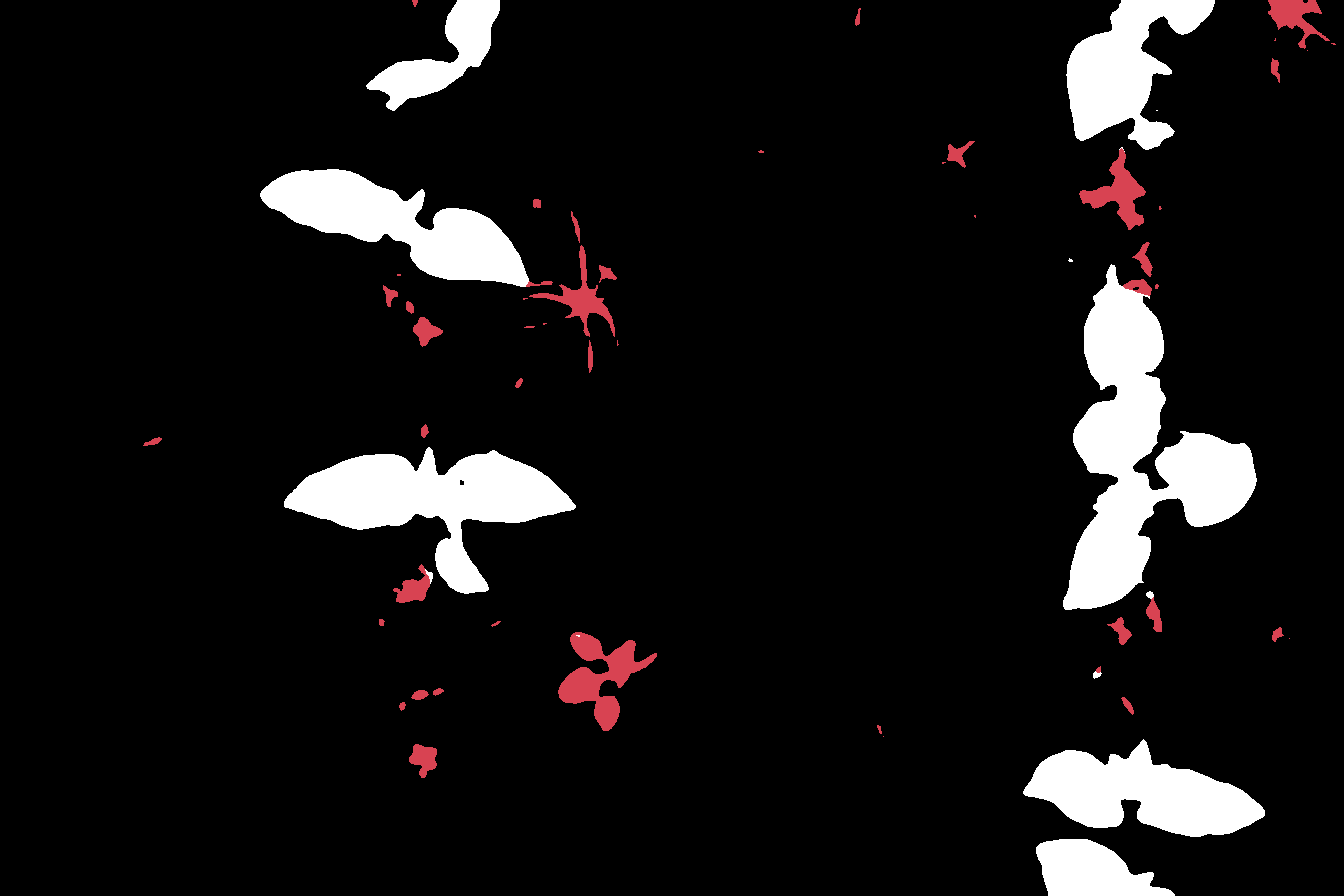}&
\includegraphics[width=.15\textwidth]{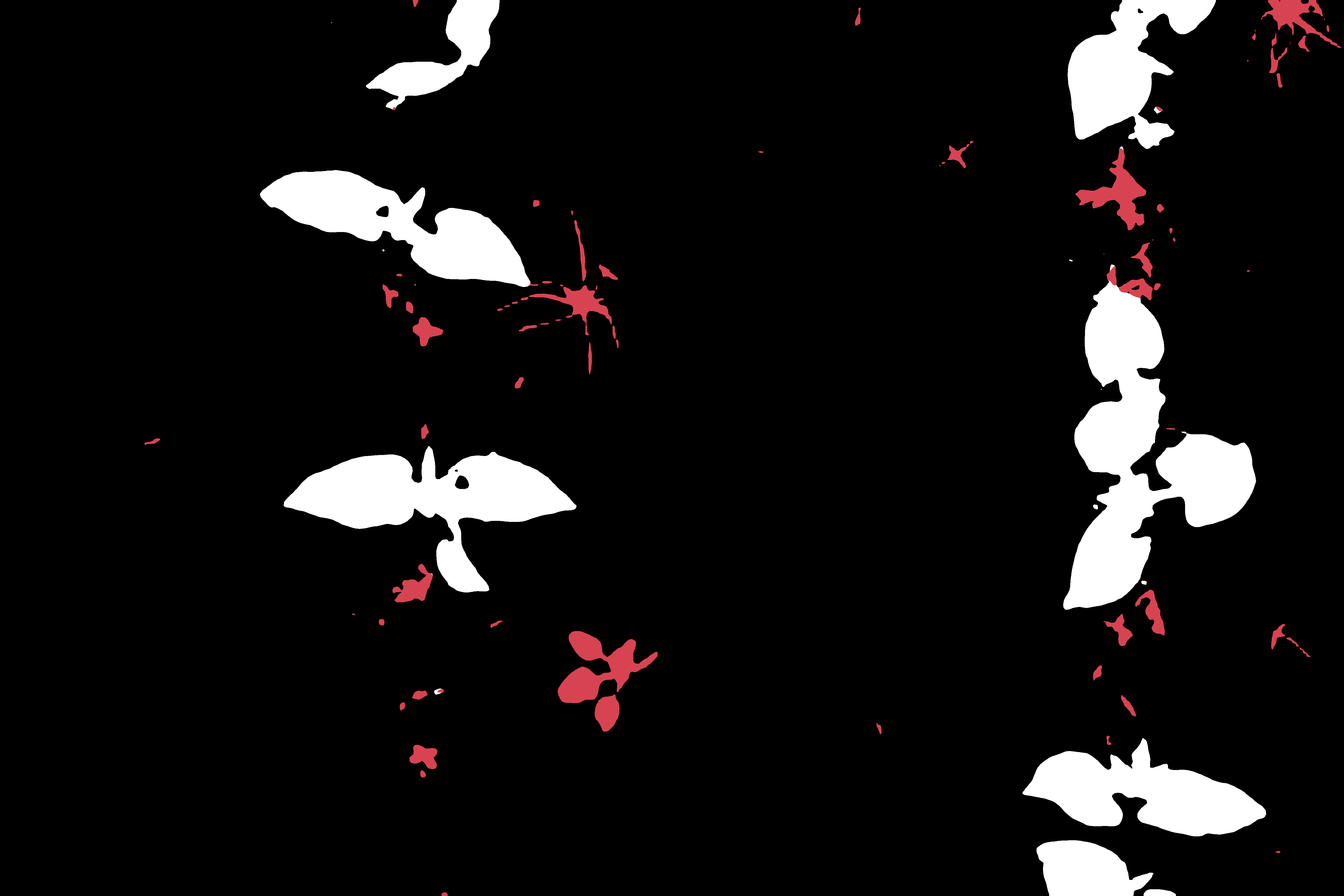}&
\includegraphics[width=.15\textwidth]{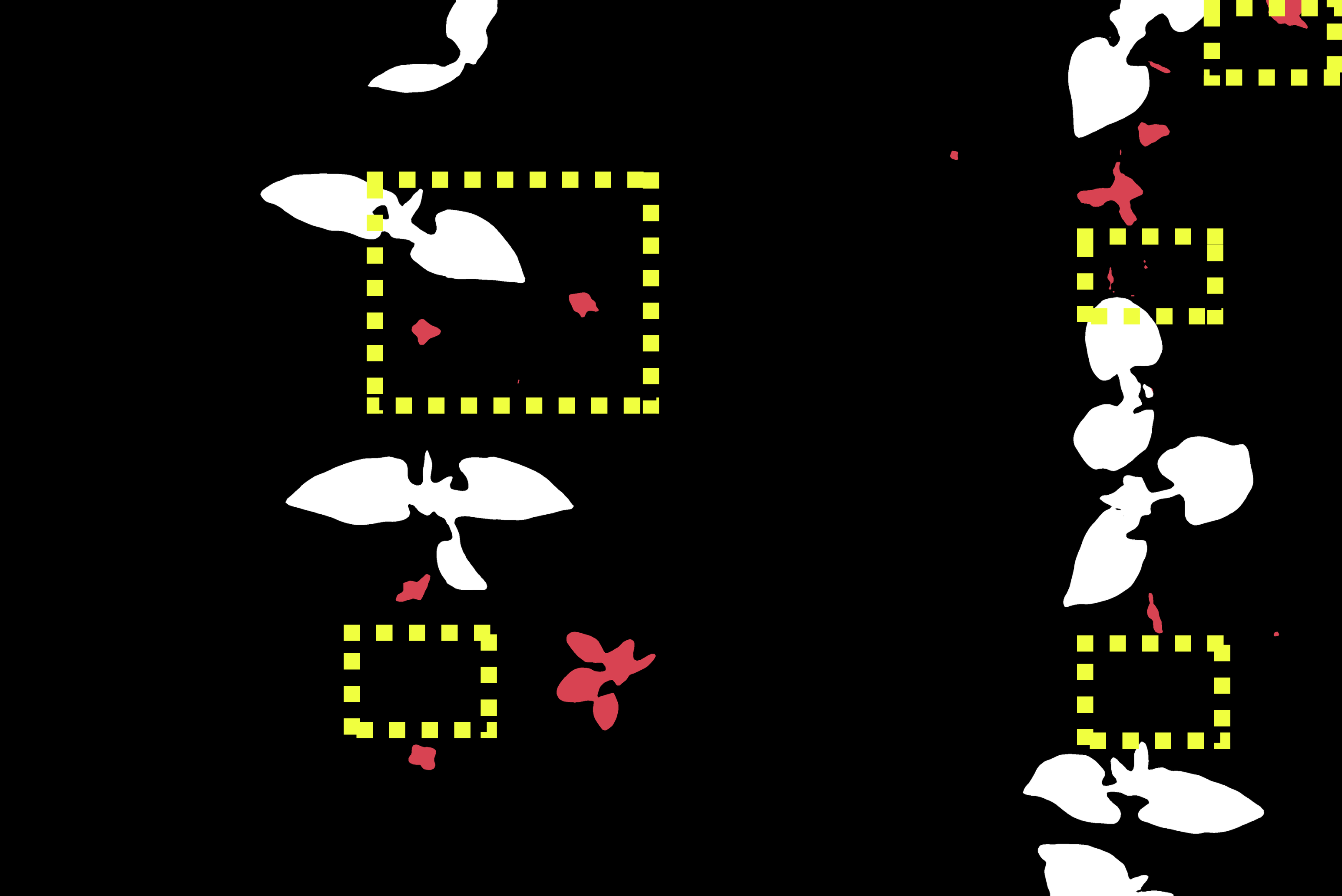}\\
 \includegraphics[width=.15\textwidth]{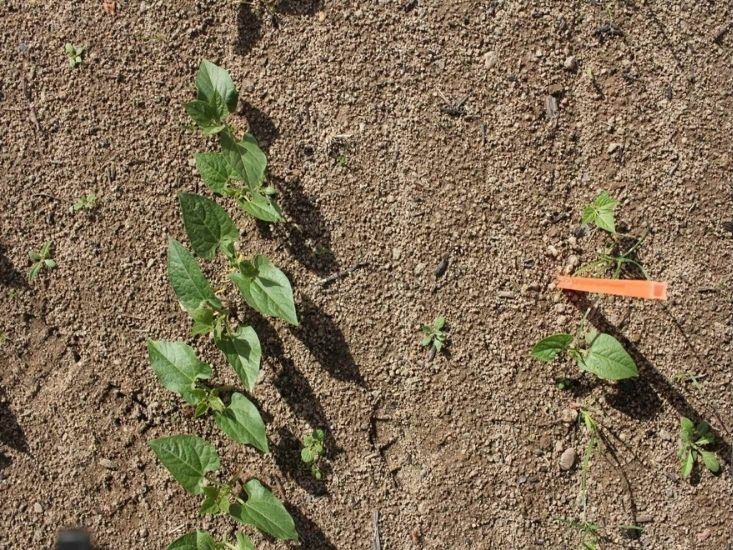}&\includegraphics[width=.15\textwidth]{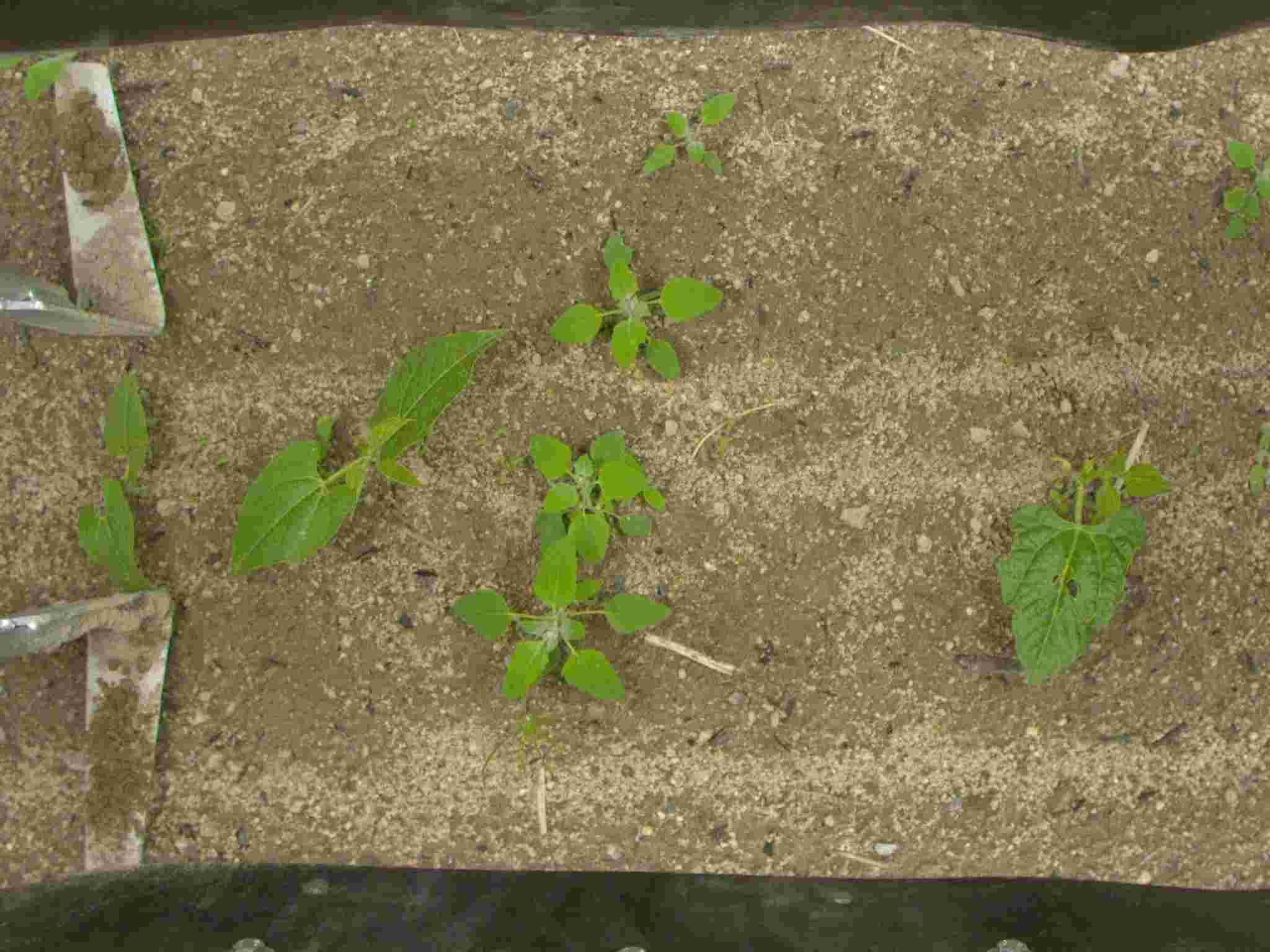} &\includegraphics[width=.15\textwidth]{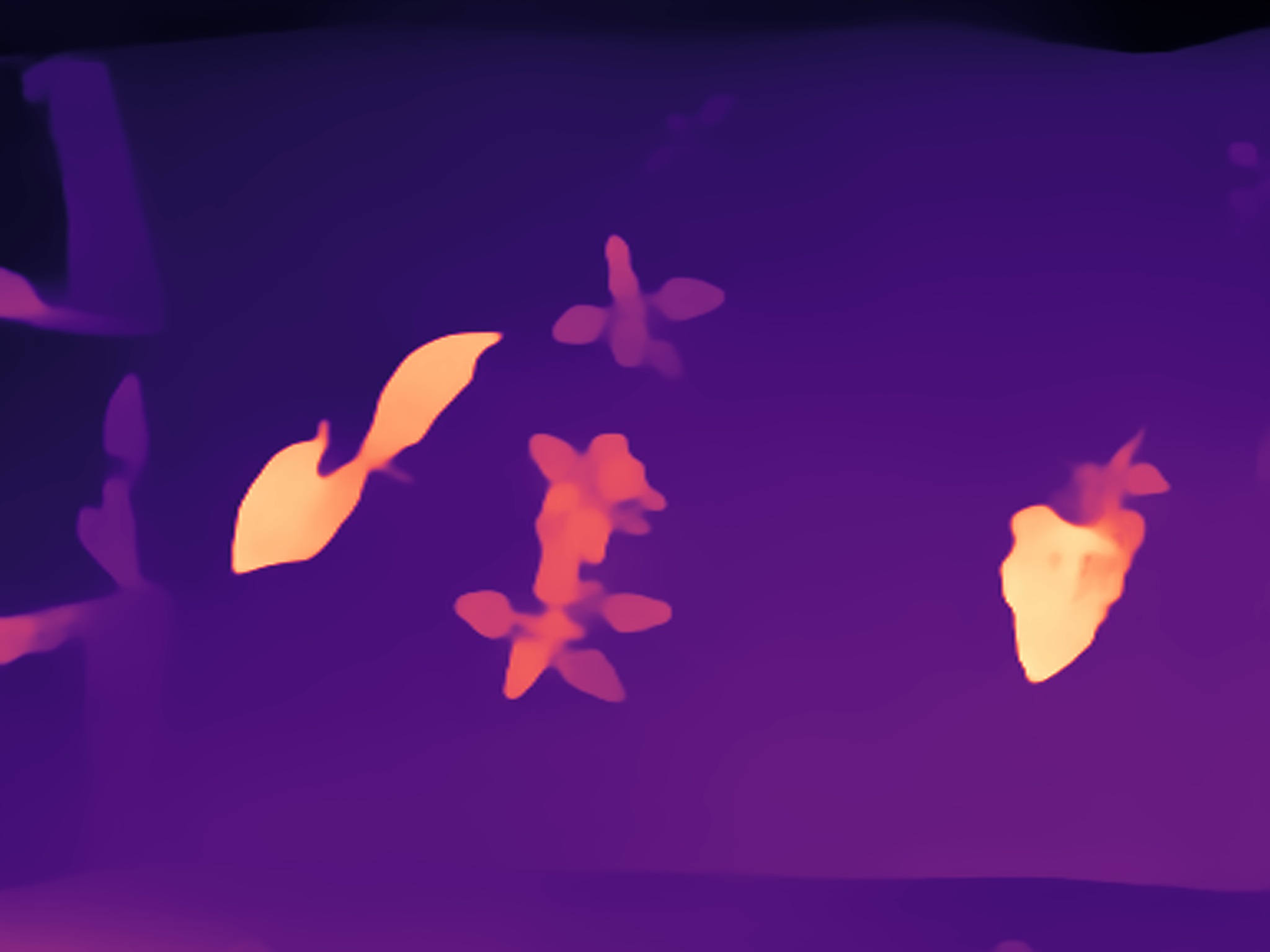}& \includegraphics[width=.15\textwidth]{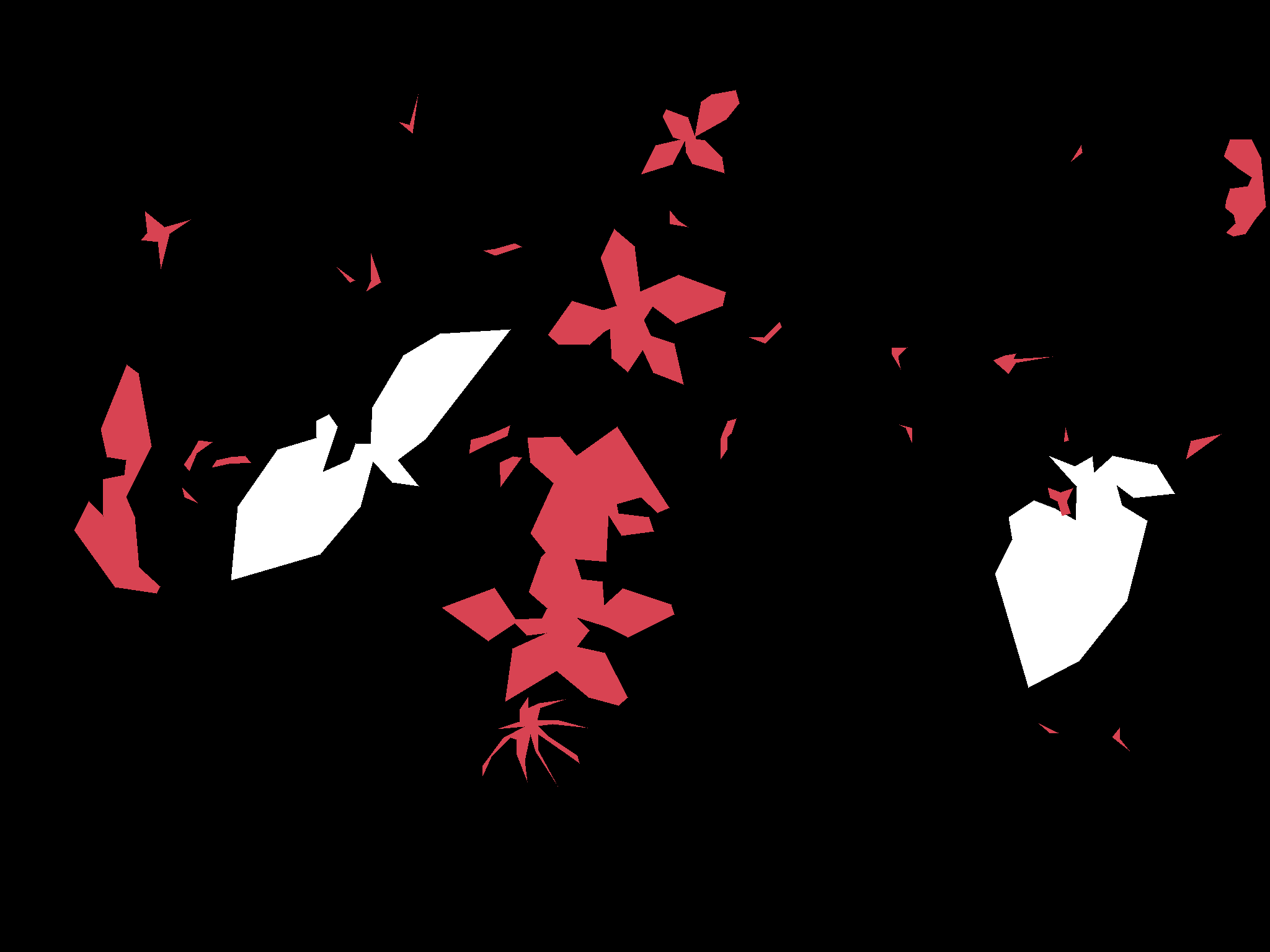}& \includegraphics[width=.15\textwidth]{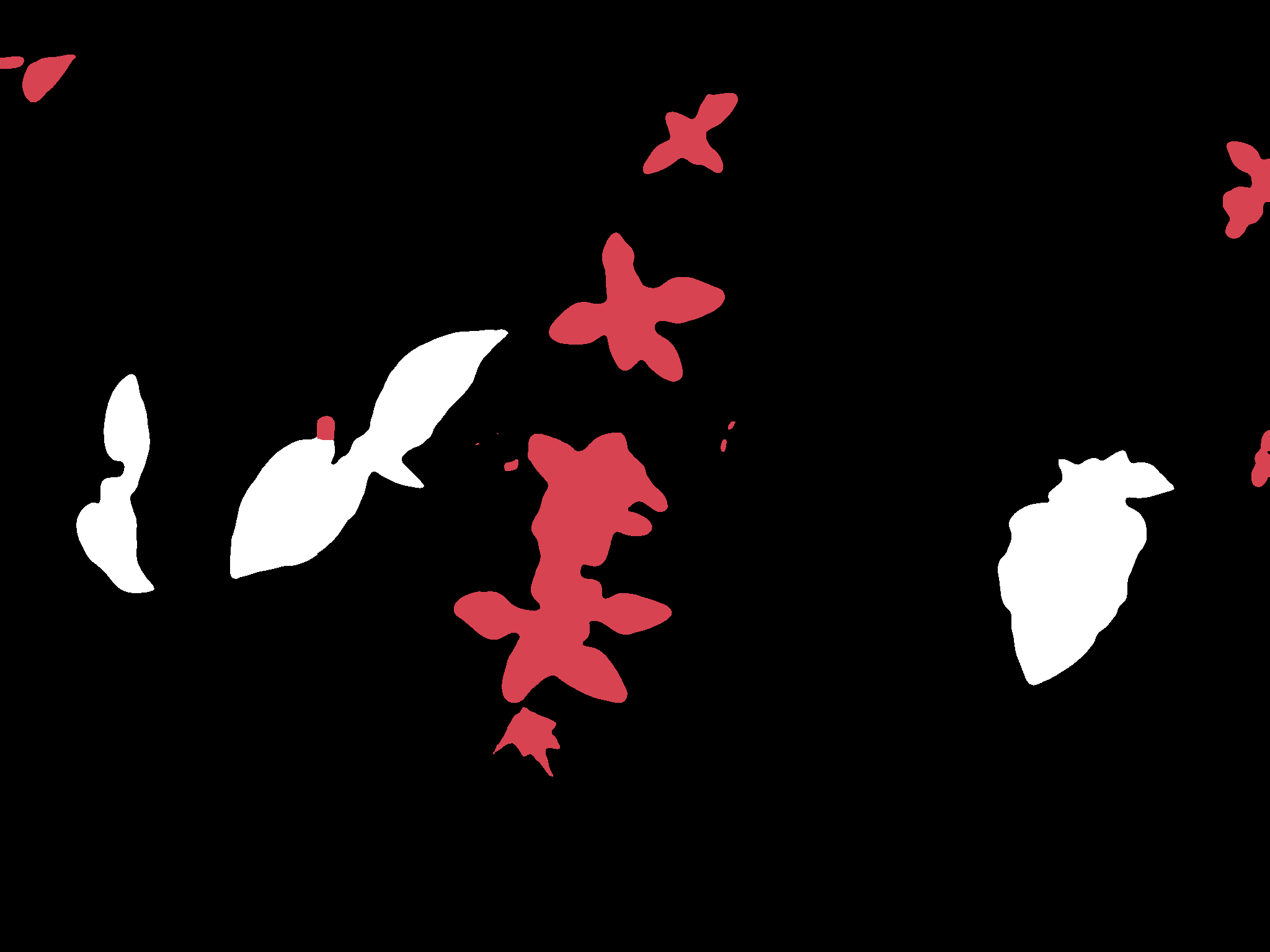}& \includegraphics[width=.15\textwidth]{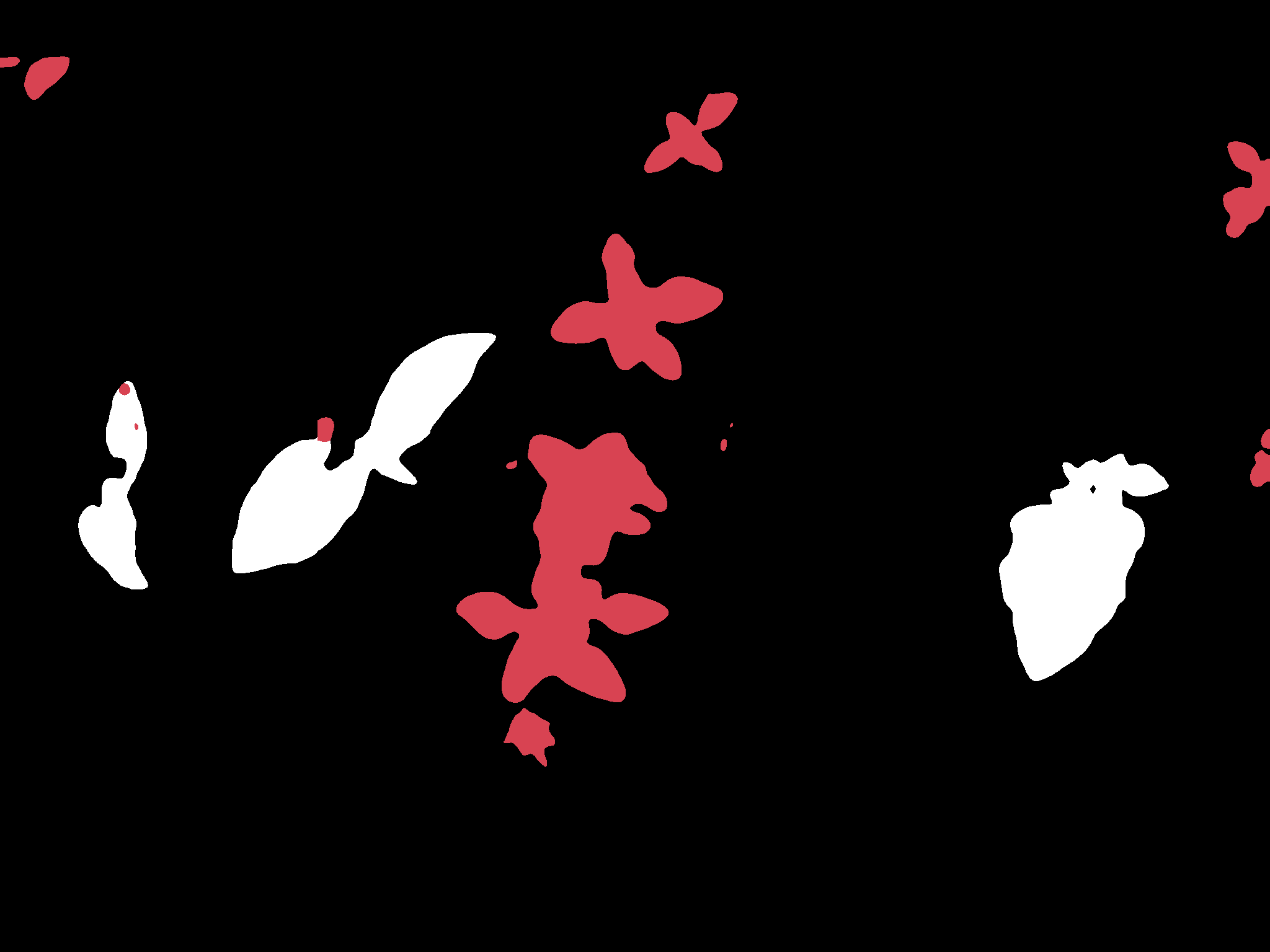}&\includegraphics[width=.15\textwidth]{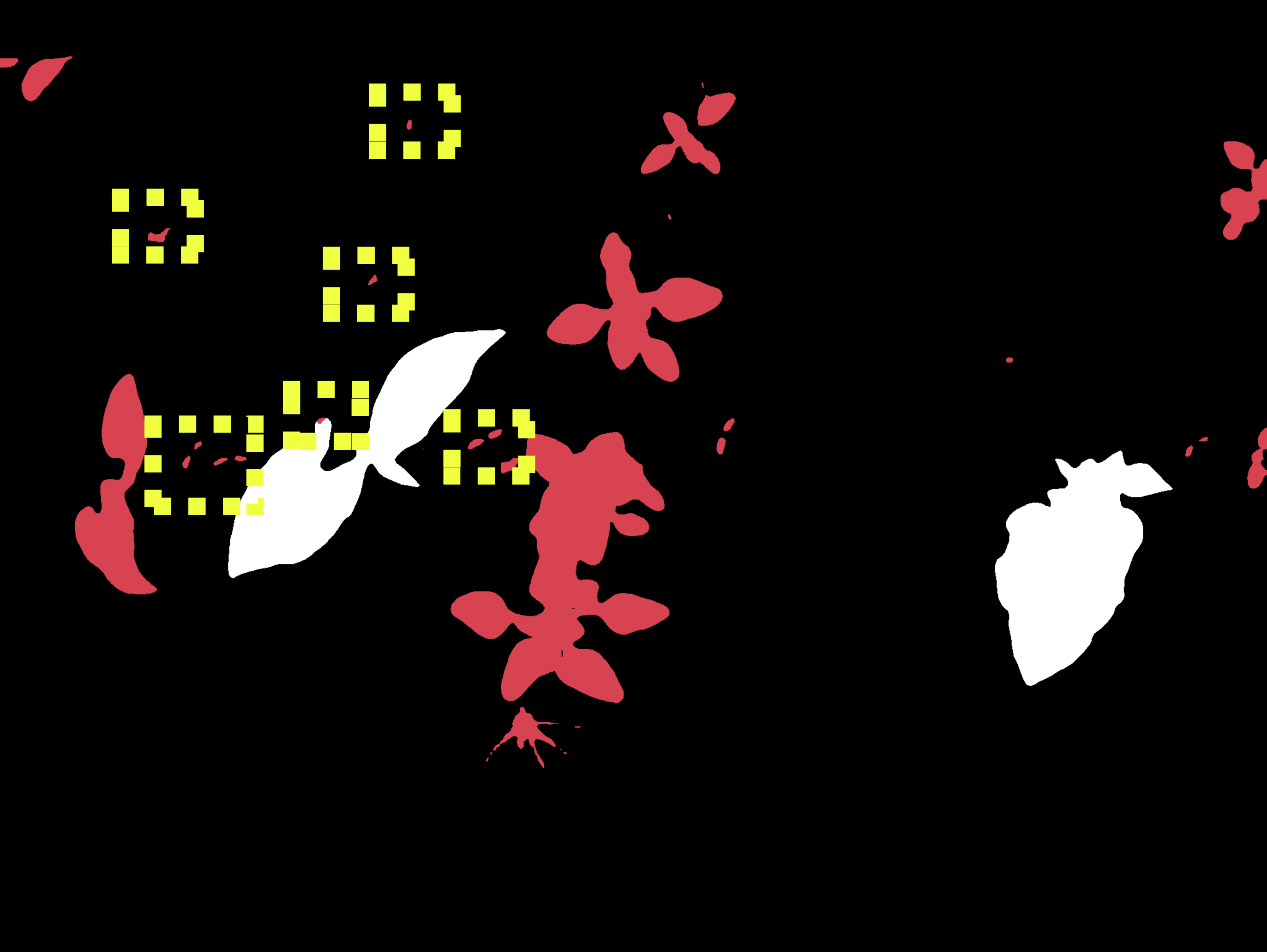}\\
 \includegraphics[width=.15\textwidth]{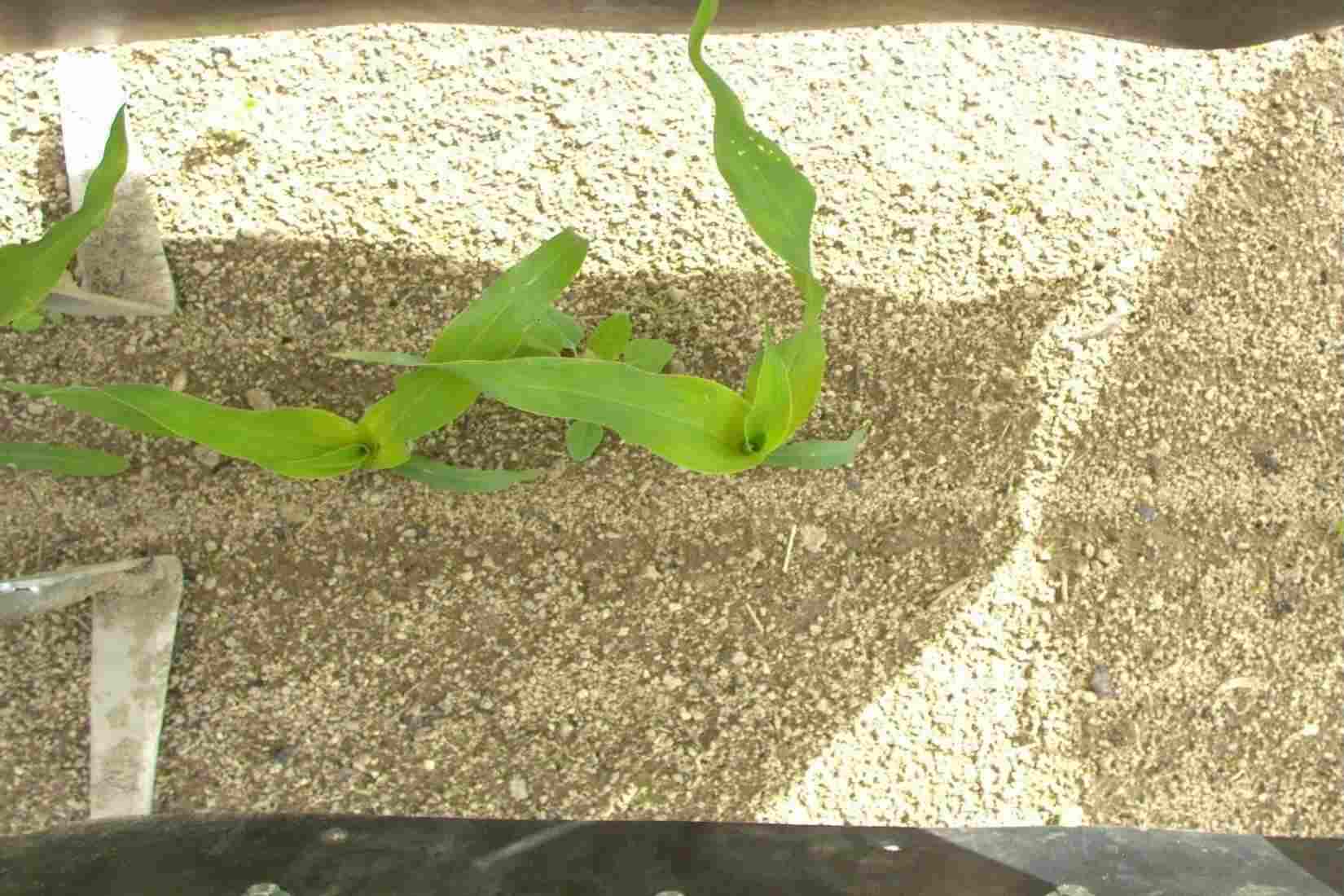}&\includegraphics[width=.15\textwidth]{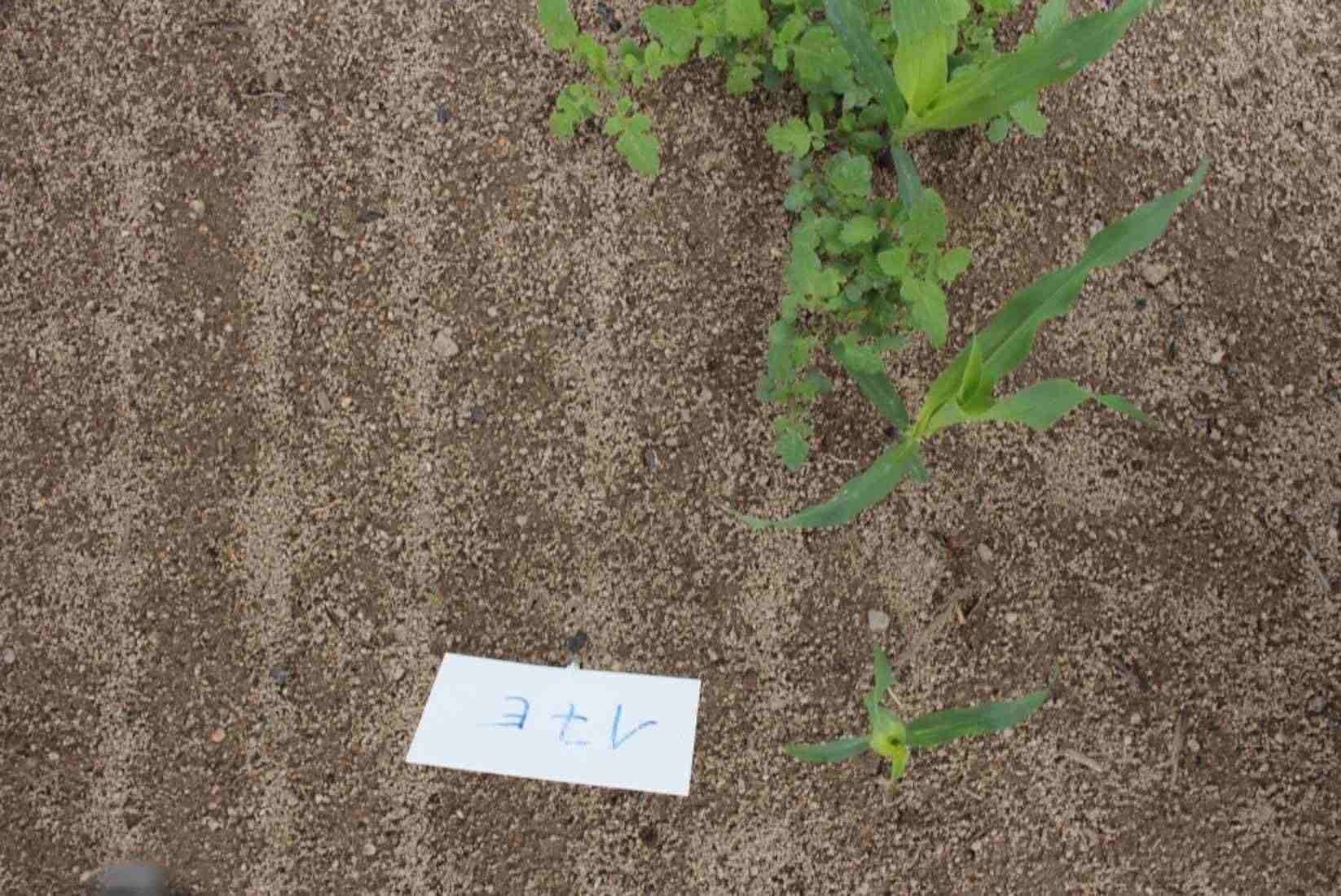}&\includegraphics[width=.15\textwidth]{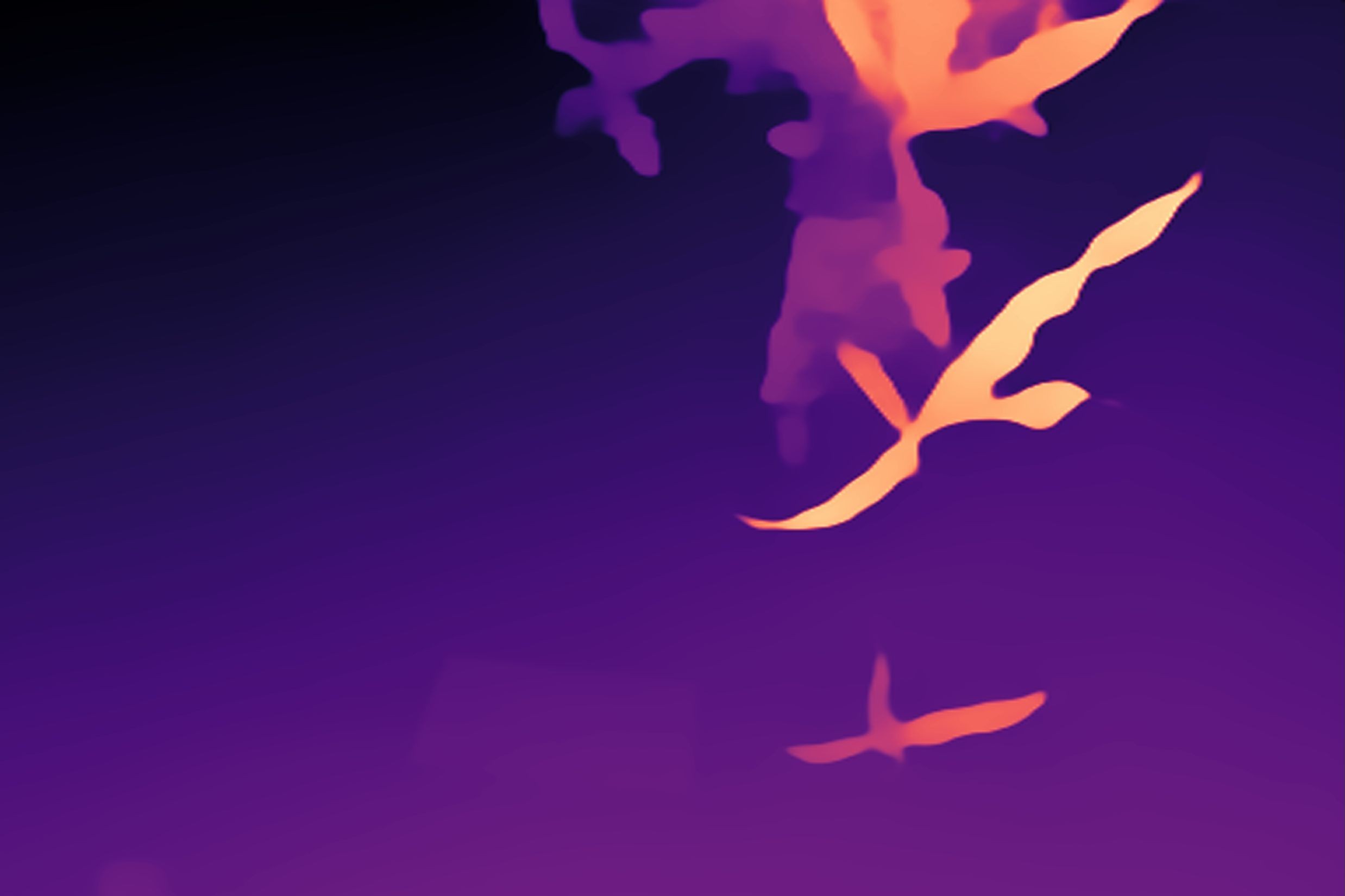}& \includegraphics[width=.15\textwidth]{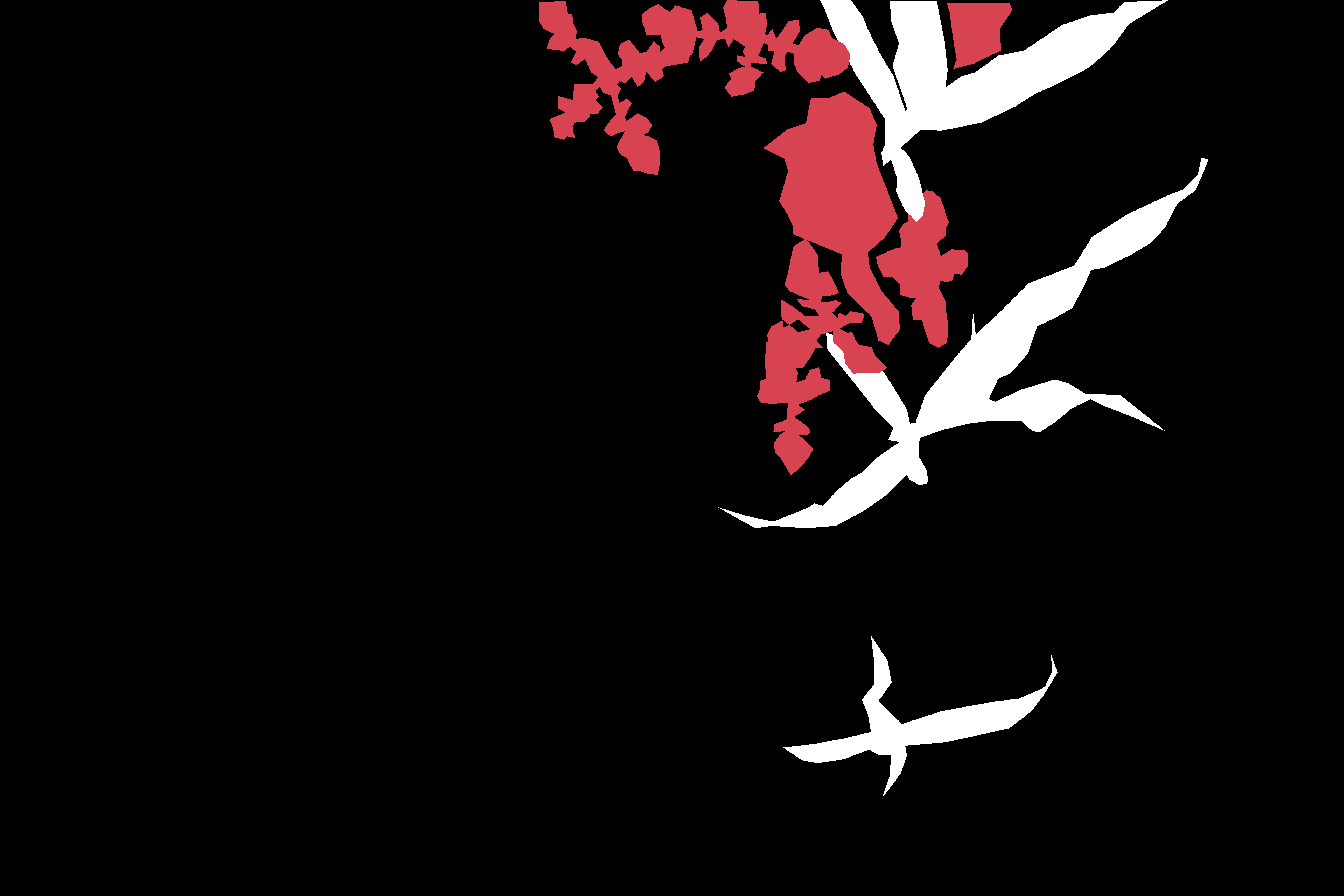}& \includegraphics[width=.15\textwidth]{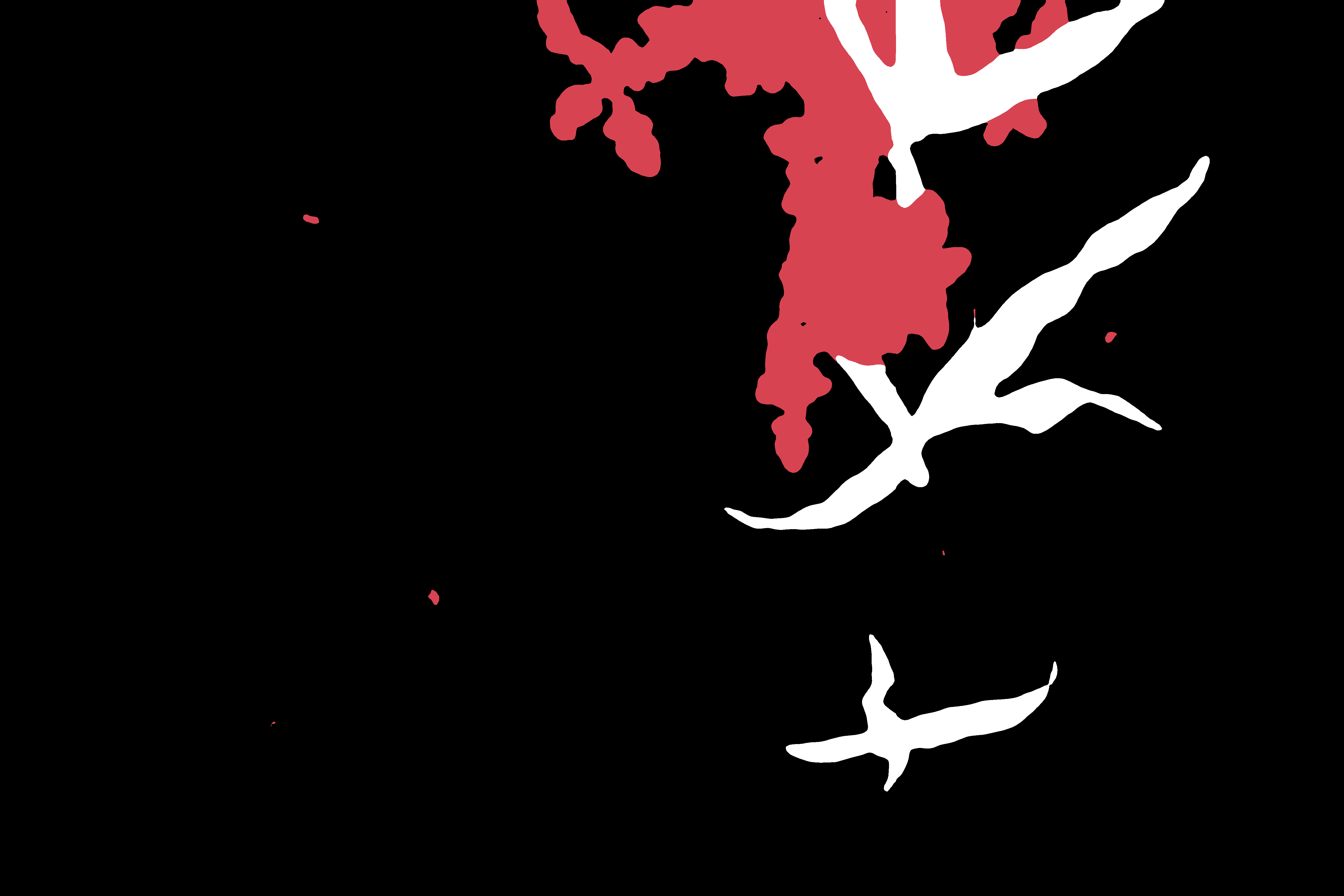}& \includegraphics[width=.15\textwidth]{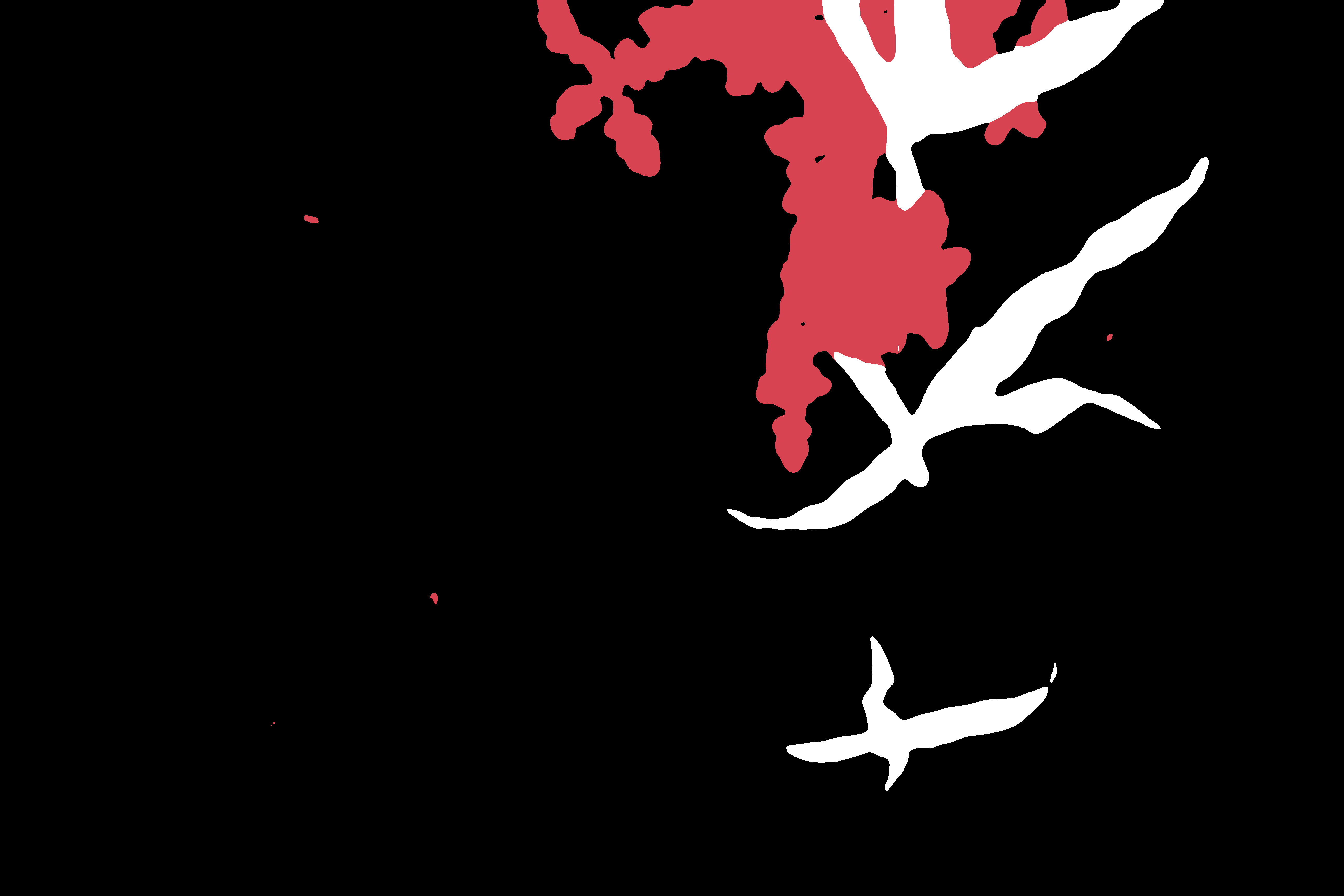}&\includegraphics[width=.15\textwidth]{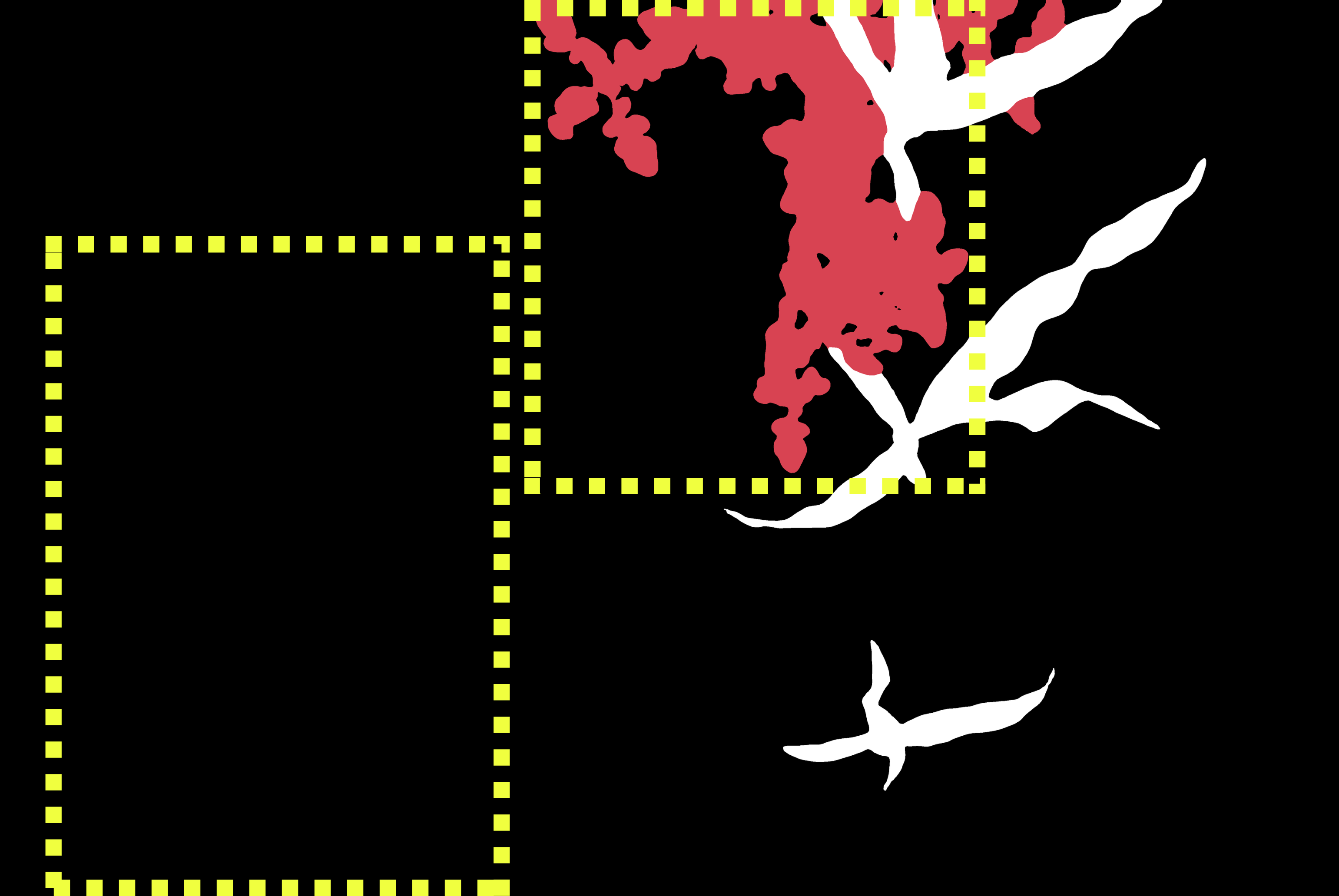}\\
 \includegraphics[width=.15\textwidth]{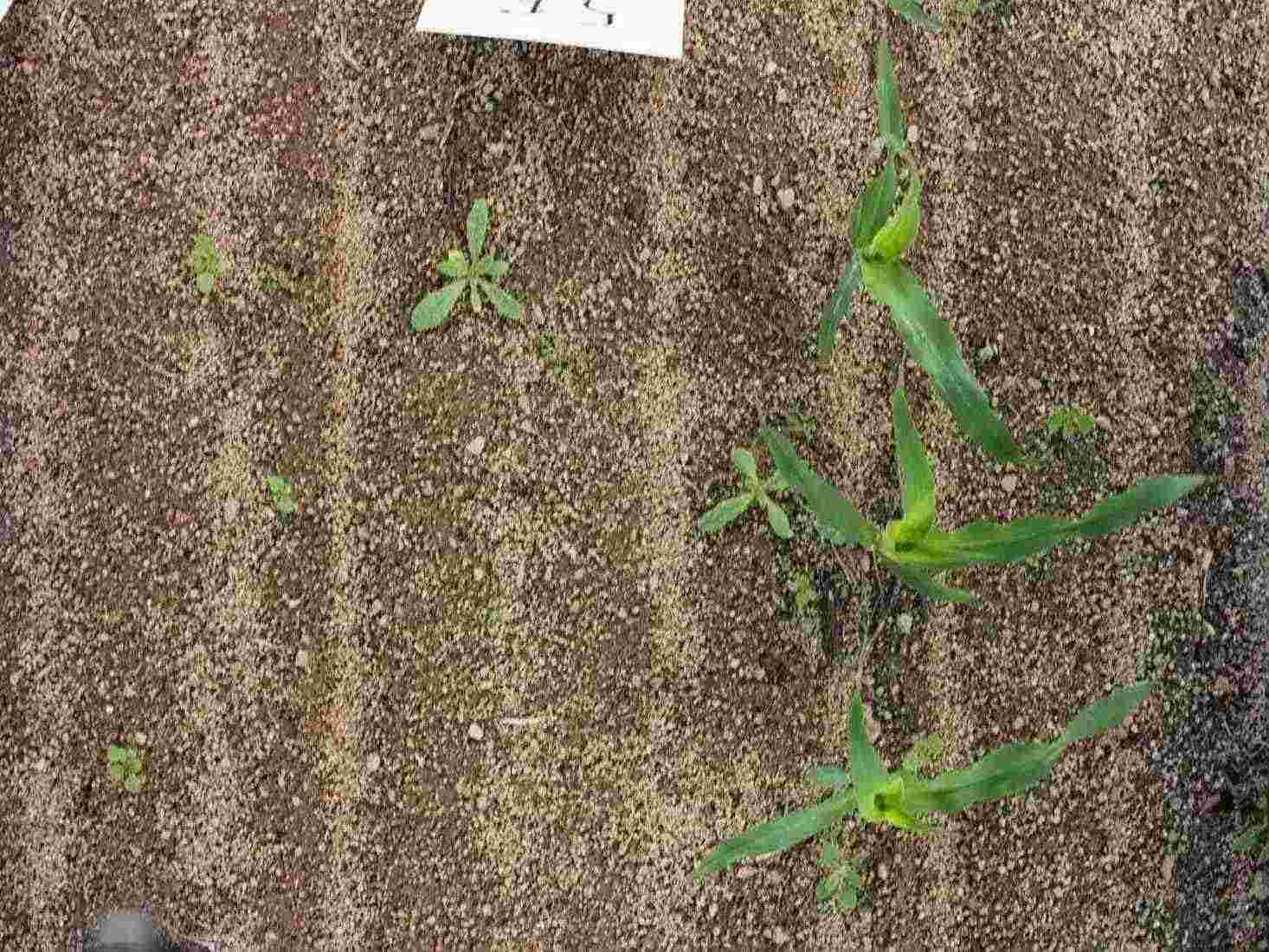}&\includegraphics[width=.15\textwidth]{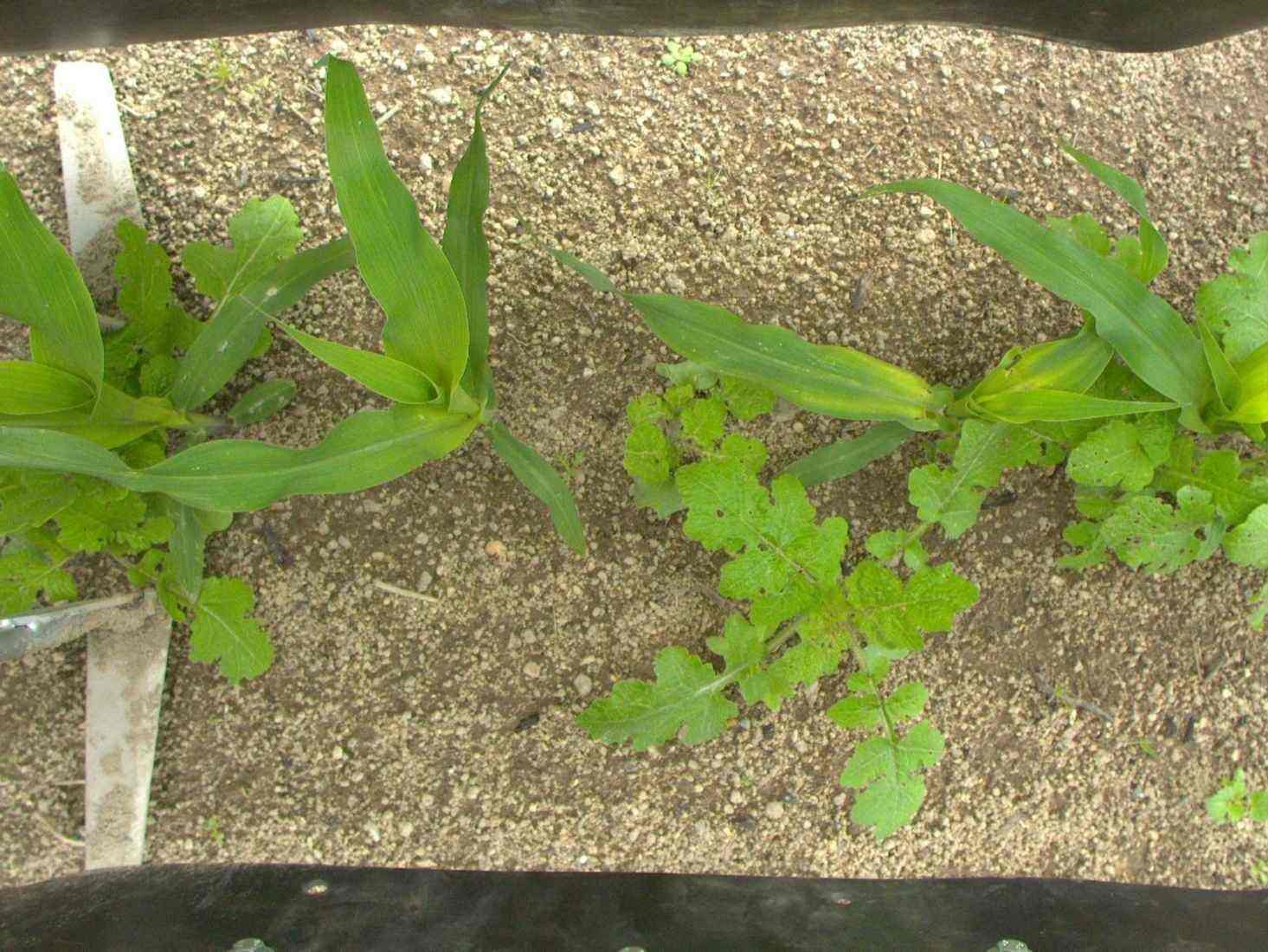} &\includegraphics[width=.15\textwidth]{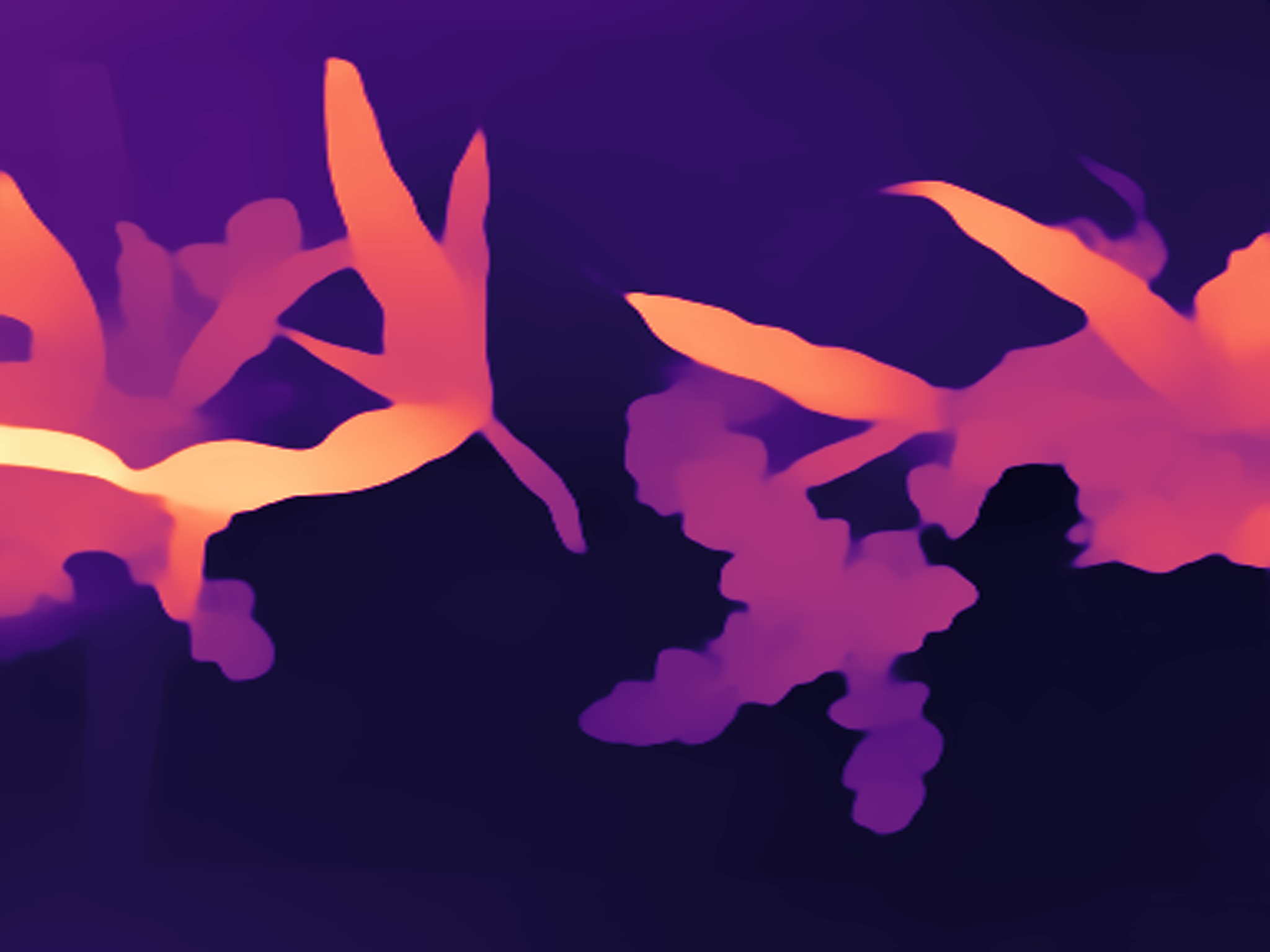}& \includegraphics[width=.15\textwidth]{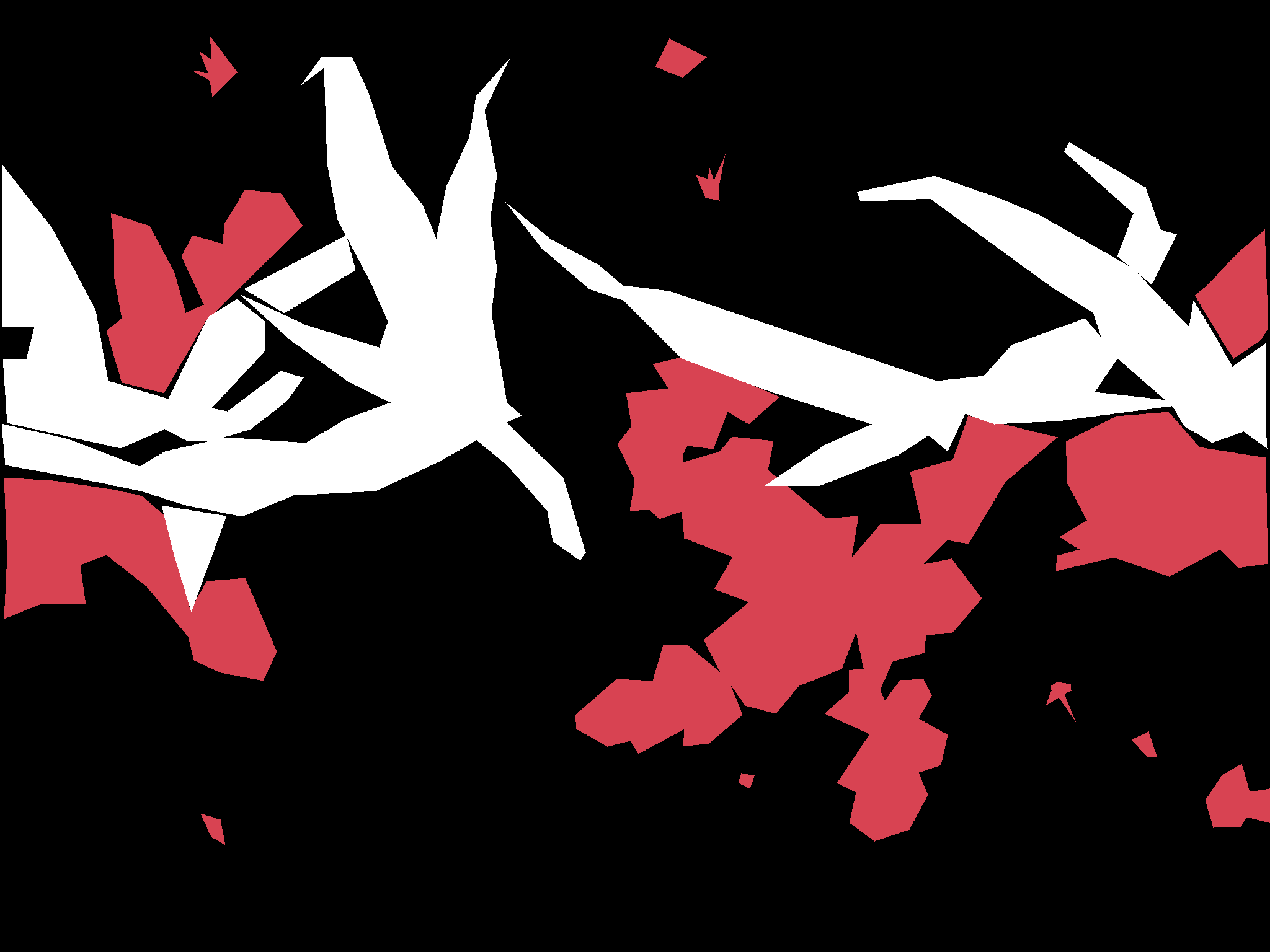}& \includegraphics[width=.15\textwidth]{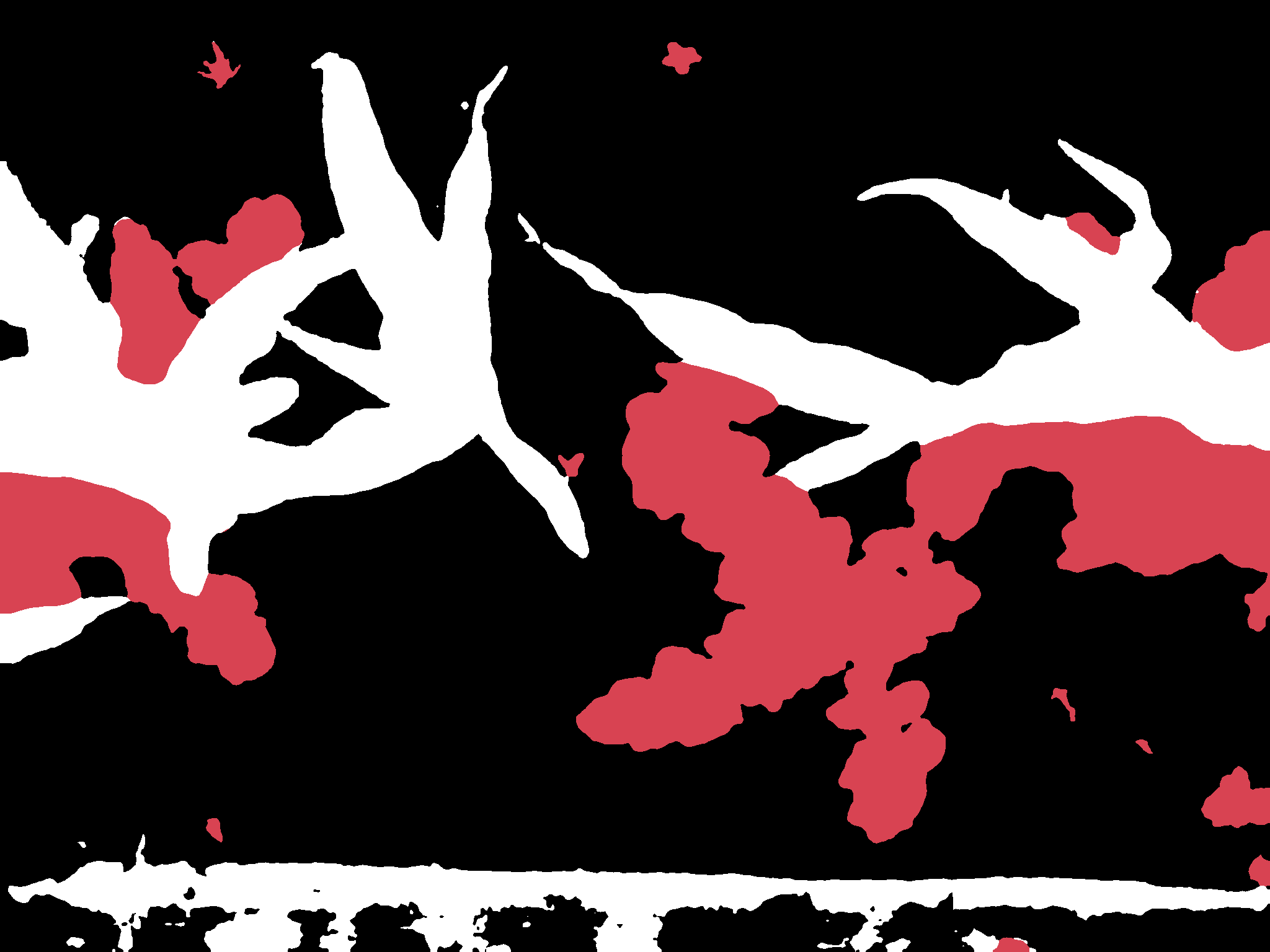}& \includegraphics[width=.15\textwidth]{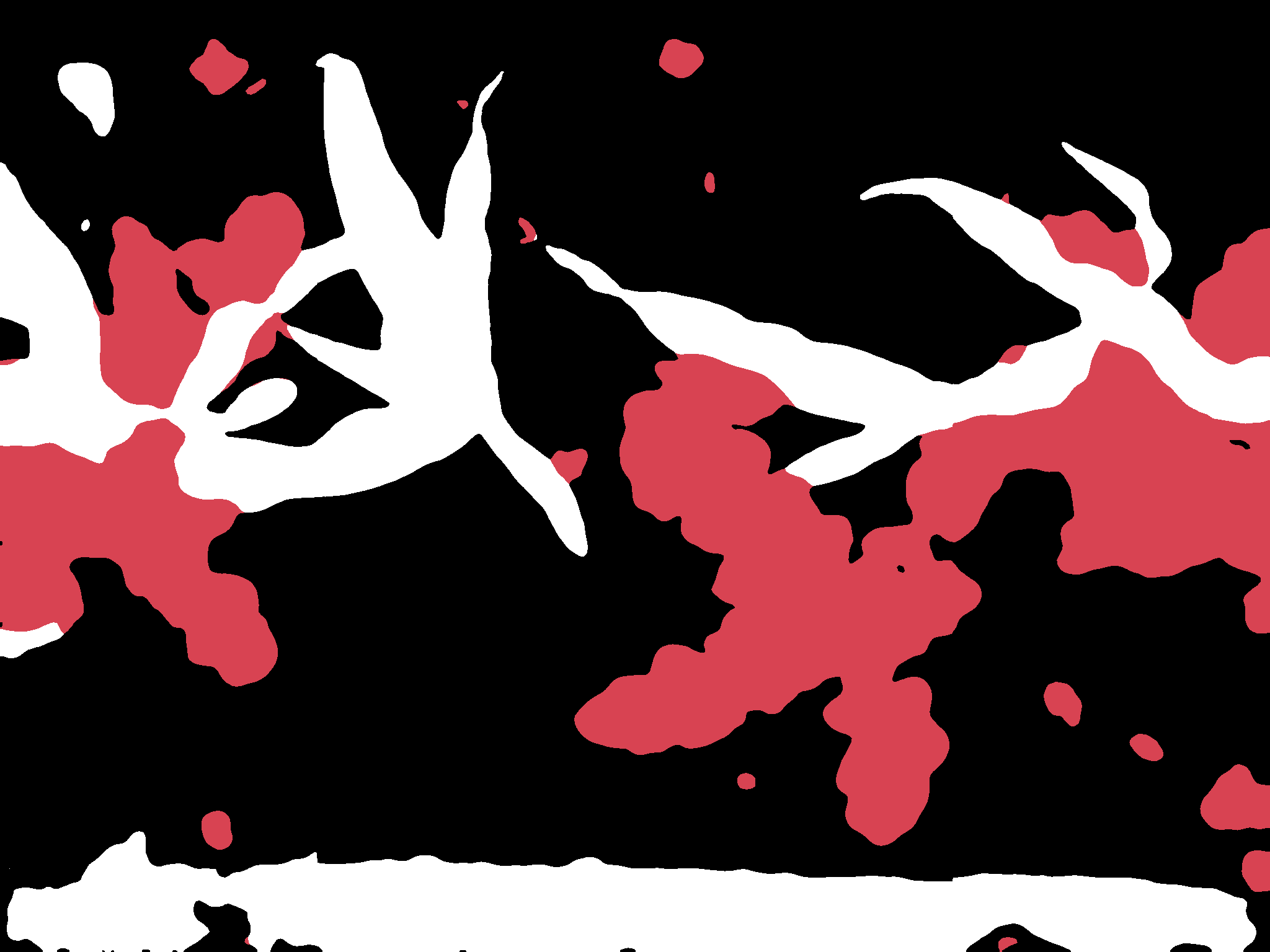}&\includegraphics[width=.15\textwidth]{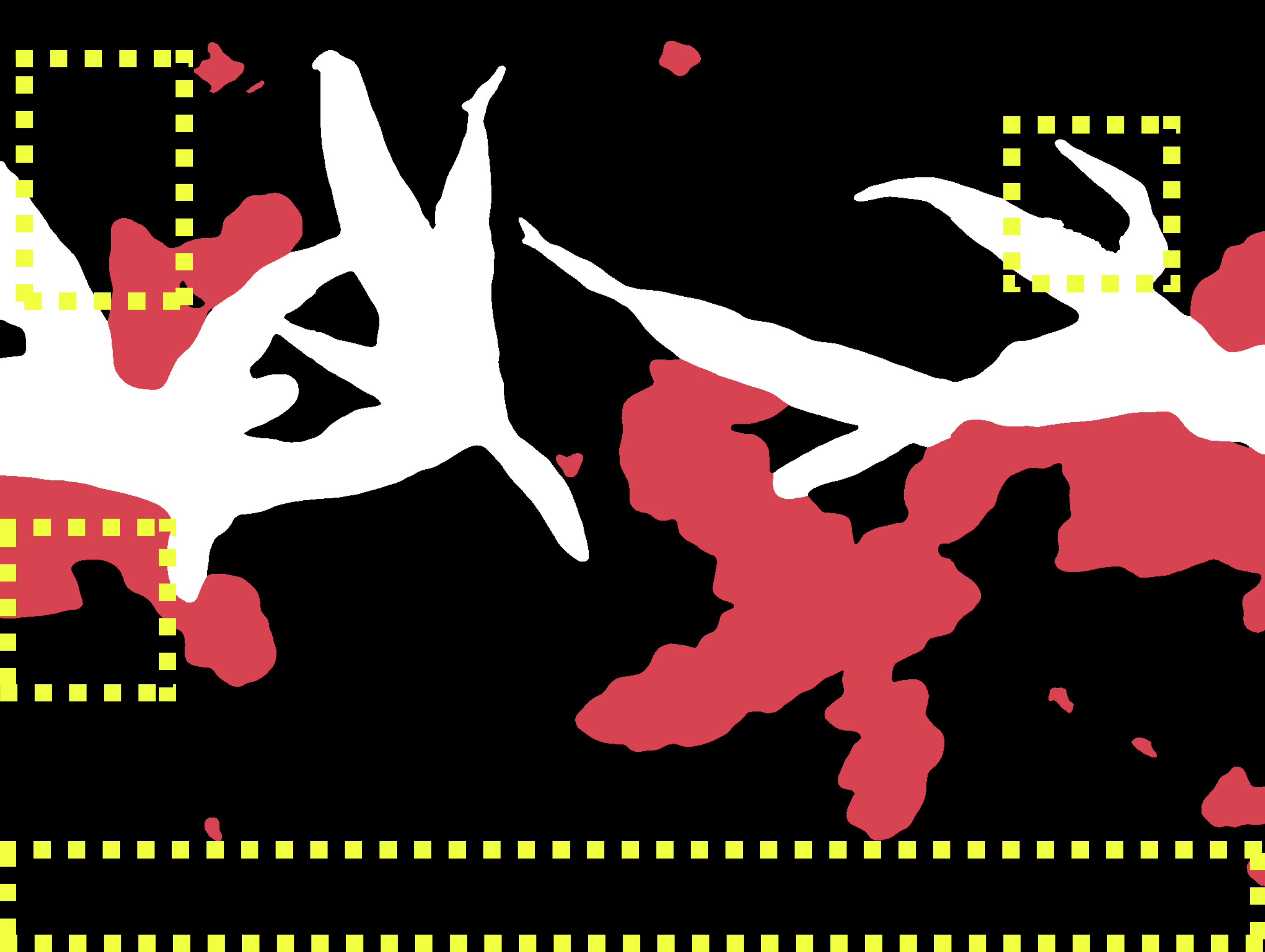}\\
 \includegraphics[width=.15\textwidth]{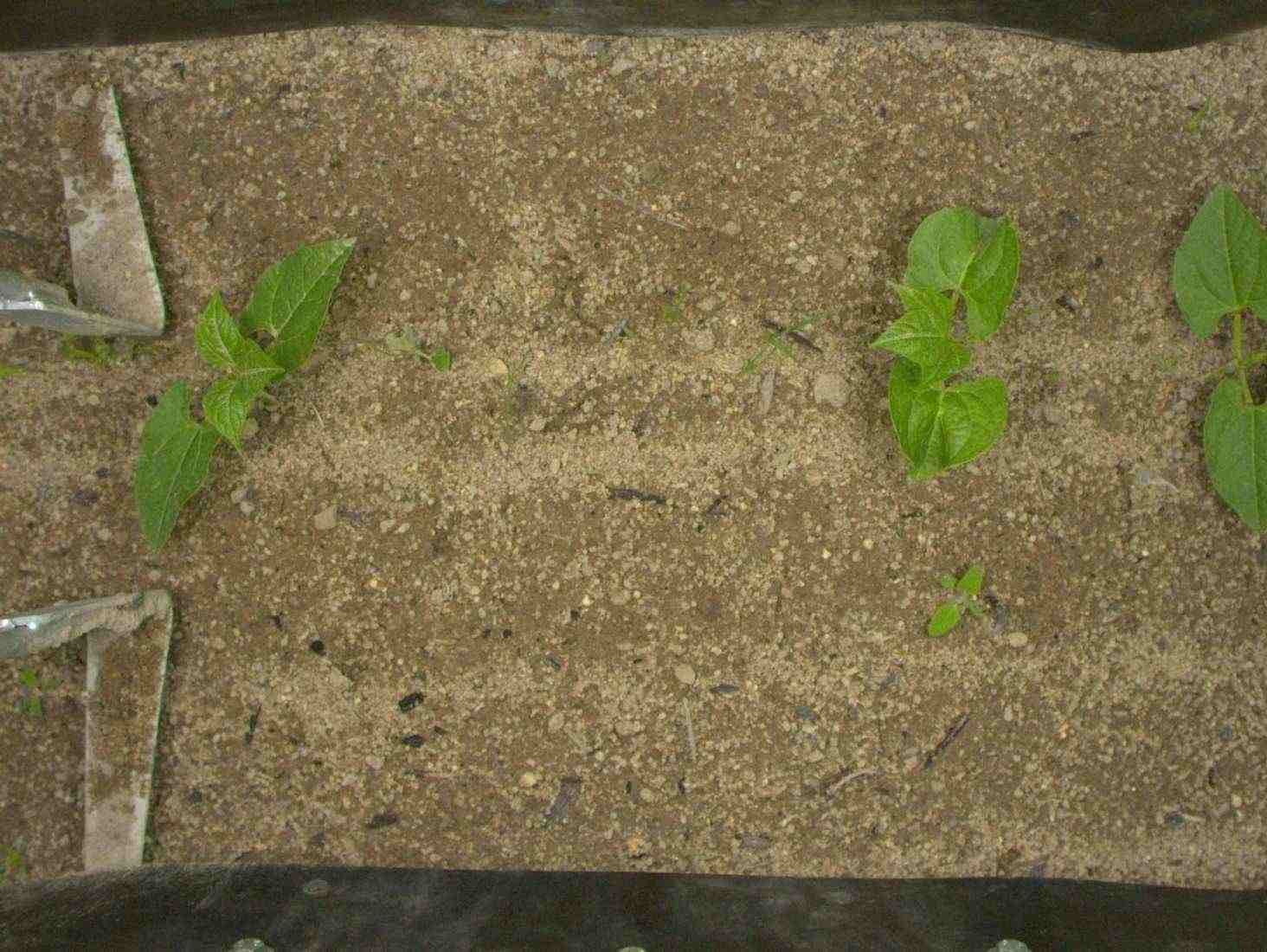}& \includegraphics[width=.15\textwidth]{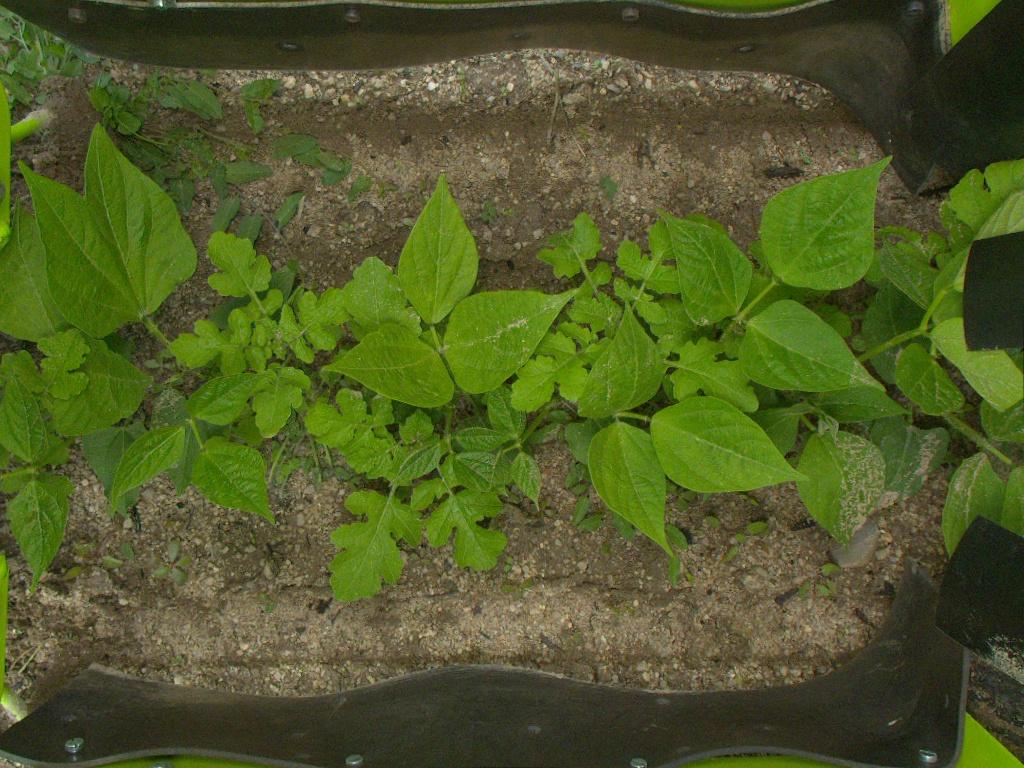} &\includegraphics[width=.15\textwidth]{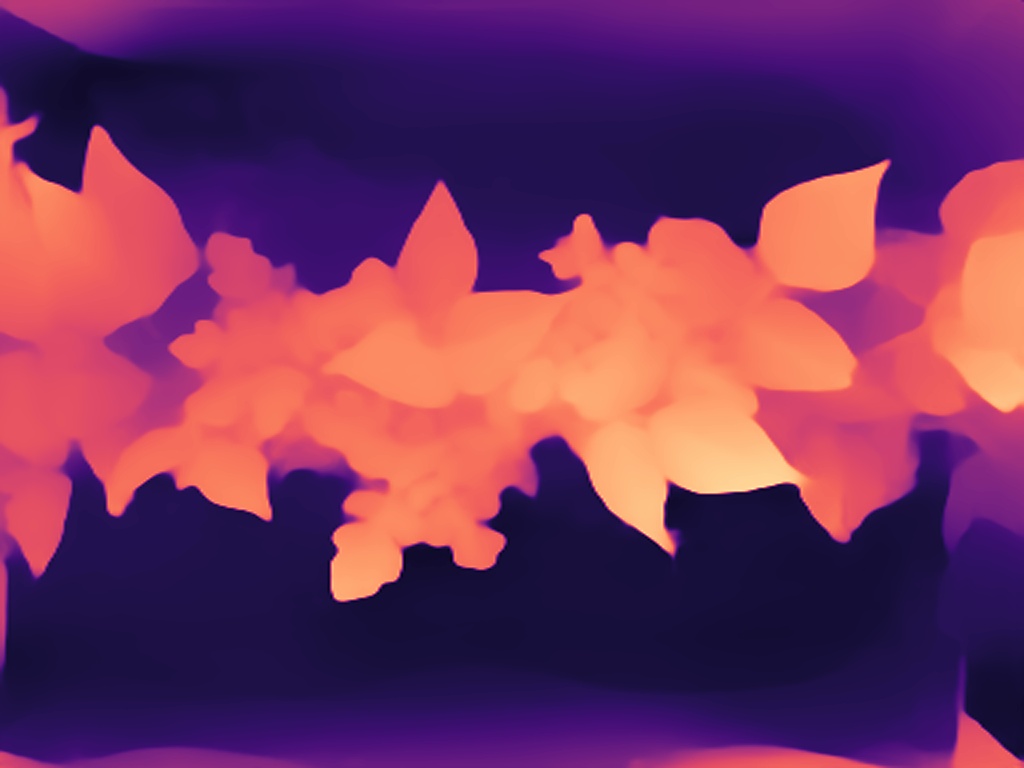}& \includegraphics[width=.15\textwidth]{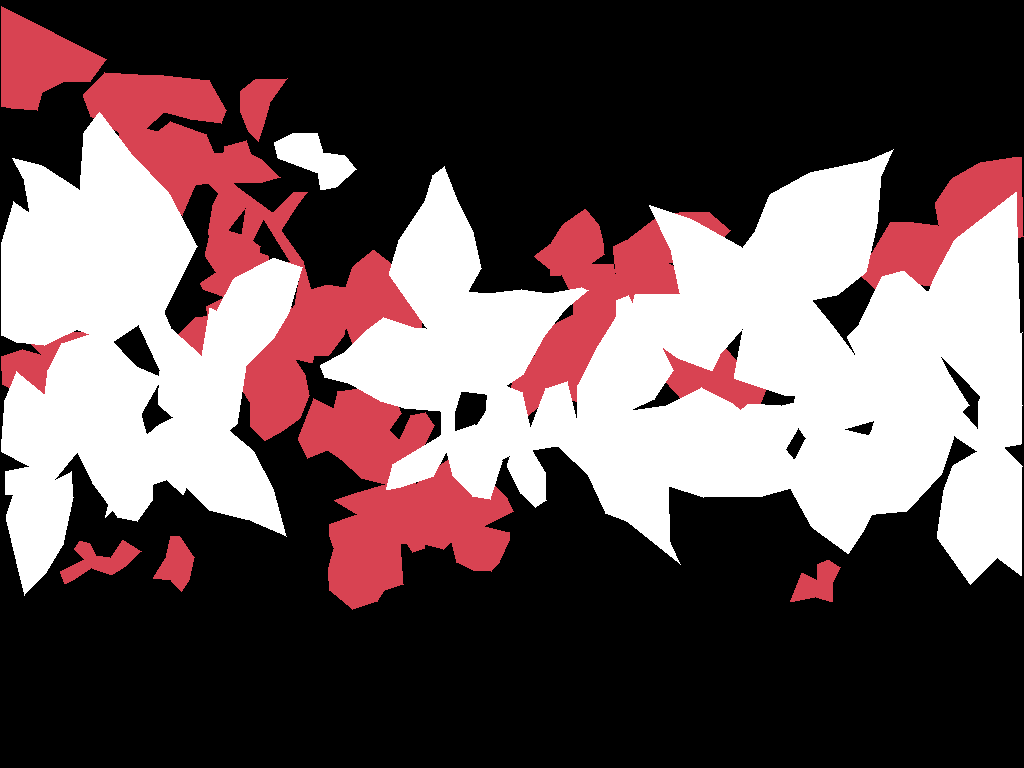}& \includegraphics[width=.15\textwidth]{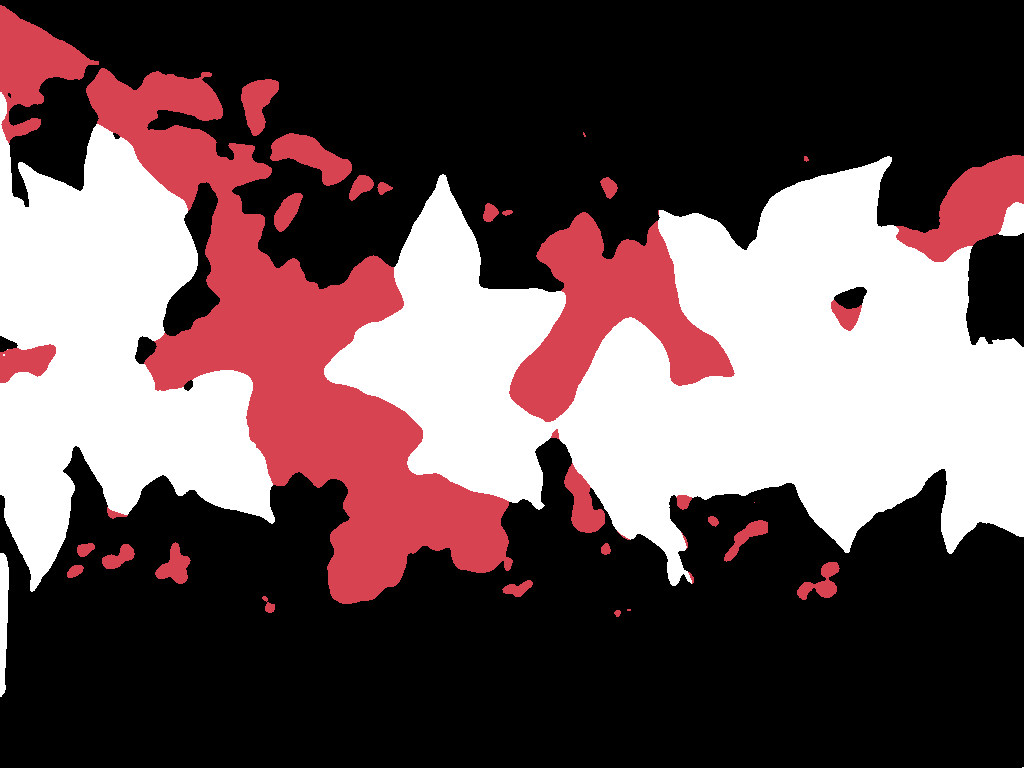}& \includegraphics[width=.15\textwidth]{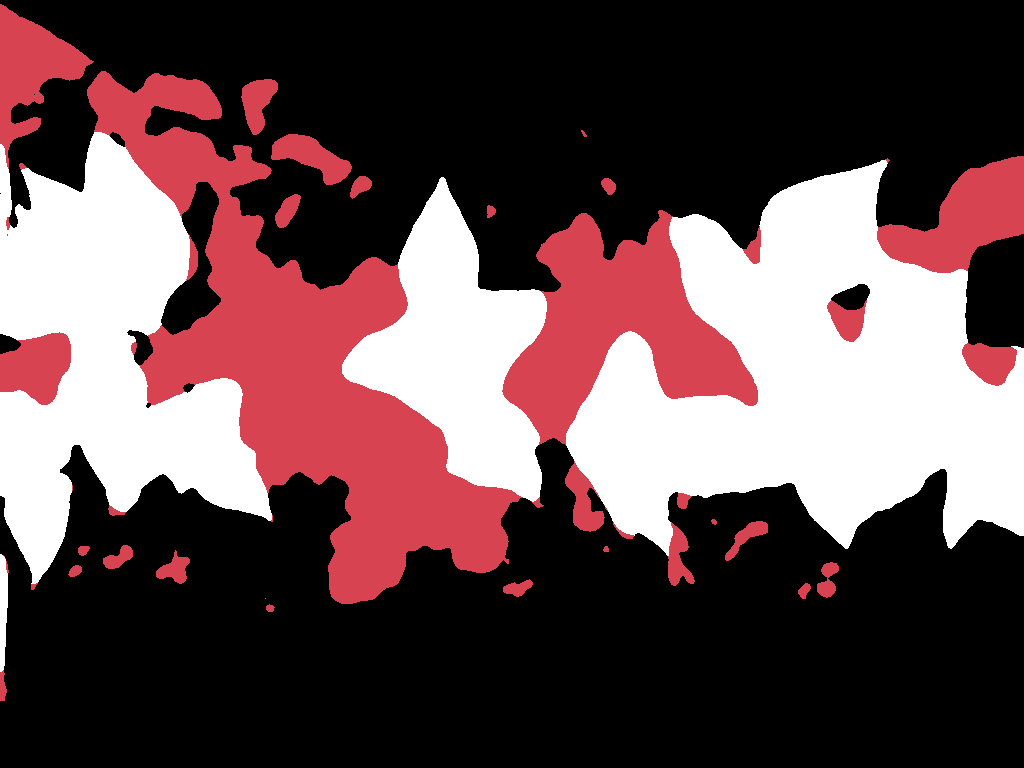}&\includegraphics[width=.15\textwidth]{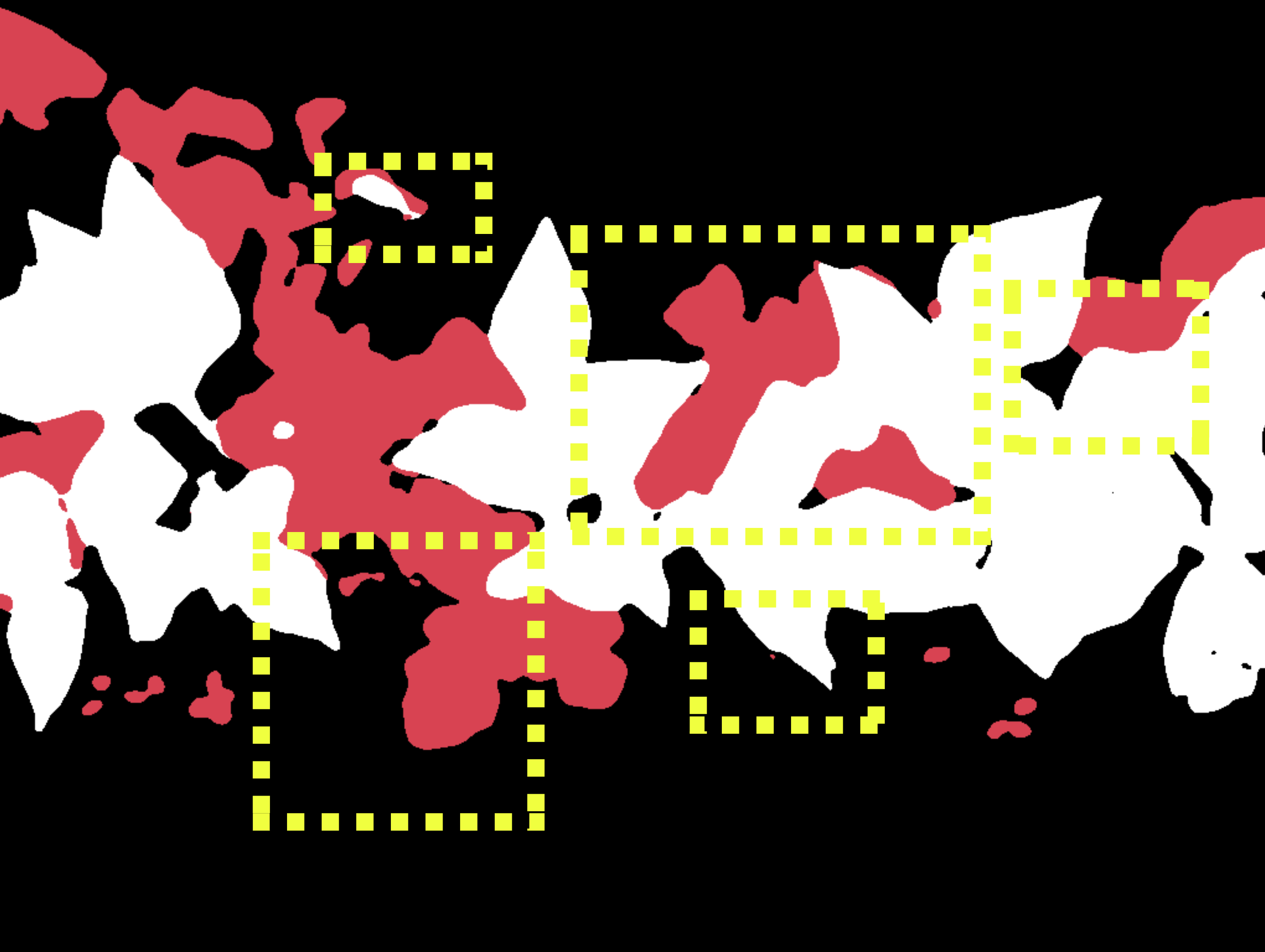}\\
 \includegraphics[width=.15\textwidth]{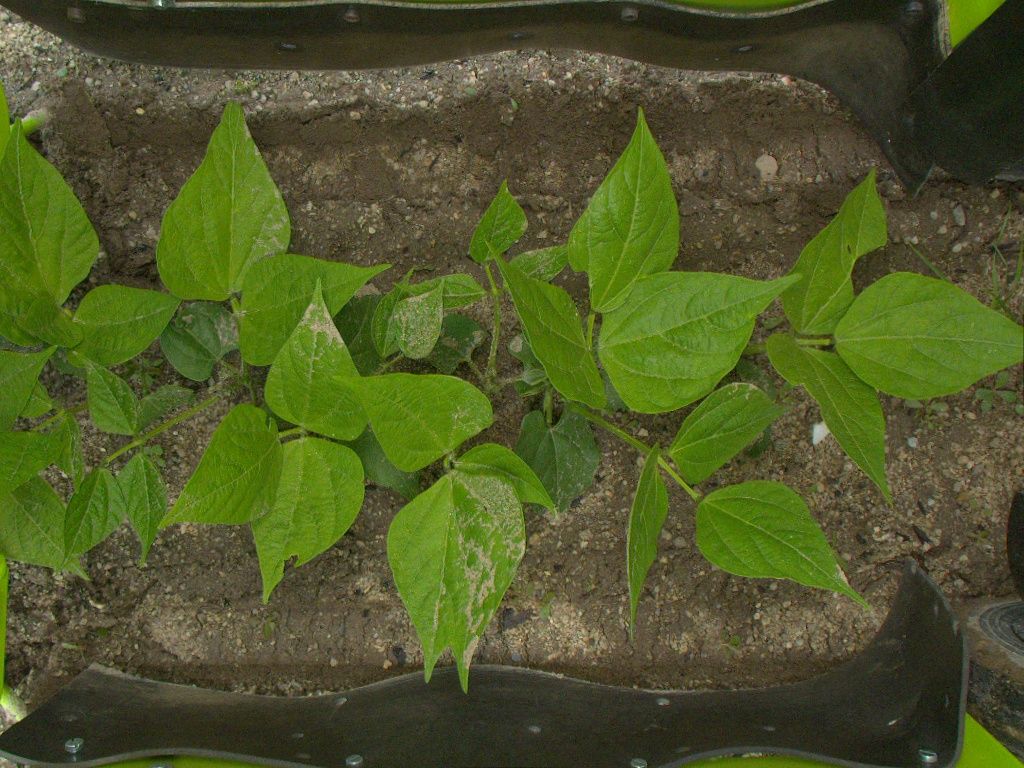}& \includegraphics[width=.15\textwidth]{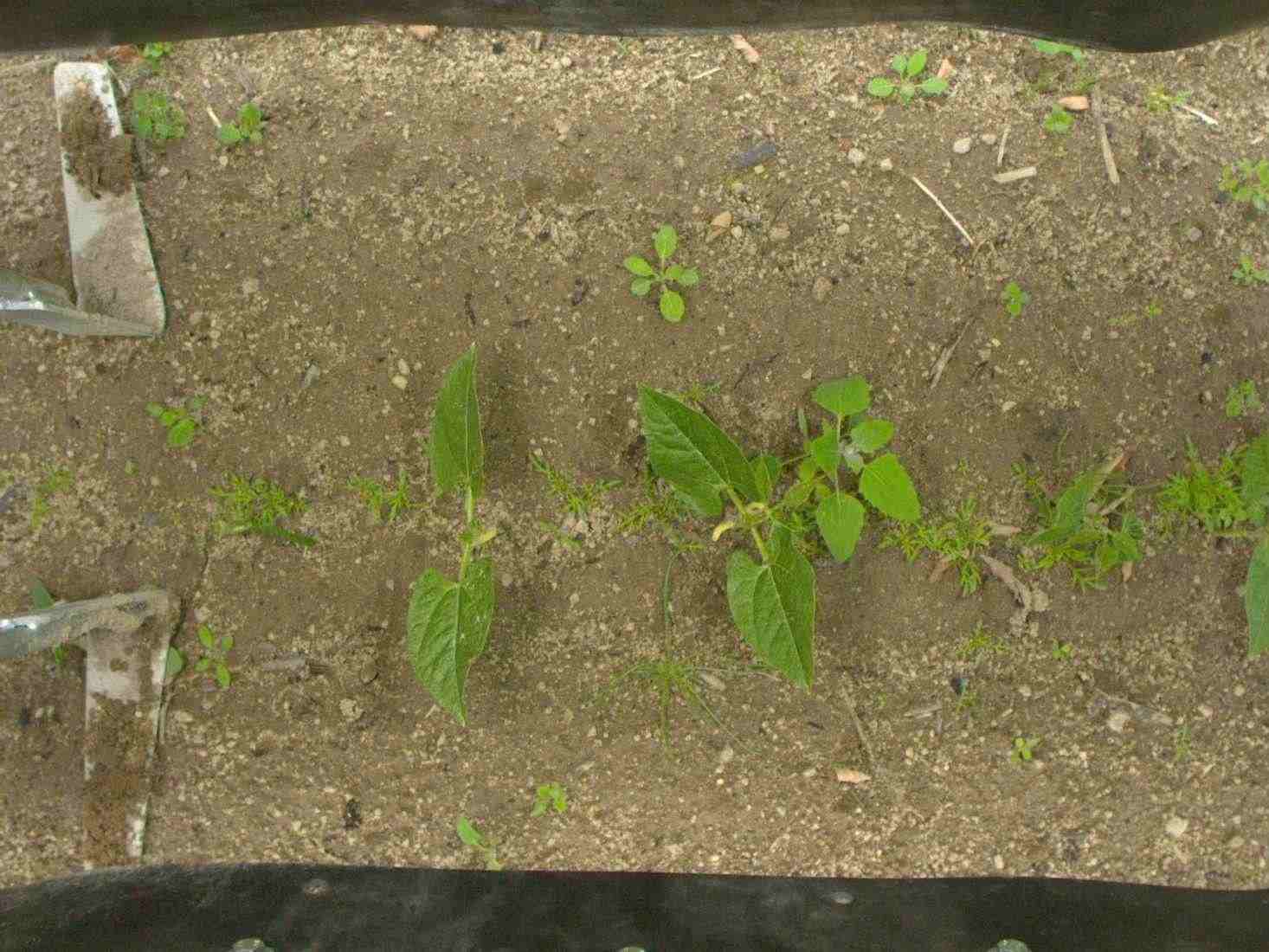} &\includegraphics[width=.15\textwidth]{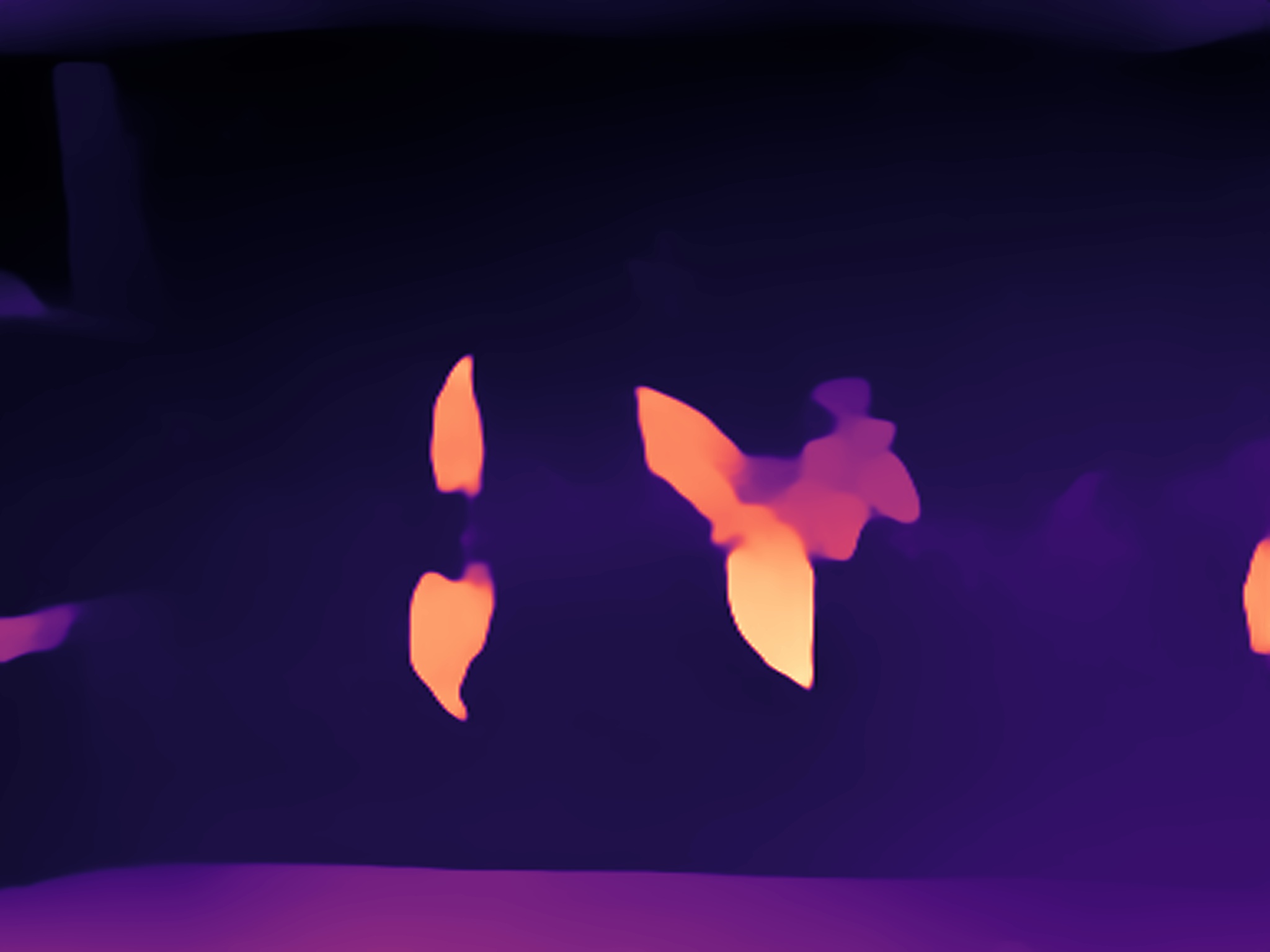}& \includegraphics[width=.15\textwidth]{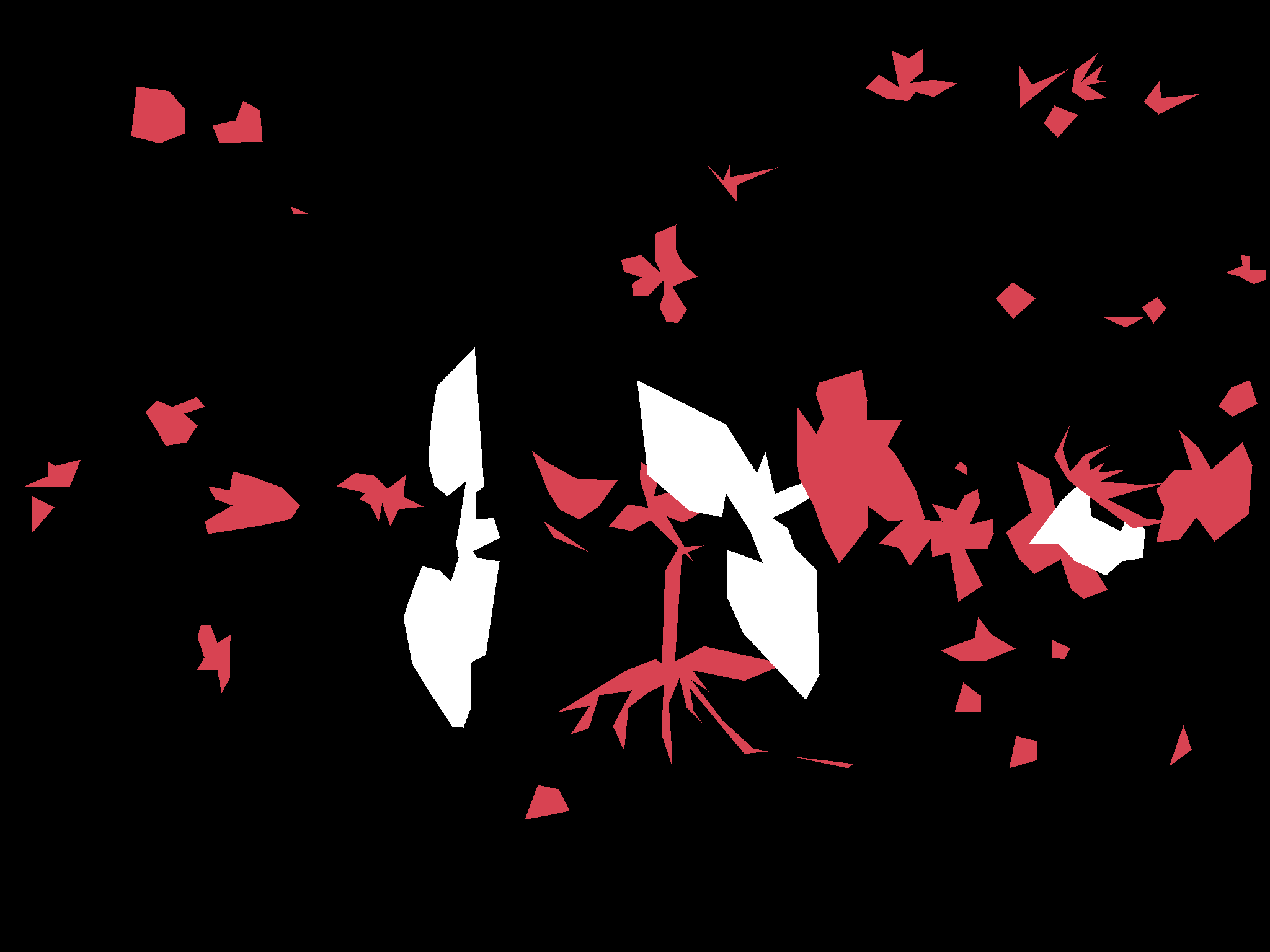}& \includegraphics[width=.15\textwidth]{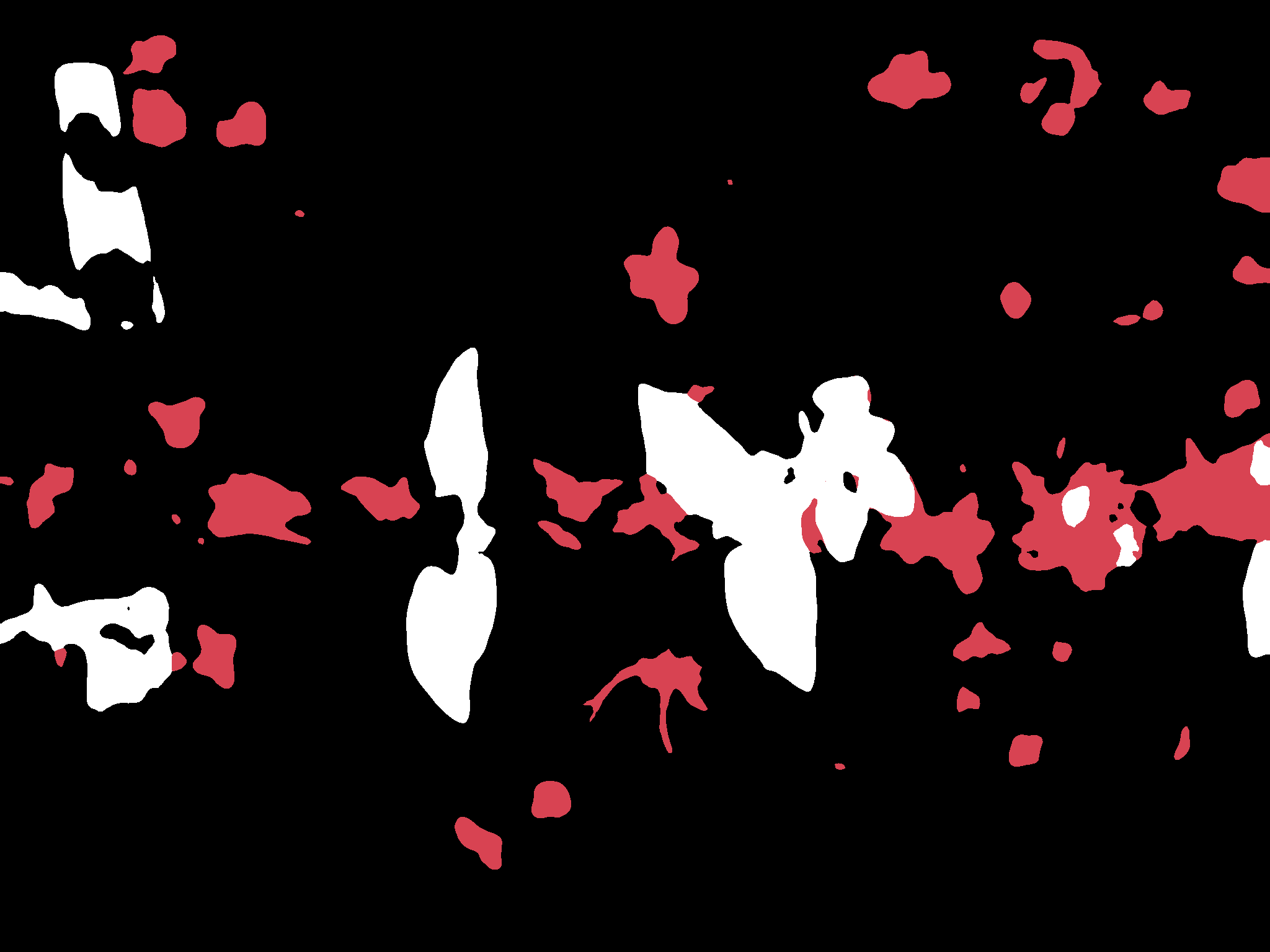}& \includegraphics[width=.15\textwidth]{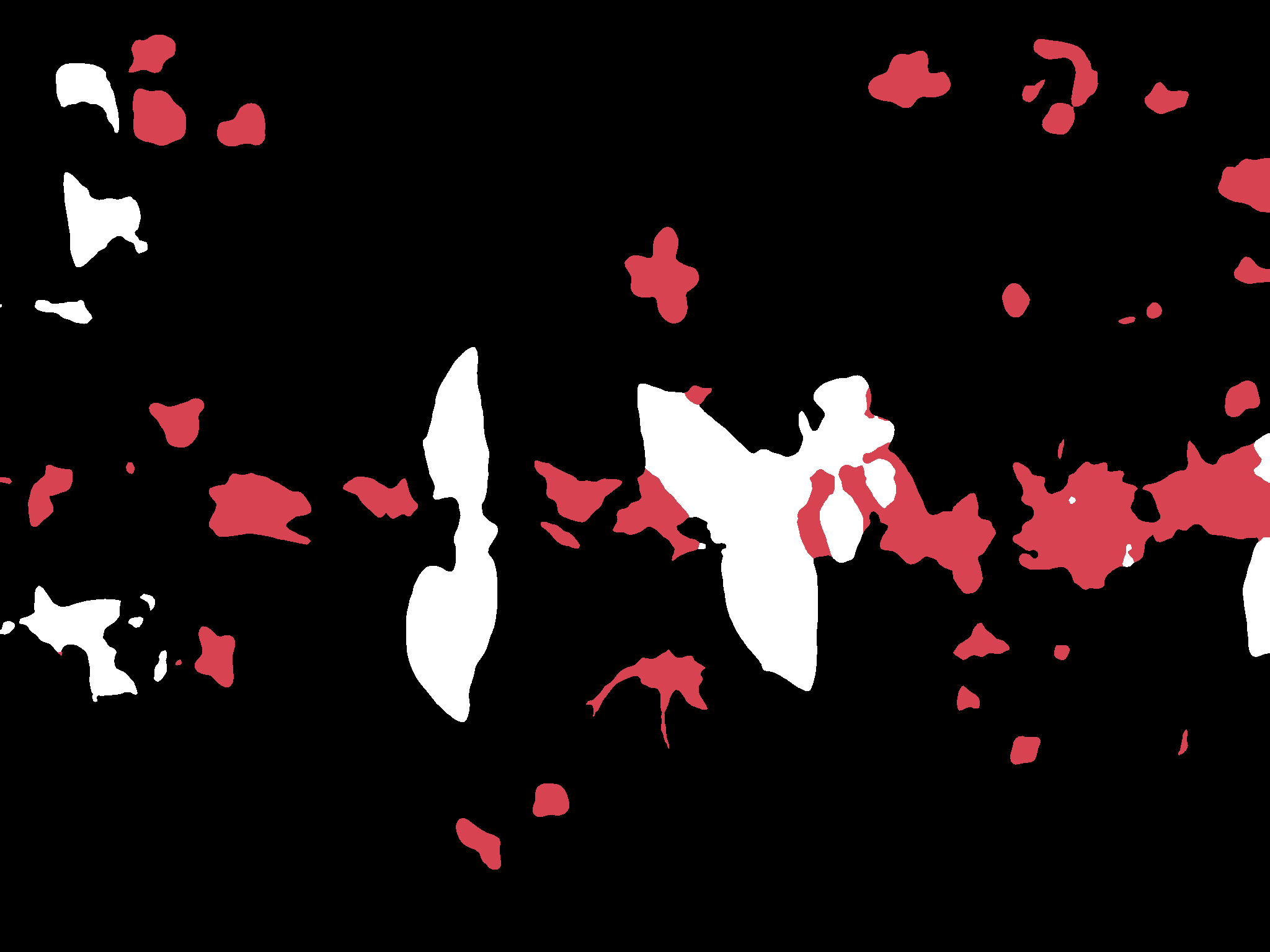}&\includegraphics[width=.15\textwidth]{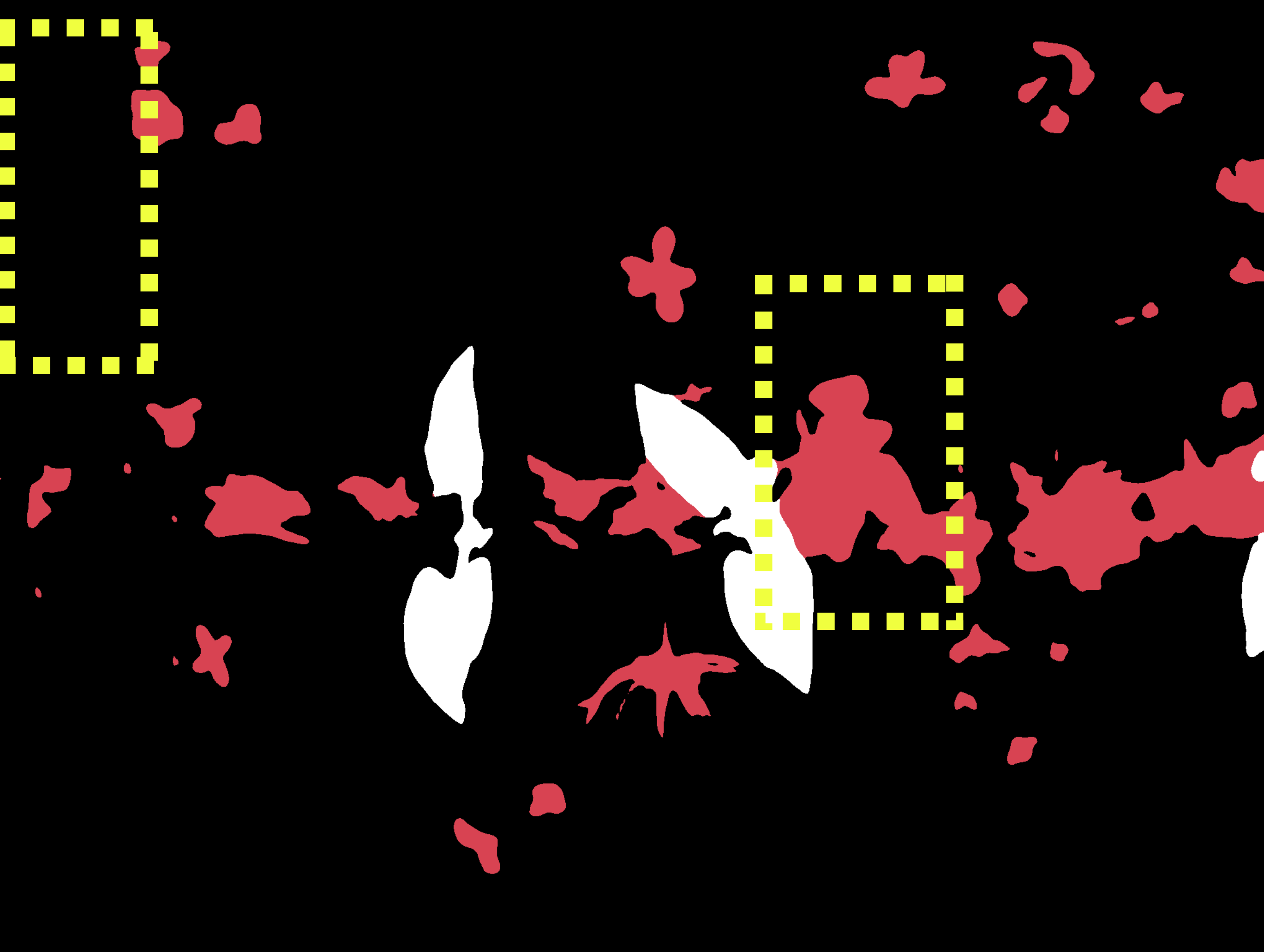}\\
 \includegraphics[width=.15\textwidth]{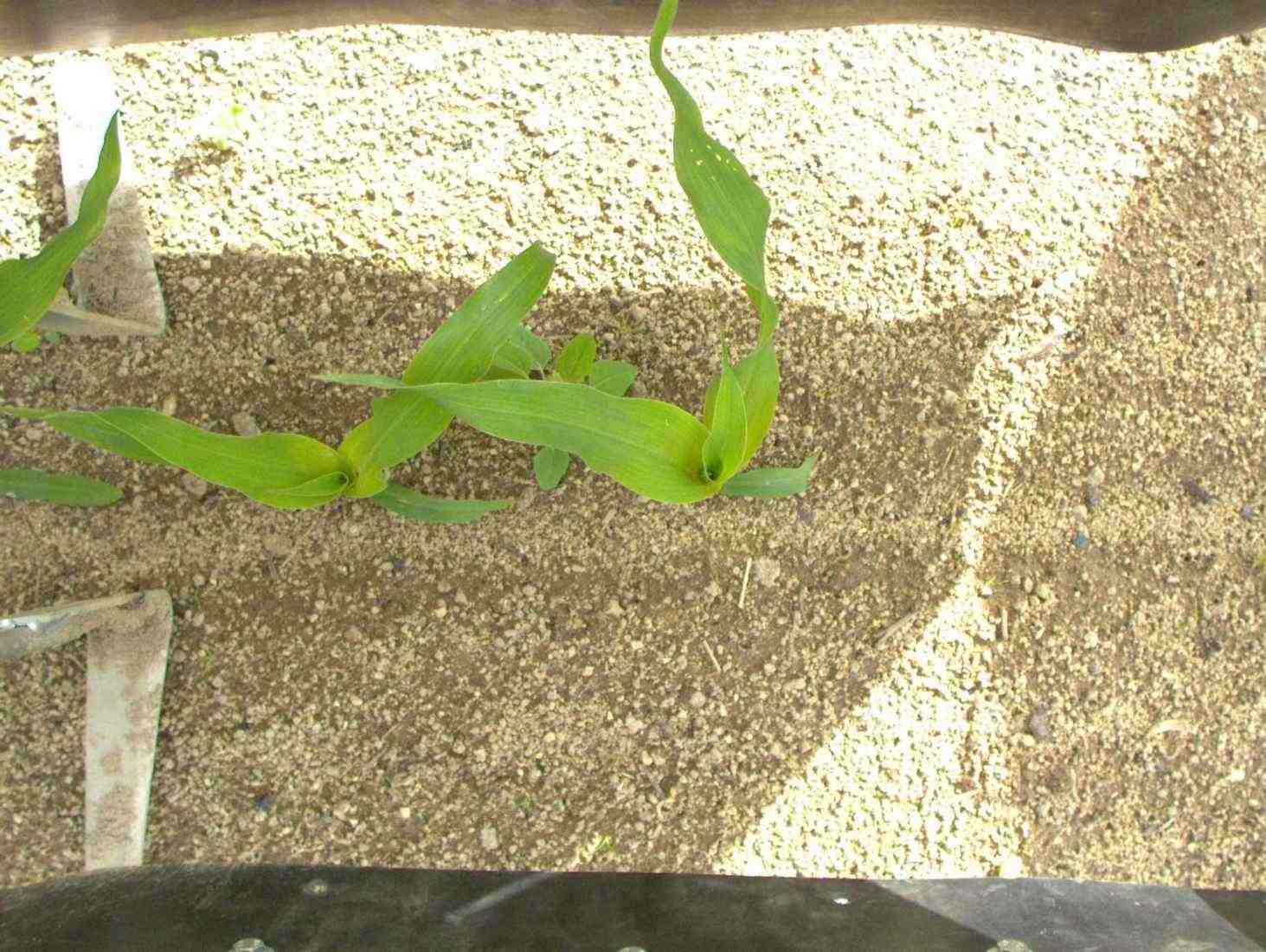}& \includegraphics[width=.15\textwidth]{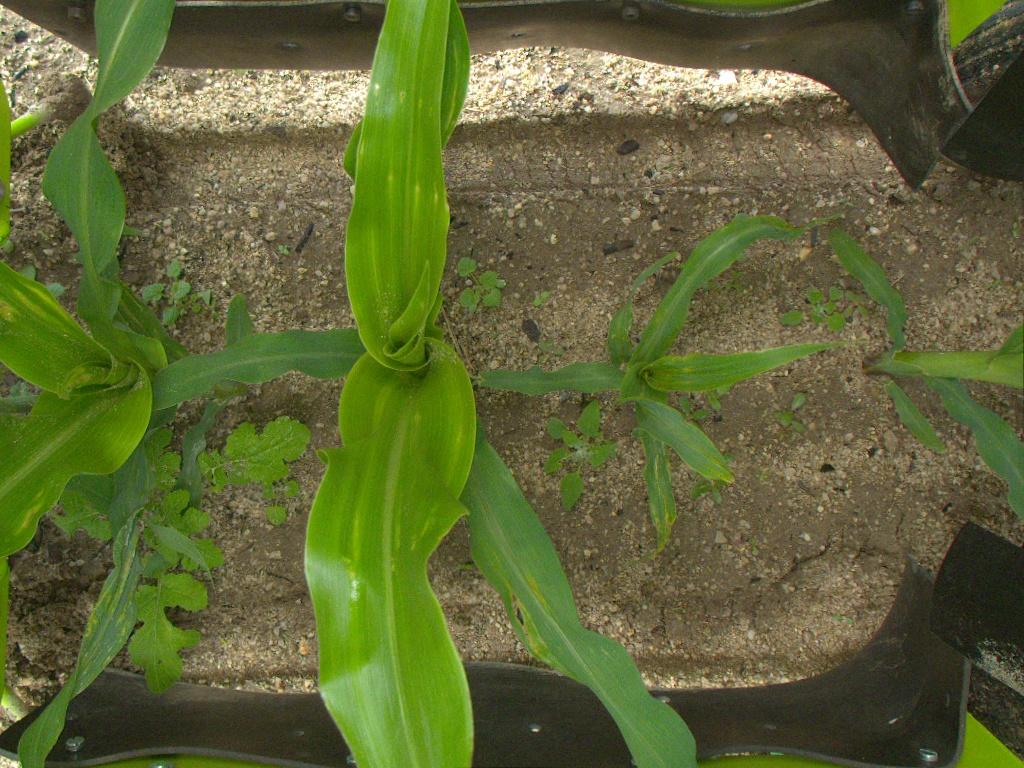} &\includegraphics[width=.15\textwidth]{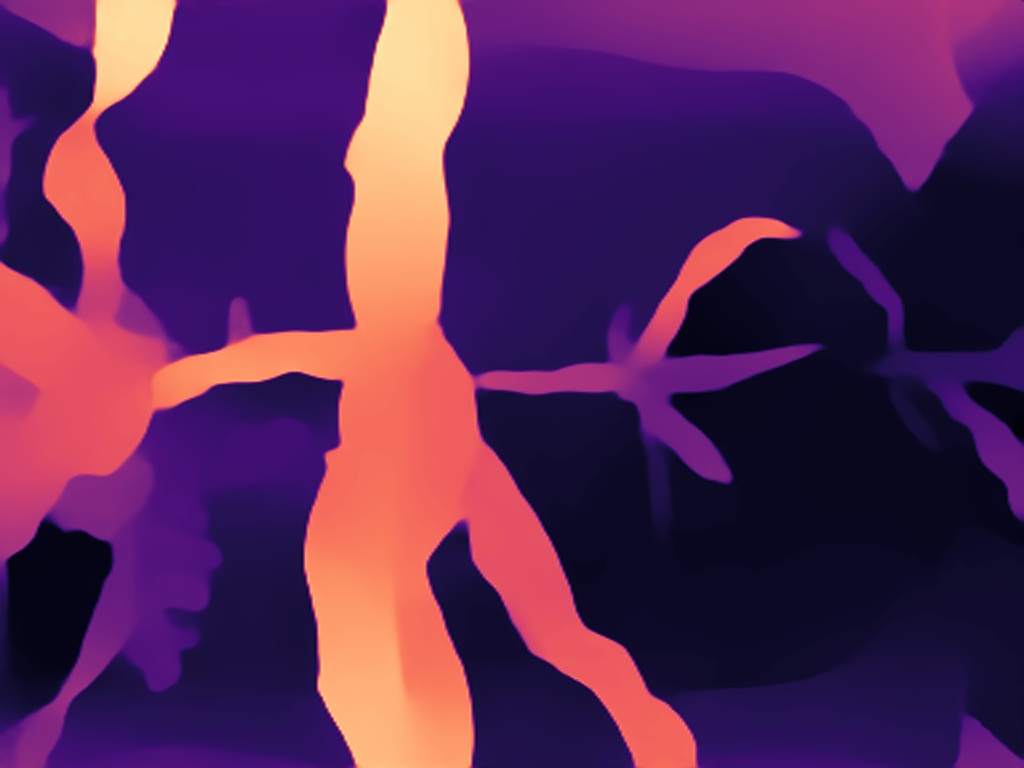}& \includegraphics[width=.15\textwidth]{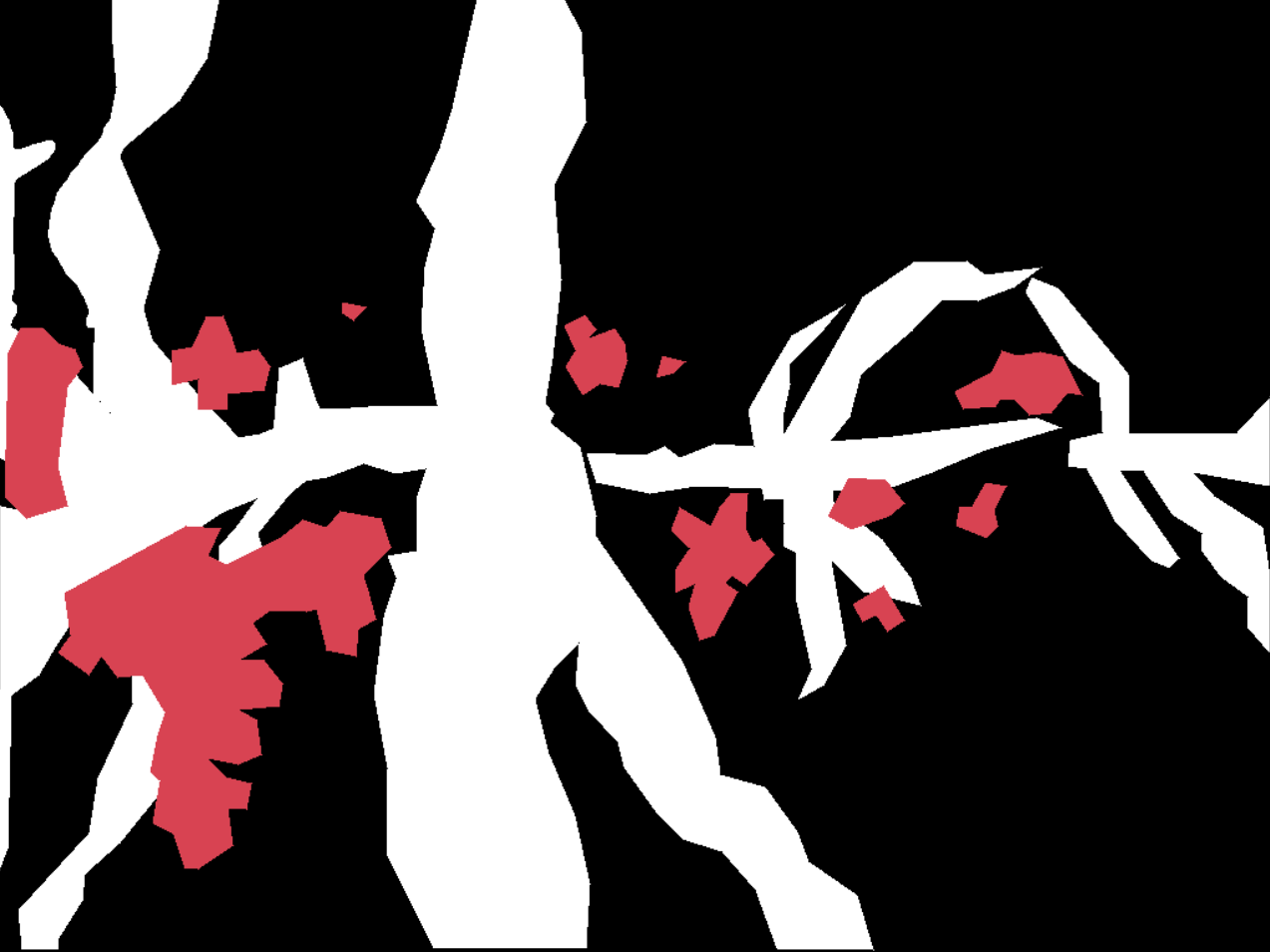}& \includegraphics[width=.15\textwidth]{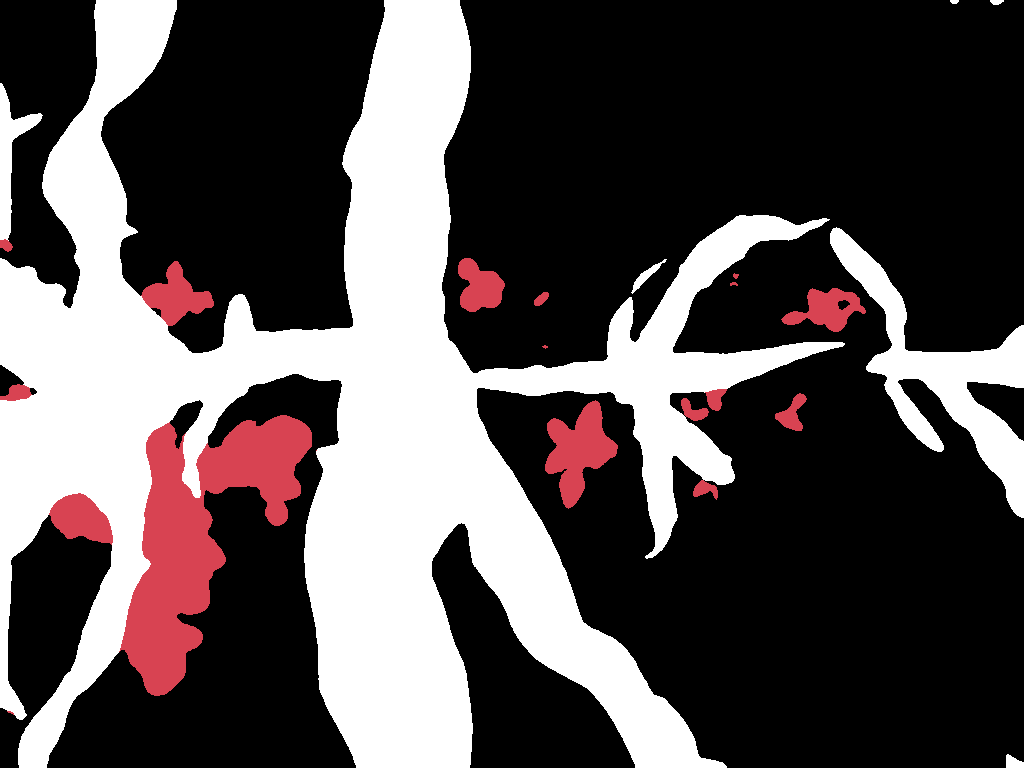}& \includegraphics[width=.15\textwidth]{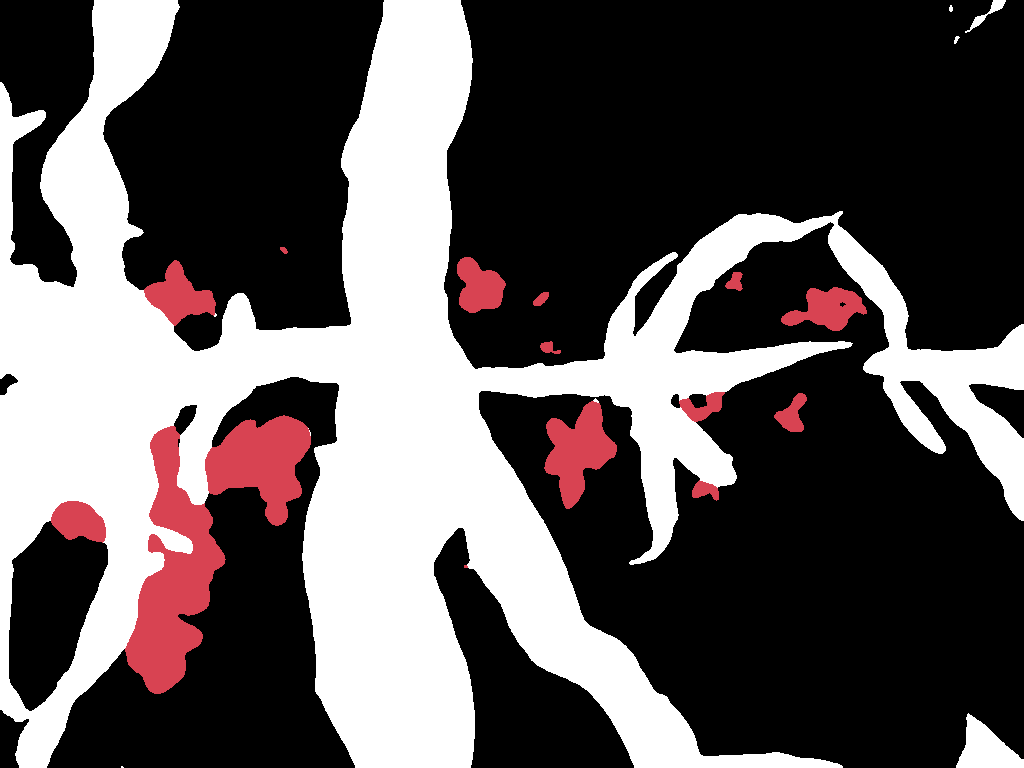}&\includegraphics[width=.15\textwidth]{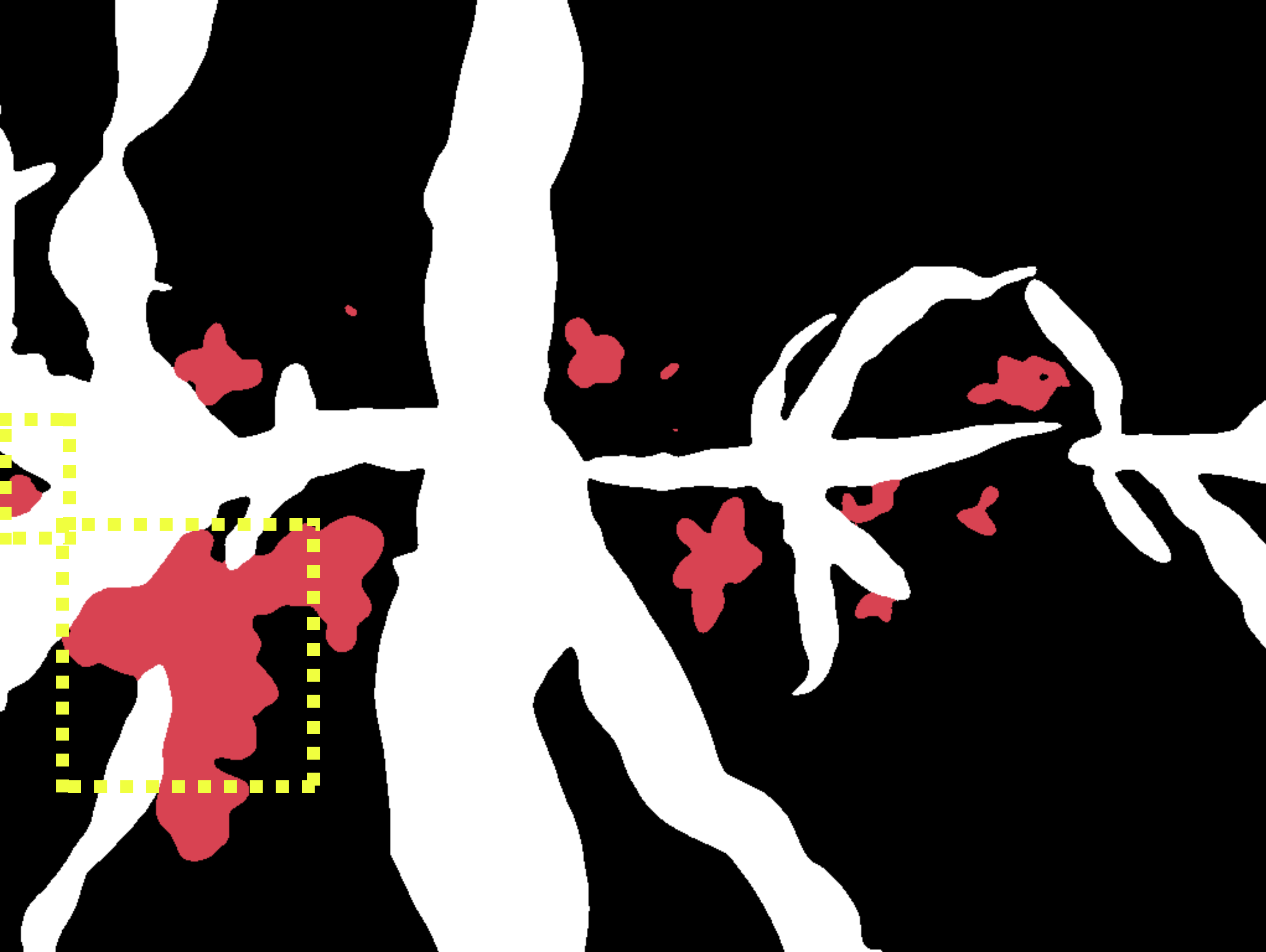}\\
 \includegraphics[width=.15\textwidth]{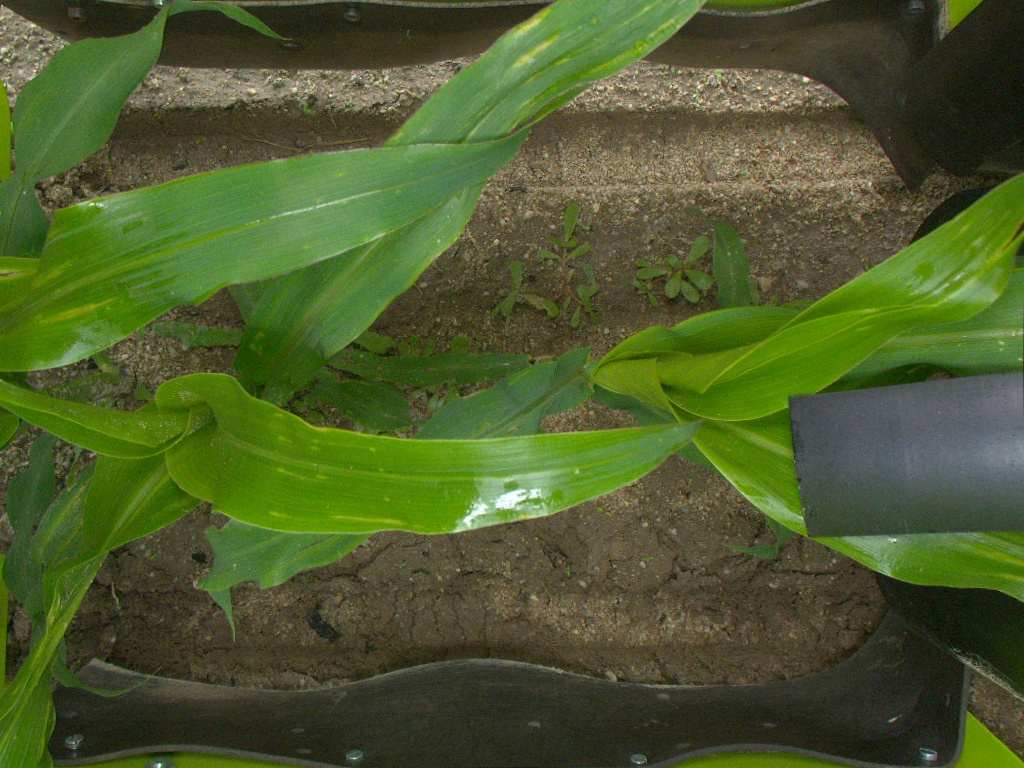}& \includegraphics[width=.15\textwidth]{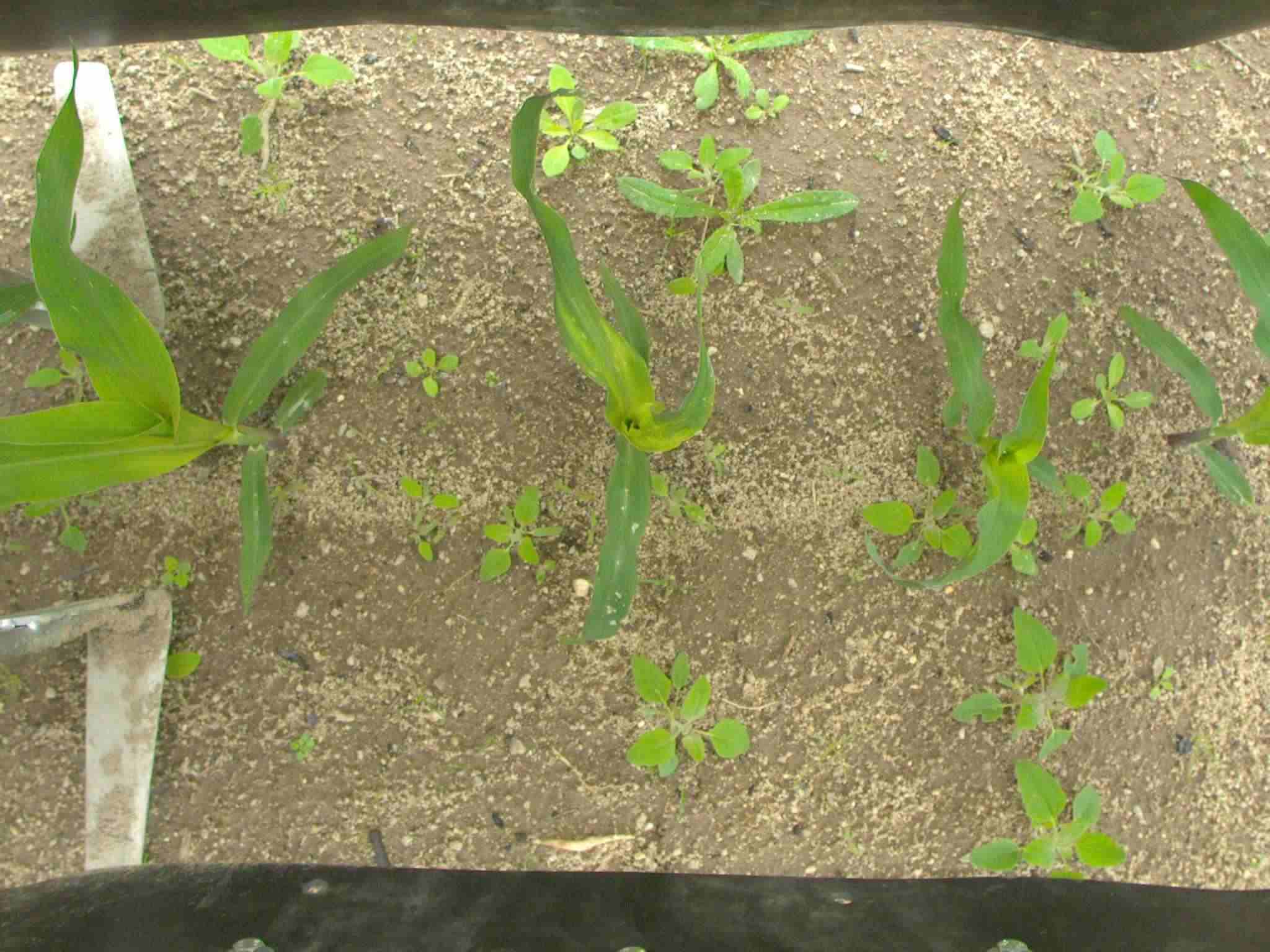} &\includegraphics[width=.15\textwidth]{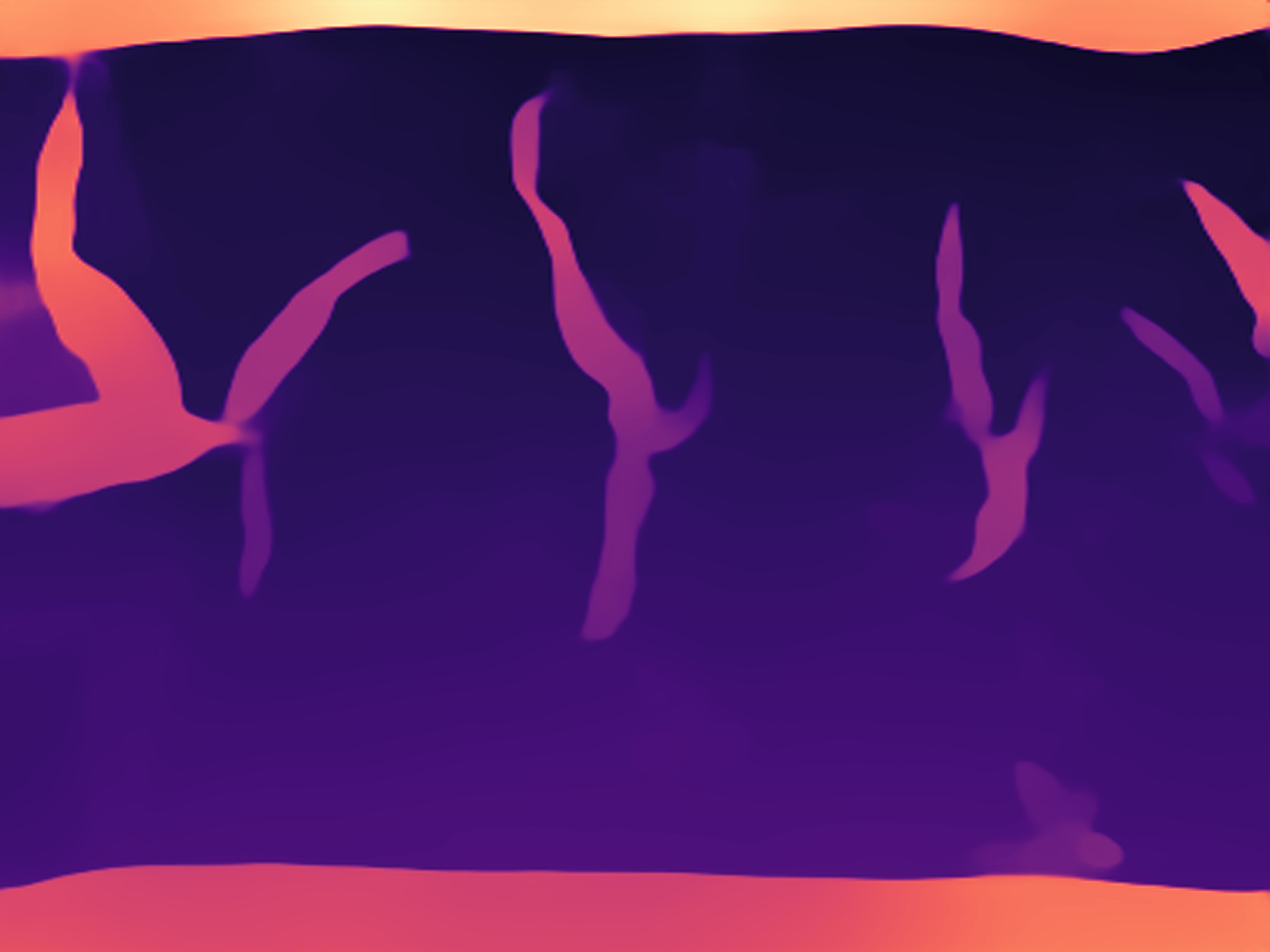}& \includegraphics[width=.15\textwidth]{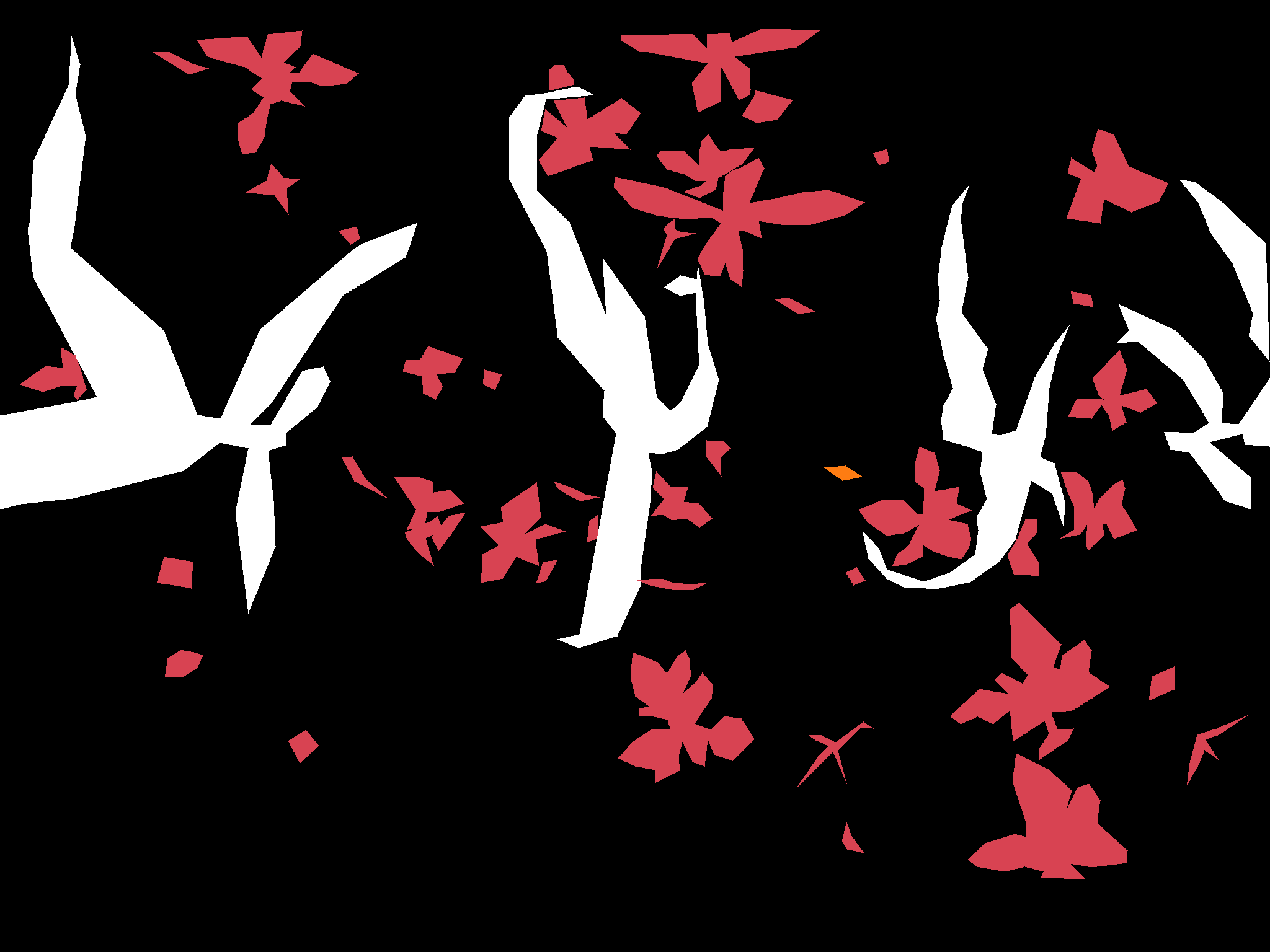}& \includegraphics[width=.15\textwidth]{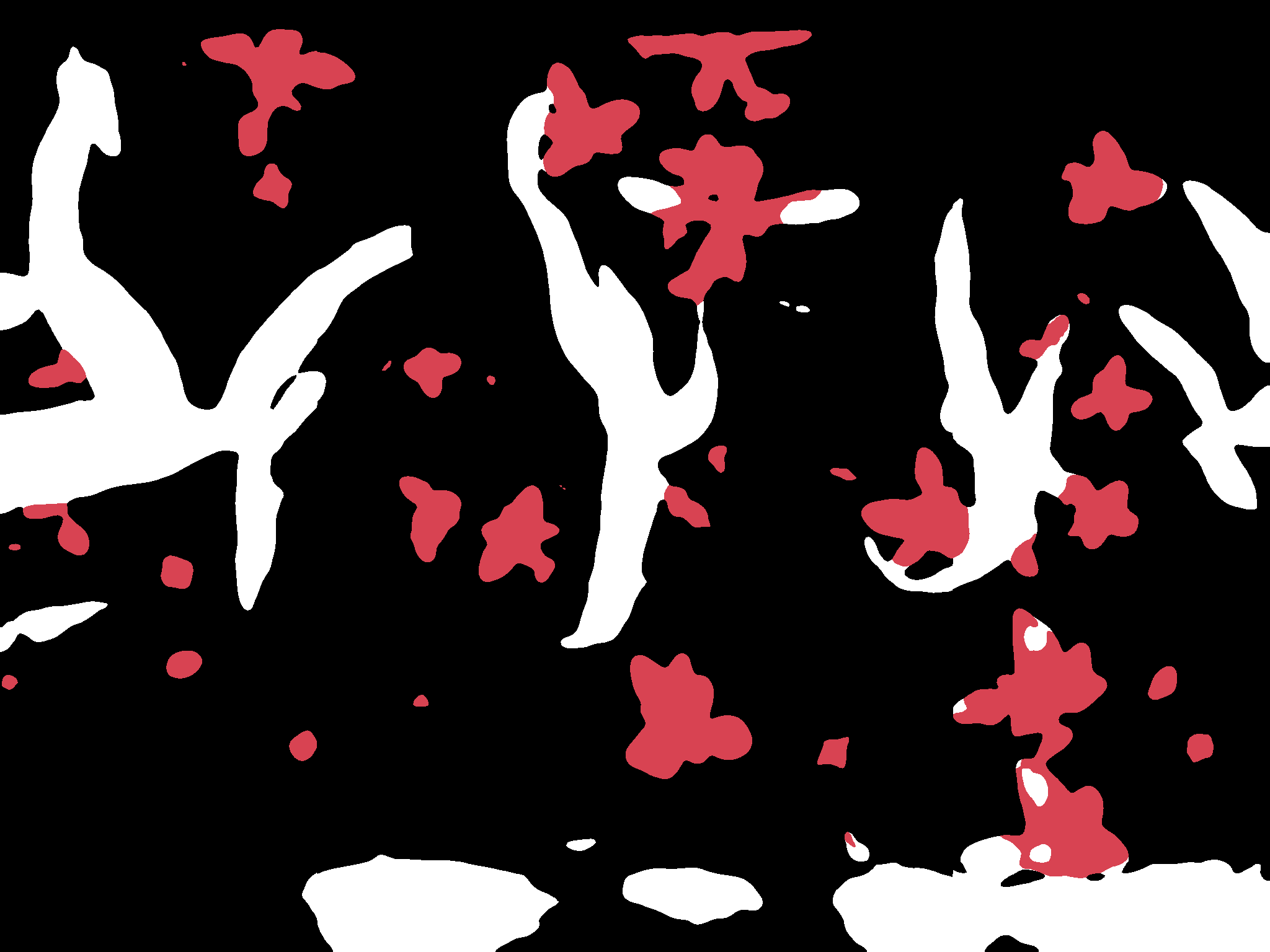}& \includegraphics[width=.15\textwidth]{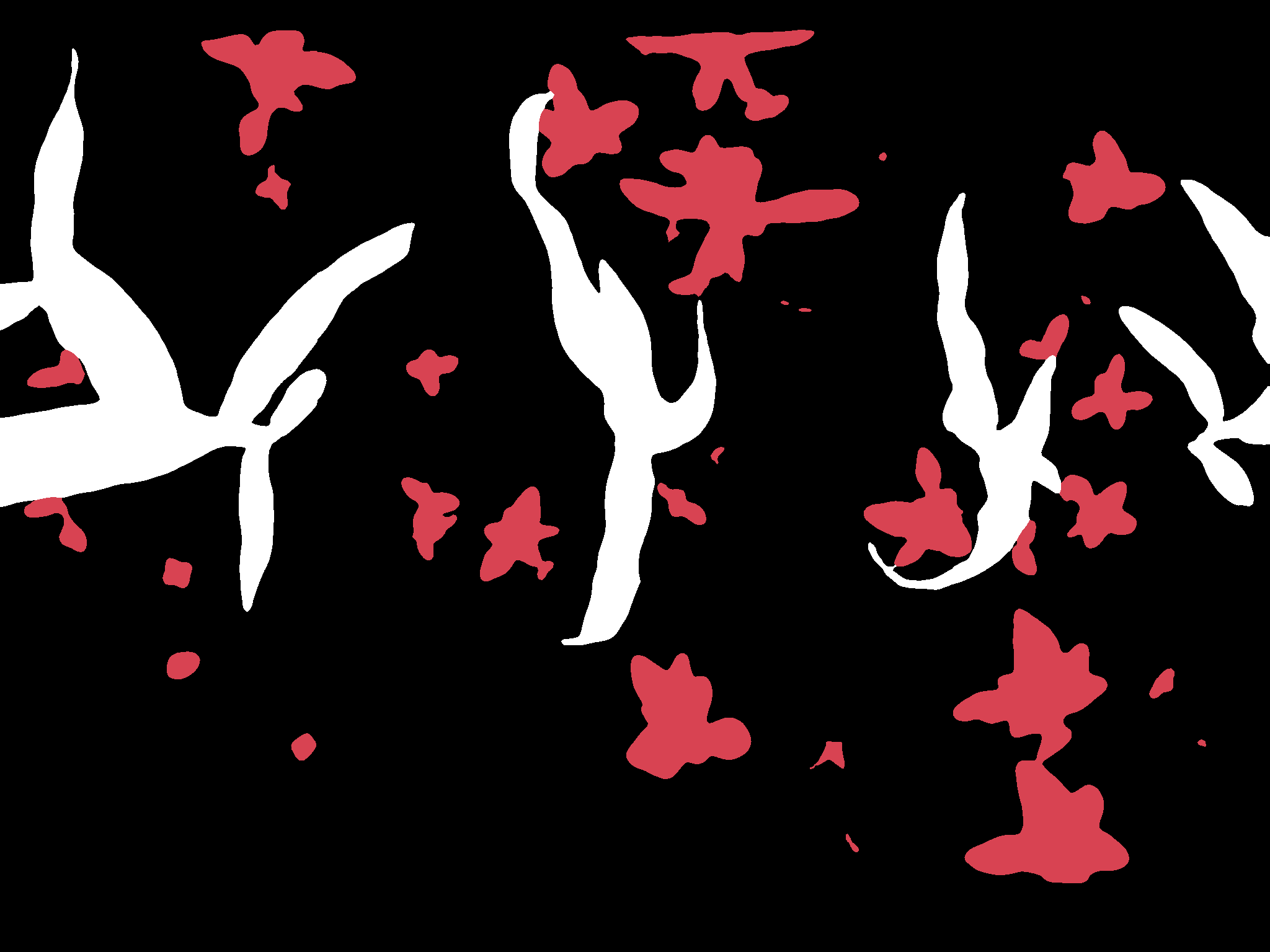}&\includegraphics[width=.15\textwidth]{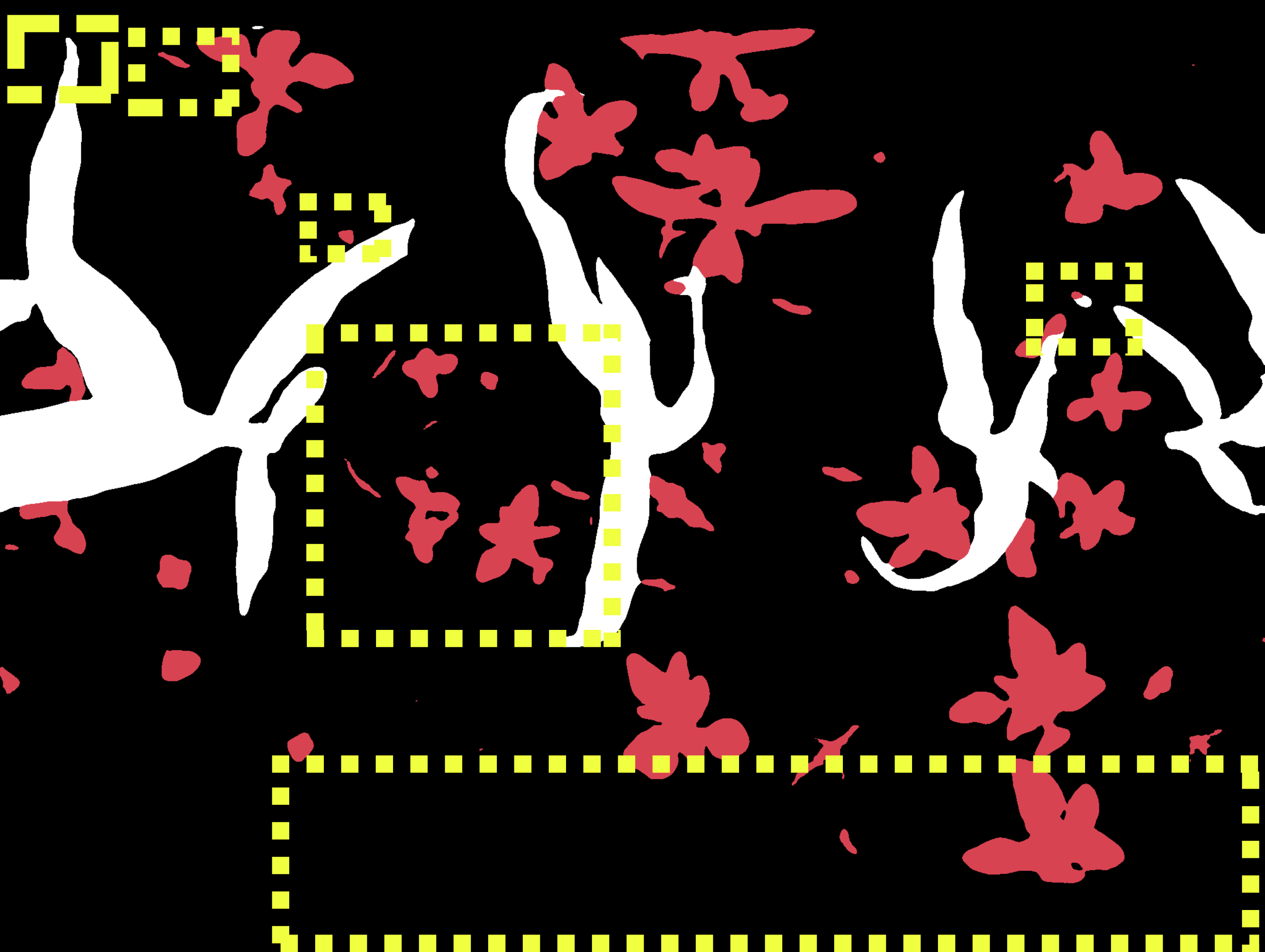}\\
 (a) Source & (b) Target &(c) Depth Map& (d) Ground Truth& (e) HRDA\cite{hoyer2022hrda}& (f) MIC\cite{hoyer2023mic}& (g) Ours\\
\end{tabular}}
\end{center}
\vspace{-3mm}
\caption{Qualitative results are shown for Bean BIPBIP$\rightarrow$WeedElec (Row 1), Bean WeedElec$\rightarrow$BIPBIP (Row 2), Maize BIPBIP$\rightarrow$WeedElec (Row 3), Maize WeedElec$\rightarrow$BIPBIP (Row 4), Bean 2019$\rightarrow$2021 (Row 5), Bean 2021$\rightarrow$2019 (Row 6), Maize 2019$\rightarrow$2021 (Row 7), and Maize 2021$\rightarrow$2019 (Row 8). Dotted rectangles highlight regions with major improvements. MaskAdapt outperforms all SOTA UDA methods and significantly enhances performance over baseline (MIC\cite{hoyer2023mic}), excelling in distinguishing weeds and soil from desired crops.}
\label{seg_comp}
\end{figure*}

\vspace{1mm}\noindent\textbf{Training}. We use the AdamW optimizer~\cite{loshchilov2017decoupled} with learning rates of $5 \times 10^{-5}$ for the depth encoder and $5 \times 10^{-4}$ for the decoder. The learning rate employs a linear warm-up over the initial 1.5k iterations, followed by polynomial decay with a 0.9 factor. An EMA~\cite{tarvainen2017mean}  teacher updates with a momentum of $\alpha = 0.999$ at each step. We set the batch size to two~\cite{hoyer2022daformer,hoyer2023mic}, applying augmentations such as color jitter, Gaussian blur, and mixing of cross-domain classes. For a baseline comparison, we pre-train MIC\cite{hoyer2023mic} in our dataset with its intrinsic masking strategy to establish reference performance. For the implementation of our MaskAdapt method, we pre-train MIC on the same dataset, incorporating geometry-aware horizontal, vertical, and stochastic masking strategies; initially, masking is applied to the source domain, switching to the target domain when pseudo-label confidence reaches a threshold of 90\% to enhance domain adaptation. The RGB encoder and decode head leverage the pre-trained weights~\cite{hoyer2023mic}, the depth encoder uses ImageNet weights, and the RGB encoder remains frozen. Subsequently, we implemented the enhanced feature fusion module and trained the depth encoder. Training occurs over 10k iterations on source and target datasets with cross-entropy loss for both domains, supplemented by periodic unmasked forward pass to align feature distributions. The experiments were performed on an NVIDIA-GeForce RTX-3090 GPU with 24GB of memory, taking 19 hours with MIC on MaskAdapt. Evaluation metrics include classwise IoU and mIoU.

\subsection{Comparison with the SOTA UDA Methods}
To evaluate the performance of our method, we compare it with our baseline approach, MIC~\cite{hoyer2023mic}, and eight SOTA UDA methods, including three transformer-based approaches~\cite{hoyer2022hrda,hoyer2022daformer,hoyer2023mic} designed for challenging domain shifts. The results are presented in Table~\ref{SOTA_comp}, which shows the superior performance of our method in adapting to datasets with different agricultural fields, camera configurations, and stages of crop growth.
For same-year and growth stages but different camera and agriculture field-based adaptations, MaskAdapt achieves an improvement in mIOU compared to the existing best methods across all pairs. The improvement ranges from a minimum of $2.02\%$ on Bean BIPBIP-to-WeedElec to a maximum of $4.47\%$ on Maize WeedElec-to-BIPBIP, with weed and crop IOU increasing by at least $2.98\%$ and $2.81\%$ to a maximum of $11.63\%$and $5.72\%$ , respectively.

\begin{table}[t]
\centering
\caption{Performance Evaluation of Masking Strategies and Types}
\label{tab:combined_masking}
\resizebox{\columnwidth}{!}{
\begin{tabular}{p{5cm}cc|c}
    \toprule
    \textbf{Masking Strategy/Type} & \textbf{RGB} & \textbf{D} & \textbf{mIOU ($\uparrow$)} \\
    \midrule
    Baseline (Stochastic)& \checkmark & $\times$          & 78.98 $\pm$ 0.20 \\
    Vertical Only                  & \checkmark & \checkmark & 84.72 $\pm$ 0.15\\
    Horizontal Only                & \checkmark & \checkmark & 83.65$\pm$ 0.15\\
    Dynamic $m_t$ + Stochastic& \checkmark & \checkmark & 82.78 $\pm$ 0.15\\
    Dynamic $m_t$+ Vertical + Horizontal + Stochastic& \checkmark & \checkmark & 87.27 $\pm$ 0.13\\ \hline
    Complementary + Dynamic $m_t$ + Vertical + Horizontal + Stochastic& \checkmark & \checkmark & \textbf{88.58 $\pm$ 0.10}\\
    \bottomrule
\end{tabular}}
\vspace{-2mm}
\end{table}
\begin{table}[t]
\centering
\caption{Performance of Our Scheduled Masking Technique: This table shows mIOU for 'Source Only' and 'Target Only' masking versus our scheduled masking approach on a target dataset.}
\label{tab:scheduled_masking}
\resizebox{\columnwidth}{!}{
\begin{tabular}{p{5cm}c}
    \toprule
    \textbf{Approach} & \textbf{mIOU ($\uparrow$)} \\
    \midrule
    Source Only       & 85.75 $\pm$ 0.20 \\
    Target Only       & 87.91 $\pm$ 0.30 \\
    \hline
    Ours (Scheduled)  & \textbf{88.58 $\pm$ 0.10} \\
    \bottomrule
\end{tabular}}
\vspace{-3mm}  
\end{table}

\vspace{1mm}
\noindent Moreover, MaskAdapt also surpasses SOTA UDA methods on bean and maize datasets across varying years and growth stages, demonstrating robustness to temporal and phenological domain shifts. It consistently enhances the mIOU by at least $0.95\%$ on Bean 2021-to-2019  and a maximum of $5.19\%$ on Maize 2021-to-2019. The weed and crop class IOU experience an increase on almost all source-target pairs, with the least improvement of $0.24\%$  and $0.74\%$  to a maximum of $9.37\%$ and $5.58\%$, respectively. MaskAdapt’s contributions complement image masking in UDA as it consistently outperformed MIC~\cite{hoyer2023mic} in all adaptations.
Figure~\ref{seg_comp} validates our Table~\ref{SOTA_comp} results, showing MaskAdapt's superiority over SOTA UDA methods.While existing methods misclassify weed as crop or vice versa, particularly in occluded regions and dense canopies, MaskAdapt leverages depth gradients to correctly identify these challenging cases while maintaining precise boundaries at crop-weed interfaces. The visual comparisons clearly demonstrate MaskAdapt's dual advantage: it not only captures weeds missed by other methods but also preserves boundary accuracy where crops and weeds overlap, outperforming all SOTA methods across growth stages and domain shifts

\subsection{Ablation Studies}
This section explores MaskAdapt ablation studies, dissecting geometry-aware multimodal masking in cross-domain learning, the enhanced feature fusion module, and source-target mask scheduling. Using the Maize WeedElec  $\rightarrow$ BIPBIP dataset, we assess how these elements enhance performance beyond a standard baseline, revealing their roles in achieving robust adaptation for agricultural scenes. 

\subsubsection{Types of Geometry-Aware Masking}
Table~\ref{tab:combined_masking} assesses masking strategies in MaskAdapt on the Maize WeedElec $\rightarrow$ BIPBIP dataset, utilizing geometry-aware masks and dynamic $m_t$, the fraction of input masked. The baseline, employing RGB with stochastic masking, scores a mIOU of 78.98\%, improving significantly to 84. 72\% with vertical masking (+5.74\%)  and depth inclusion, excelling at mixed crop-weed segmentation. Horizontal masking reaches 83.65\%, aiding crop-soil sharp boundaries, while Dynamic $m_t$  along with Stochastic masking, which extends the baseline's stochastic approach by varying masked fractions on both RGB and depth, achieves 82.78\% (+3.80\% over baseline). This modest gain suggests that dynamic masking with depth adds resilience, yet falls short of geometry-aware strategies, underscoring the need for structured masking beyond stochastic variation. Combining dynamic masking with all three types of masking strategies, mIOU increases to 87.27\% (+4.49\%), highlighting the synergy of integrated mask types.

Integrating complementary masking along, achieves mIOU of 88.58\%, a +9.60\% gain over the baseline. This approach, with coordinated complementary masking (e.g., 80\% RGB, 20\% depth), random mask selection per epoch, and dynamic $m_t$ adjustment, adds improvement over the combined dynamic strategy, affirming the power of cross-modal coordination. The complementary framework unifies these strengths, ensuring robust segmentation classes.

\begin{table}[t]
\centering
\caption{Impact of Fusion Components in MaskAdapt: This table compares mIOU for baseline (no fusion), cross-attention, and cross-attention with depth gradients, with the best result in \textbf{bold}.}
\label{tab:fusion_components}
\begin{tabular}{c|c}
    \toprule
    \textbf{Fusion Strategy} & \textbf{mIOU ($\uparrow$)} \\
    \midrule
    Baseline & 78.98 $\pm$ 0.20 \\
    Cross Attention w/o Depth Gradients& 79.46 $\pm$ 0.12 \\
    \hline
    Cross Attention w/ Depth Gradients& \textbf{81.35 $\pm$ 0.10}\\
    \bottomrule
\end{tabular}
\vspace{-3mm}
\end{table}

\subsubsection{Impact of Scheduled Masking}
Our scheduled masking technique, starting with source masking and progressively incorporating target masking as confidence increases across epochs, reaches a mIOU of 88.58\% (+2.83\% over source only, +0.67\% over target only), as shown in Table~\ref{tab:scheduled_masking}. This strategy balances source supervision with target adaptation, reducing early pseudo-label noise compared to simultaneous source-target masking, and outperforms both single-domain extremes.

\subsubsection{Effect of Multimodal Fusion}
Table~\ref{tab:fusion_components} evaluates the enhanced feature fusion module, starting with a baseline mIOU of 78.98\% without depth features. Adding cross-attention without depth gradients increases this by 0.48\%, by aligning RGB and depth features more effectively. Including depth gradients with cross-attention pushes the mIOU to 81.35\%,  sharpening weed and crop boundaries. Cross-attention alone offers a slight boost in feature integration, but adding depth gradients enhances edge detail critical for crop-weed segmentation.

\section{Conclusion}
\label{conc: conclusion}In conclusion, our proposed framework addresses the limitations of semantic segmentation in SSFM under varying field conditions by integrating RGB-depth gradients and geometry-aware multimodal masking strategy to achieve UDA. By leveraging depth-derived spatial transitions, we enhance feature fusion, enabling more precise delineation of crops, weeds, and soil even in challenging scenarios where only RGB-based models struggle. The depth-gradient-guided cross-attention mechanism effectively refines RGB features, sharpening boundaries, and reducing texture ambiguities, while the geometry-aware multimodal masking strategy further strengthens adaptation across different fields. \section*{Acknowledgment}
This research project is supported by an NSERC Alliance - Mitacs Accelerate grant (NSERC Ref: ALLRP 581076 - 22 \& Mitacs Ref: IT34971), titled "Developing Machine Learning Methods for RGB Images to Quantify Crop and Weed Populations Across Agricultural Fields" in partnership with Croptimistic Technology Inc.
{
    \small
    \bibliographystyle{ieeenat_fullname}
    \bibliography{main}
}


\end{document}